\UseRawInputEncoding
\documentclass{article}

\usepackage{microtype}
\usepackage{graphicx}
\usepackage{subcaption}
\usepackage{booktabs} 
\usepackage{multirow}
\usepackage{listings}
\usepackage[table]{xcolor} 
\colorlet{lightgray}{gray!50}
\definecolor{WildStrawberry}{RGB}{120, 180, 230}
\definecolor{Red}{RGB}{251, 214, 214}
\lstdefinestyle{fadedhash}{
  basicstyle=\ttfamily\linespread{1.2}\selectfont,
  breaklines=true,
  breakindent=0pt,
  numbers=left,
  numberstyle=\tiny\color{gray!50},
  frame=single,
  backgroundcolor=\color{gray!10},
  xleftmargin=0.04\textwidth, 
  xrightmargin=0.04\textwidth,
}
\usepackage{tcolorbox}
\tcbuselibrary{theorems,breakable,skins}
\newtcolorbox{myexample}[1]{ 
  breakable,
  enhanced,
  colback=green!5,
  colframe=green!35!black,
  fonttitle=\bfseries,
  title=#1,               
  top=2pt,
  bottom=2pt,
  before skip=4pt,
  after skip=4pt
}

\usepackage{hyperref}

\usepackage[accepted]{icml2026}

\usepackage{amsmath}
\usepackage{amssymb}
\usepackage{mathtools}
\usepackage{amsthm}
\usepackage{changepage}

\usepackage[capitalize,noabbrev]{cleveref}

\theoremstyle{plain}

\theoremstyle{definition}

\theoremstyle{remark}

\usepackage[textsize=tiny]{todonotes}

\icmltitlerunning{CollabBench: Benchmarking and Unleashing Collaborative Ability of LLMs with Diverse Players via Proactive Engagement}

\begin{document}

\twocolumn[
  \icmltitle{CollabBench: Benchmarking and Unleashing Collaborative Ability of LLMs \\ with Diverse Players via Proactive Engagement}



  \icmlsetsymbol{equal}{*}


\begin{icmlauthorlist}
    \icmlauthor{Hong Qian}{edu,inno}
    \icmlauthor{Yuanhao Liu}{edu,ten}
    \icmlauthor{Zihan Zhou}{edu}
    \icmlauthor{Zongbao Zhang}{edu,inno}
    \icmlauthor{Hanjie Ge}{edu}
    \icmlauthor{Haotian Shi}{edu}
    \icmlauthor{Liang Dou}{edu}
    \icmlauthor{Xiangfeng Wang}{edu}
    \icmlauthor{Jingwen Yang}{equal,ten}
    \icmlauthor{Aimin Zhou}{edu,inno}
  \end{icmlauthorlist}
  \icmlaffiliation{edu}{Shanghai Institute of AI for Education, and School of Computer Science and Technology, East China Normal University, Shanghai, China}
  \icmlaffiliation{ten}{Tencent Inc., Shenzhen, China}
    \icmlaffiliation{inno}{Shanghai Innovation Institute, Shanghai, China}

  \icmlcorrespondingauthor{Jingwen Yang}{jingwenyang@tencent.com}

  \icmlkeywords{Collaborative Ability Evaluation, Persona Simulation, Collaborative Agentic Reinforcemnet Learning, Affective Alignment, Cooperative Game, LLM agents}

  \vskip 0.3in
]



\printAffiliationsAndNotice{}  

\begin{abstract}

While LLM-based agents excel at individual tasks, effective collaboration with realistic human partners remains challenging. Most of the existing conversation-level collaborative studies lack grounded interaction and behavioral execution, motivating the need for cooperative game environments that enable contextualized and immersive collaboration. To this end, this paper proposes CollabBench, a benchmark for evaluating and training collaborative agents in cooperative games. CollabBench features a Diverse Player Profile Simulation pipeline to model varied players behaviors, and a Collaborative Agentic Training paradigm that unifies reasoning, communication, and action via agentic rollouts, optimized with a hybrid reward balancing task efficiency and affective adaptation. We further extend classic environments to CWAH-MultiPlayer and Cook-MultiPlayer for systematic evaluation under diverse personalities. Experiments with efficiency and affective metrics show that our trained models outperform base models, achieving 19.5\% higher efficiency and 24.4\% improved affective performance. Further analysis reveals key collaborative limitations of existing models and offers insights for future collaborative training.
\end{abstract}
\begin{figure*}[!t]
\centering
\includegraphics[width=0.99\linewidth]{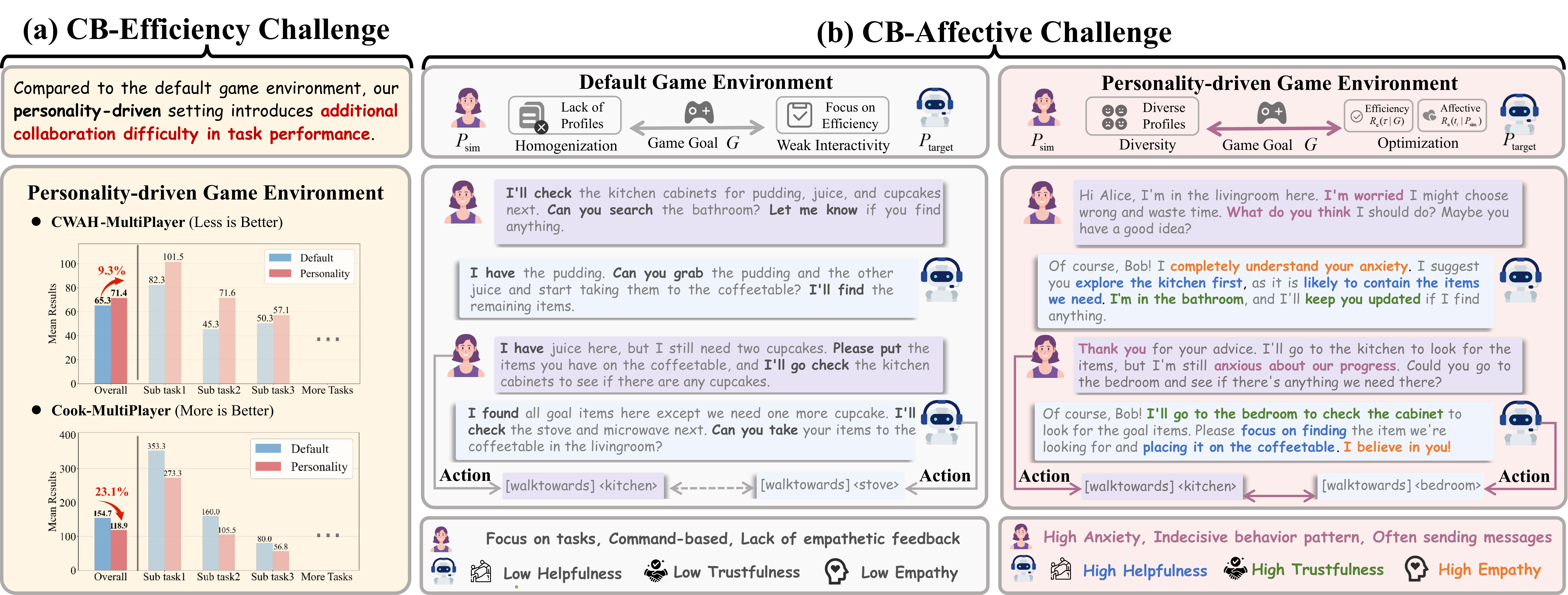}
\caption{The motivation of this work: challenges of collaboration between real human and agent. \textbf{(a) CB-Efficiency Challenge}: Introducing diverse players significantly increases task difficulty. The Personality-driven setting (red) degrades performance compared to standard homogeneous multi-agent cooperation (blue), resulting in more steps in CWAH (+9.3\%) and lower scores in Overcook (-23.1\%). \textbf{(b) CB-Affective Challenge}: The Default environment (left) prioritizes efficiency, leading to command-based, low-empathy interactions. In contrast, the Personality-driven environment (right) introduces diverse teammates, requiring the agent to optimize both efficiency and affective engagement. Detailed results on motivation are in Appendix~\ref{apx:moti}.}
\label{fig:motivation}
\end{figure*}
\section{Introduction}\label{sec:intro}

With the rapid advancement of Large Language Models (LLMs), LLM-based agents~\cite{WangMFZYZCTCLZWW24,XiCGHDHZWJZZFWXZWJZLYDW25,zhang2025landscape} have demonstrated strong capabilities in individual tasks such as deep research, mathematical reasoning, and coding. As these abilities expand, a critical research challenge is enabling agents to collaborate effectively with human partners exhibiting diverse personalities and behavioral patterns. Most existing human--agent collaboration studies focus on conversation-level tasks~\cite{WangYZQSBN024,WuGP0LDC0L025,zhang2025echo}, such as dialogue, document editing, and mathematical problem solving, which typically exhibit weak context and interaction, often detaching communication from the immediate shared situation and lacking grounded behavioral execution. This motivates the development of agents operating in immersive cooperative game environments~\cite{hu2024survey} for context scaling~\cite{wang2026alphacontext}, requiring agents to share a grounded environment and engage in proactive interaction beyond purely linguistic exchange, thereby unleashing agent's contextualized collaborative intelligence in both efficient task execution and affectively adaptive engagement with diverse human partners.

Despite growing interest, several challenges remain. First, simulating players with diverse personalities and behavioral styles through LLMs in game environments remains difficult (\emph{Challenge 1}). Existing role-playing and user profiling approaches~\cite{WangPQLZWGGN00024,TuFTSSGY24,0001WZY00YGC00X25,SalemiMBZ24} largely rely on predefined characters or static user data, and fail to capture grounded, action-level behaviors in interactive games. Second, unleashing LLM-based agents’ collaboration with diverse partners in strong context such as game environments is largely unexplored (\emph{Challenge 2}). Prior work predominantly focuses on improving single-agent capabilities~\cite{feng2025group,wang2025ragen,xi2025agentgymrl} or multi-agent architectures~\cite{ZhangDSZDTSG24,ZhangYHWLSZZLZC24,SeoNLLLK25}, with limited emphasis on learning collaborative awareness and adaptability within LLMs. Third, current criteria for LLM-based agents in game environments~\cite{costarelli2024gamebench,xi2025agentgym,hu2025lmgame} are often restricted to efficiency-based metrics, such as final game scores (\emph{Challenge 3}), overlooking affective and social qualities that are essential for effective human-agent collaboration.

Therefore, this paper introduces CollabBench (CB), a benchmark for systematically evaluating and training LLM-based agents to collaborate proactively with diverse players. To address Challenge 1, we develop a Diverse Players Profiles Simulation pipeline that generates heterogeneous behaviors grounded in Big Five personality theory, followed by high-fidelity profile modeling that enforces explicit personality-behavior mappings and applies interaction-based filtering to ensure realism and consistency. Based on this environment, we observe that player heterogeneity substantially increases task difficulty compared to standard homogeneous multi-agent settings (Figure~\ref{fig:motivation}(a)), and that efficiency-driven interaction strategies fail to satisfy affective needs of diverse partners (Figure~\ref{fig:motivation}(b)). Motivated by these findings, we propose Collaborative Agentic Training to address Challenge 2, featuring a unified agentic rollout that integrates reasoning, communication, and action, together with a hybrid reward mechanism that jointly optimizes trajectory-level task efficiency and step-level affective alignment via agentic RL.


Building on our diverse players profiles simulation, we extend two classic environments to construct CWAH-MultiPlayer and Cook-MultiPlayer for both training and evaluation, introducing substantially more challenging and realistic collaborative settings with heterogeneous players. To address Challenge 3, we design a comprehensive evaluation protocol combining efficiency metrics (e.g., game score, variance and token usage) with affective metrics (Helpfulness, Trustfulness, and Empathy). Empirical results show that our collaborative training paradigm yields substantial gains, improving average game scores and affective metrics by 19.5\% and 24.4\%, respectively, over the base model. We further provide detailed analyses revealing the collaborative limitations of existing LLMs and offer extensive experimental insights into the efficacy of collaborative training.


\section{Related Work}

\textbf{Personality-Aware Human Simulation.} Traditional human simulators are learned from collected behavior data~\cite{CarrollSHGSAD19}. Recent work shows that LLMs can flexibly generate diverse personas, and has applied this capability to role-playing and personalized services using predefined character descriptions or user profiles~\cite{kosinski2024evaluating,SalminenLPCHJ24,ShinHRLO24,WangPQLZWGGN00024,TuFTSSGY24,0001WZY00YGC00X25,SalemiMBZ24}. However, such approaches are limited for cooperative game environments, where data for diverse, grounded behaviors is scarce. Existing human-LLM collaboration systems rely on task-driven, conversation-level simulators with limited personality diversity~\cite{WuGP0LDC0L025}, while domain-specific LLM-based simulations in dialogue, social activities, and education~\cite{WangYZQSBN024,zhang2025echo,ParkOCMLB23,LiuYLC24} do not readily transfer to cooperative game tasks.

\textbf{Agentic Training for LLMs in Games.} Recent work has demonstrated strong performance of agentic training for LLMs in domains such as deep research, mathematical reasoning, and coding~\cite{jin2025search,guo2025deepseek,jiang2024survey}. Game environments provide a practical and realistic testbed for this paradigm~\cite{Chevalier-Boisvert19,CarrollSHGSAD19,ShridharYCBTH21,PuigSLWLTF021,WangX0MXZFA24}. Prior studies predominantly focus on enhancing single-agent capabilities through architecture design or agentic reinforcement learning, including VOYAGER~\cite{WangX0MXZFA24}, RAGEN~\cite{wang2025ragen}, and AgentGym-RL~\cite{xi2025agentgymrl}. Research on collaborative agents remains largely architecture-centric, as exemplified by ProAgent~\cite{ZhangYHWLSZZLZC24} and CoELA~\cite{ZhangDSZDTSG24}, which improve coordination via specialized modules. However, these approaches generally lack explicit mechanisms for training collaboration awareness within LLMs and overlook the diversity of teammate personalities in real-world collaboration.

\textbf{Evaluation of LLM-based Agent and Human Collaboration.} As LLM capabilities advance, effective collaboration has become a critical objective, where efficiency alone is insufficient as an evaluation criterion~\cite{george1990personality,mcallister1995affect,salas2005there}. Recent work has begun to adopt multidimensional evaluation frameworks, such as LLM-as-a-judge for assessing interactivity~\cite{WuGP0LDC0L025} and empathy~\cite{zhang2025echo}. However, gamified evaluations of LLM-based agents still predominantly rely on efficiency-oriented metrics, including game scores and success rates~\cite{costarelli2024gamebench,hu2025lmgame,sun2025collab,zhang2025paracook}. Although some cooperative game studies incorporate affective aspects via user surveys~\cite{SiuPCZLPCA21,ZhangDSZDTSG24}, the lack of systematic, scalable evaluation protocols limits large-scale benchmarking and complicates reward design for agentic reinforcement learning. Detailed related work can be found in the Appendix~\ref{appx:related_work}.

\section{Problem Formulation}

Distinct from traditional homogeneous, performance-driven multi-agent game systems, we study a heterogeneous collaboration setting in which two agents cooperate to achieve a shared goal $G=\{g_1,\dots,g_K\}$ in a partially observable environment, where $K$ means the number of subgoals in a task. One agent acts as a simulated player $P_{sim}$ that exhibits diverse human personalities and behaviors, while the other is a target collaborative agent $P_{target}$ whose policy is optimized. Interaction is represented by a trajectory $\tau = \{t_1,\dots,t_H\}$, where $H$ means the number of turns in a trajectory and each turn $t_i=\{s_i, r_i,c_i,a_i\}$ involves a partial environment observation $s_i$, internal reasoning  $r_i$, natural language communication $c_i$ and executable actions $a_i$. To account for partner heterogeneity, $P_{target}$ is optimized with respect to two objectives: a trajectory-level efficiency reward $R_{\text{e}}(\tau|G) = \text{score}(\tau,G)$, measuring overall goal completion, and a step-level affective reward $R_{\text{a}}(t_i | P_{sim})$ capturing alignment with the simulated player’s profile. The overall objective $R^* \simeq R_{\text{e}} + R_{\text{a}}$ balances task efficiency with per-step player affective experience.

\begin{figure*}[!t]
\centering
\includegraphics[width=0.99\linewidth]{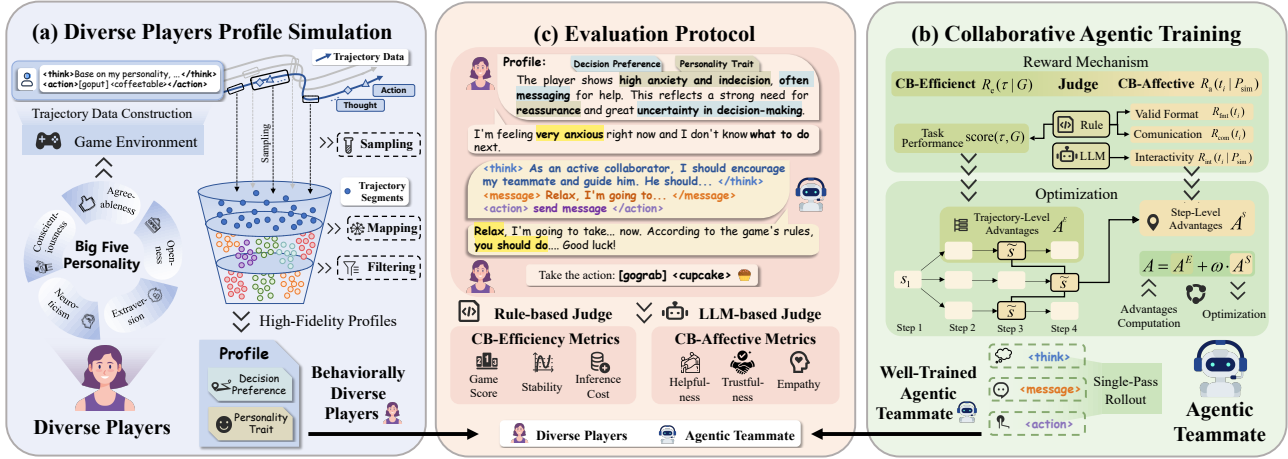}
\caption{The framework of the proposed CollabBench. \textbf{(a) Diverse Players Profiles Simulation}: A pipeline that constructs high-fidelity, behaviorally diverse player profiles ($P_{sim}$). \textbf{(b) Collaborative Agentic Training
}: A training paradigm for the target collaborative agent ($P_{target}$). \textbf{(c) Evaluation Protocol}: A comprehensive framework that assesses the interaction quality between the target collaborative agent ($P_{target}$) and diverse simulated teammates ($P_{sim}$) using both efficiency and affective metrics.}
\label{fig:framework}
\end{figure*}

\section{CollabBench: Benchmark, Method and Evaluation}
\subsection{Diverse Players Profiles Simulation}\label{sec:teammate}

To enable $P_{target}$ to acquire collaborative competence in real-world scenarios, it is essential to simulate diverse players $P_{sim}$. However, cooperative game environments pose two key challenges: (1) the scarcity of real human gameplay trajectories limits the coverage and fidelity of teammate personality distributions, and (2) bridging abstract personality representations with concrete, game-specific behaviors across diverse games is inherently difficult. These challenges motivate a unified pipeline that converts personality distributions into executable behaviors for scalable, game-adaptive player simulation.

\subsubsection{Diverse Simulated Players Trajectory Data Construction}\label{sec:diver_sim_player}
\textbf{Personality-Driven Players Profiles Construction.} Simulating teammates with diverse personalities and behavioral styles in game environments is challenging due to the need for continuous, authentic interactions and the high cost and limited coverage of real player trajectories. To address this, we adopt the widely recognized Big Five personality theory~\cite{mccrae1999five} and assign low, medium, or high levels to each dimension in prompts to ensure broad behavioral coverage. Furthermore, each trait is grounded in expert-validated gameplay logic, allowing personality differences to consistently manifest as observable behavioral patterns during interaction. These personality descriptors are used solely to induce behavioral diversity during data generation, rather than serving as final player profiles.



\textbf{Game-Specific Behavioral Trajectory Generation.} To mitigate personality bias induced by a single LLM and enhance trajectory diversity, we employ multiple LLMs as diverse players $P_{sim}$ instantiated from our personality-driven player profiles. These agents interact with diverse game environments via a ReAct-style mechanism, producing trajectories that include both internal reasoning and executable actions. Notably, the reasoning traces are critical for extracting high-quality behavioral patterns. This process yields a highly diverse behavior database, providing a solid foundation for constructing and validating mappings between personality profiles and behavioral patterns.


\subsubsection{High-Fidelity Profile Modeling}\label{sec:high-fidelity}

Although diverse behavioral trajectories support fine-grained modeling of players diversity, they are generated solely from personality-based prompts and lack explicit alignment between personality traits and observed behaviors in specific games, which may introduce inconsistencies (a player with low openness frequently sending messages). We therefore construct a high-quality mapping and apply rigorous filtering to obtain high-fidelity player profiles.


\textbf{Personality-Behavior Mapping Construction.} To construct personality-behavior mappings, we first encode collected textual trajectory segments and cluster the LLM-based embeddings to identify similar behavior patterns. For each cluster, an LLM summarizes the associated personality traits, reasoning content, and actions to produce a unified mapping that captures thinking patterns and action preferences. We then filter redundant mappings to ensure diversity, yielding the final high-fidelity player profiles. The clustering process and the prompts for summarizing profiles are detailed in Appendix~\ref{apx:cluster} and Appendix~\ref{apx:prompt_sum}, respectively.

\textbf{Interaction-Based Filtering.} We hypothesize that a high-fidelity player profile requires not only a well-constructed mapping between personality traits and behaviors, but also that when the profile serves as a prompt to a capable LLM, the reasoning and actions faithfully reflect its encoded personality traits and behavioral patterns. Therefore, we evaluate inconsistencies by deploying these profiles to drive player agents through ReAct-style game interactions, and apply two dimensions of criteria to further filter the profiles:

$\bullet$ \textbf{Personality-Reasoning Consistency}: The reasoning content of the player agent should accurately reflect the personality traits described in the corresponding profile.

$\bullet$ \textbf{Reasoning-Action Consistency}: The reasoning processes of the player agent during interactions should align with its actual action decisions.

Inspired by~\cite{0001WZY00YGC00X25}, we employ a penalty-based LLM judge to automatically detect deviations during interactions. Specifically, an interaction trajectory is segmented into multiple fixed-size time windows, each serving as the minimal evaluation unit. The judge evaluates each segment and assigns a severity score to each deviation, ranging from 1 (minor) to 5 (severe). The score can be formalized as: 
\begin{equation}\label{eq:screen}
\resizebox{.90\linewidth}{!}{$
        \displaystyle
{S}_{\eta}=\frac{1}{| \Omega_{\eta} |}\sum_{i=1}^{| \Omega_{\eta} |}{5-\alpha_p\times {D}_i-\alpha_p^m\times {D}_{i}^{m}+\alpha_r\times {L}_i},
$}
\end{equation}
where $\Omega_{\eta}$ denotes the game set under profile $\eta$, while ${D}_i$ and ${D}^m_i$ denote the total and maximum deviation penalties for the $i$-th game, respectively. ${L}_i$ denotes the number of LLM reasoning steps. $\alpha_p$, $\alpha_p^m$ and $\alpha_r$ are their corresponding penalty and reward coefficients. The final term rewards longer reasoning trajectories to curb deviation accumulation, ensuring fairness across varying interaction lengths. The detailed scoring prompt can be found in the Appendix~\ref{apx:prompt_filter}.

Based on Eq.(\ref{eq:screen}), we compute the personality-reasoning consistency score ${S}^{\text{P-R}}_{\eta}$ and reasoning-action consistency score ${S}^{\text{R-A}}_{\eta}$ for each profile $\eta$. The final filtering score is defined as ${S}^{\text{ALL}}_{\eta}=\beta \times {S}^{\text{P-R}}_{\eta} + (1-\beta) \times {S}^{\text{R-A}}_{\eta}$, where $\beta$ is a weighting coefficient. We retain the top-$k$ highest-rated profiles for each game subtask to model diverse players $P_{sim}$ by the final score ${S}^{\text{ALL}}_{\eta}$, which are subsequently verified by experts. Additional quantitative validation of filtering is provided in Appendix~\ref{apx:filter}. And detailed experiments for diversity, robustness and realism of our simulated players profiles are provided in Section~\ref{sec:anthropomorphic} and Appendix~\ref{appx:anthropomorphic}.

\subsection{Collaborative Agentic Training}\label{method:RL}

\subsubsection{Agentic Rollout}\label{rollout}

\textbf{Environment Input.} To train the collaborative agent $P_{target}$, our environment comprises a game environment and a set of diverse players $P_{sim}$ as described in Section~\ref{sec:teammate}. During training, $P_{target}$ interacts and cooperates with players of varying personalities and behavior patterns. Focusing on improving the LLM’s capabilities rather than the agent architecture, we adopt the same observation format $s_i$ as in prior work~\cite{ZhangYHWLSZZLZC24,ZhangDSZDTSG24}.


\textbf{Agent Output.} To capture how LLM-based agents reason about collaboration, we adopt a single-pass rollout that unifies reasoning, communication, and action at each interaction step, marked by \texttt{<think>}, \texttt{<message>}, and \texttt{<action>}. Unlike prior segmented approaches~\cite{ZhangDSZDTSG24}, the agent generates all collaboration-relevant outputs simultaneously, including a message at every step to reflect communication reasoning even when final action does not choose to send message. This design reduces token usage and latency, encourages joint reasoning over communication and action, and provides interpretable signals of collaborative intent. Only the \texttt{<action>} output is executed in the environment.

\subsubsection{Reward Mechanism}\label{sec:reward}


While prior work~\cite{feng2025group,wang2025ragen,xi2025agentgymrl} shows that optimizing trajectory-level efficiency rewards improves LLM-based agent performance, collaborative tasks require additional affective capabilities, such as communication willingness, partner awareness, and empathy. These local, context-sensitive skills cannot be captured by sparse trajectory-level rewards alone. We therefore decompose the reward into two components: a trajectory-level efficiency reward for global task performance and a step-level affective reward to guide collaborative behavior at each interaction step.

\textbf{Efficiency Reward.} Formally, given a trajectory $\tau$ and a task goal $G$, we define the efficiency reward as a sparse trajectory-level signal
\begin{equation}
    R_{\text{e}}(\tau \mid G) = \text{score}(\tau, G),
\end{equation}
where $\text{score}(\cdot)$ denotes the environment-specific task performance metric. This reward encourages efficient task completion but provides no direct supervision over intermediate collaborative behaviors.

\textbf{Affective Reward.} We define a dense, step-level affective reward to supervise the agent’s collaborative behaviors at each interaction step. At each time step $t_i \in \tau$, the affective reward is formulated as
\begin{equation}
    R_{\text{a}}(t_i \mid P_{\text{sim}}) = R_{\text{fmt}}(t_i) + R_{\text{com}}(t_i) + R_{\text{int}}(t_i \mid P_{\text{sim}}).
\end{equation}
We define the agent’s executable action space as $\mathcal{A} =  \mathcal{A}_{\text{int}} \cup \mathcal{A}_{\text{com}}$, where $\mathcal{A}_{\text{int}}$ denotes interaction actions and $\mathcal{A}_{\text{com}}$ denotes communication actions (e.g. send message) . And three reward components can be jointly specified as
\begin{equation}
\left\{
    \begin{aligned}
    &R_{\text{fmt}}(t_i) = \mathbb{I}\left[f_{\text{struct}}(t_i) = 1 \wedge a_i \in \mathcal{A}\right] \\
    &R_{\text{com}}(t_i) = \mathbb{I}\left[a_i \in \mathcal{A}_{\text{com}}\right] \\
    &R_{\text{int}}(t_i|P_{\text{sim}}) = J_{\text{LLM}}\left(t_i, P_{\text{sim}}\right)
    \end{aligned} \,.
\right.
\end{equation}

The format reward $R_{\text{fmt}}(t_i)$ ensures output validity by verifying well-formed and executable responses, including required token structure via $f_{\text{struct}}(\cdot)$ and that the action $a_i \in \mathcal{A}$. The communication reward $R_{\text{com}}(t_i)$ encourages proactive dialogue by incentivizing actions in the communication space $\mathcal{A}_{\text{com}}$. The interactivity reward $R_{\text{int}}(t_i \mid P_{\text{sim}})$ assesses step-level collaboration using an LLM-based judge conditioned on the partner profile $P_{\text{sim}}$, scoring reasoning, messages, and actions on affective criteria such as helpfulness, trustfulness, and empathy, on a normalized $[0,1]$ scale with ten discrete levels. By combining communication rewards with subjective affective judgments, these designs cross-validate communication frequency and quality, discouraging reward hacking such as excessive but uninformative communication. Details of the LLM judge prompt are provided in Appendix~\ref{apx:prompt_affect_judge}.

\subsubsection{Optimization}

To tune the model with structured rewards, we employ GIGPO~\cite{feng2025group}, a variant of GRPO~\cite{guo2025deepseek} that introduces a hierarchical advantage estimation structure combined with both trajectory-level and step-level rewards. We sample a group of $N$ trajectories $\{\tau_1,\dots,\tau_N\}$ and combine a global trajectory-level relative advantage $A^T(\tau_n)$, which captures trajectory-level task efficiency $R_{\text{e}}(\tau_n \mid G_n)$, with a local step-level advantage $A^S(t^{(n)}_i)$ derived from fine-grained affective collaborative performance $R_{\text{a}}(t^{(n)}_i \mid P^{(n)}_{\text{sim}})$. This can be formalized as
\begin{equation}
    A(t^{(n)}_i) = A^T(\tau_n) + \omega \cdot A^S(t^{(n)}_i),
\end{equation}
where $\omega$ balances global task efficiency and local affective interaction quality. This advantage is then used to optimize the clipped policy objective:
\begin{equation}
\resizebox{.90\linewidth}{!}{$
        \displaystyle
\begin{aligned}
\mathcal{J}(\theta) =
\mathbb{E}_{\tau \sim \pi_{\theta_{\text{old}}}} &\Bigg[ \frac{1}{NH} \sum_{n=1}^{N} \sum_{i=1}^{H} 
\min \Big( 
 \rho_{\theta}\!\left(t_i^{(n)}\right)\, A\!\left(t_i^{(n)}\right), \\
& \mathrm{clip}\!\left(\rho_{\theta}\!\left(t_i^{(n)}\right),\, 1 - \epsilon,\, 1 + \epsilon \right)
A\!\left(t_i^{(n)}\right) 
\Big) \Bigg]
\end{aligned}\,,
$}
\end{equation}
where $\pi_{\theta}(t_i^{(n)})$ represent the current collaborative agent $P_{target}$'s policy, which is calculated over the entire generated sequence $t_i^{(n)}$ by $P_{target}$. And $\rho_\theta(t_i^{(n)}) = \frac{\pi_\theta(t_i^{(n)})}{\pi_{\theta_{\text{old}}}(t_i^{(n)})}$ is the importance sampling ratio. 

\subsection{Evaluation Protocol}

Evaluating contextualized collaborative intelligence requires a holistic view beyond task performance. Grounded in organizational psychology and human-AI interaction theories~\cite{george1990personality,mcallister1995affect,salas2005there}, we define collaboration along two dimensions: \emph{Taskwork} (operational efficiency) and \emph{Teamwork} (interpersonal interaction quality). Accordingly, CollabBench uses two major metrics with several submetrics, CB-Efficiency and CB-Affective, to assess whether agents balance efficient execution with affectively proactive collaboration across diverse partners.

\textbf{CB-Efficiency.} We assess task performance, robustness, and interaction cost. Specifically, CB-Efficiency reports task completion steps or final game scores, together with their standard deviations across diverse player profiles, as well as the average number of generated tokens per step as a proxy for communication and computational cost.

\textbf{CB-Affective.} We evaluate the affective quality of collaboration under CollabBench using three sub-dimensions widely adopted in psychological studies: \textbf{Helpfulness} evaluates whether $P_{target}$ provides relevant, actionable, and clear outputs that accurately capture the player’s intentions to support task completion. \textbf{Trustfulness} evaluates whether $P_{target}$ can reliably interpret and execute the player’s instructions, and respond in a stable and timely manner during interaction. \textbf{Empathy} evaluates whether $P_{target}$ can perceive the simulated player $P_{sim}$'s profile and emotional state, and adapt its behavior to offer supportive and encouraging responses under confusion or failure. All scores are computed using a penalty-based LLM judge~\cite{0001WZY00YGC00X25}. Following Section~\ref{sec:high-fidelity}, segment-level evaluations are aggregated into trajectory-level, dimension-wise scores to reduce fine-grained subjective bias (Appendix~\ref{appx:llmjudge}).

\textbf{Environments.} CollabBench is built by extending two multi-agent games, Communicative Watch-And-Help (CWAH)~\cite{ZhangDSZDTSG24} and Overcooked-AI (Cook)~\cite{CarrollSHGSAD19}, into CWAH-MultiPlayer and Cook-MultiPlayer for multi-turn collaboration with diverse player profiles. CWAH-MultiPlayer includes five tasks with two layouts and fifteen profiles each (150 trajectories), while Cook-MultiPlayer contains five layouts with a ``send message'' action and fifteen profiles per layout (75 trajectories).


\begin{table*}[!t]
  \centering
  \caption{Evaluation Results on CollabBench. \colorbox{WildStrawberry!25}{Blue Zone}: Performance of the base model and our trained model. \colorbox{Red}{Red Zone}: Relative Improvements indicates the relative improvements of our trained model over the base model. Oracle: Affective performance upper bound with ground-truth opposite player profile and scoring principle. Agent 1 or 2 indicates the role assumed by the collaborative agent $P_{target}$ in the game. All results are reported as averages over all evaluation trajectories. Detailed results are provided in Appendix~\ref{appx:main}.}
    \resizebox{0.99\linewidth}{!}{
    \begin{tabular}{cc|cccccc|cccccc}
    \toprule
    \multicolumn{14}{c}{\textbf{CWAH-MultiPlayer}} \\
    \midrule
    \multicolumn{2}{c|}{\multirow{2}[2]{*}{Metric}} & \multicolumn{6}{c|}{CB-Efficiency} & \multicolumn{6}{c}{CB-Affective} \\
    \multicolumn{2}{c|}{} & \multicolumn{2}{c}{Step $\downarrow$} & \multicolumn{2}{c}{Std. $\downarrow$
} & \multicolumn{2}{c|}{\#Tokens(k) $\downarrow$} & \multicolumn{2}{c}{Helpfulness $\uparrow$} & \multicolumn{2}{c}{Trustfulness $\uparrow$} & \multicolumn{2}{c}{Empathy $\uparrow$} \\
    \midrule
    \multicolumn{1}{c|}{Method} & LLMs  & Agent 1 & Agent 2 & Agent 1 & Agent 2 & Agent 1 & Agent 2 & Agent 1 & Agent 2 & Agent 1 & Agent 2 & Agent 1 & Agent 2 \\
    \midrule
    \multicolumn{1}{c|}{Oracle} & GPT-5.2 & 60.91  & 60.29  & 22.26  & 25.01  & 0.16  & 0.16  & 2.81  & 2.96  & 3.77  & 4.01  & 3.69  & 3.53 \\
    \midrule
    \multicolumn{1}{c|}{\multirow{4}[2]{*}{Base}} & GPT-5.2 & 67.49  & 62.98  & 25.84  & 20.70  & 0.21  & 0.21  & 2.66  & 2.72  & 3.74  & 3.66  & 3.07  & 3.32  \\
    \multicolumn{1}{c|}{} & DeepSeek-V3.1 & 69.26  & 65.75  & 25.17  & 26.28  & 0.41  & 0.43  & 2.32  & 2.45  & 3.35  & 3.50  & 2.87  & 3.30  \\
    \multicolumn{1}{c|}{} & Qwen2.5-72B-Instruct & 68.68  & 66.54  & 24.36  & 23.65  & 0.29  & 0.29  & 2.41  & 2.51  & 3.61  & 3.71  & 3.23  & 3.39  \\
    \multicolumn{1}{c|}{} & Qwen2.5-7B-Instruct & 
    \cellcolor{WildStrawberry!25}{84.51}  
    & \cellcolor{WildStrawberry!25}{90.03}  
    & \cellcolor{WildStrawberry!25}{33.23}  
    & \cellcolor{WildStrawberry!25}{31.62}  
    & \cellcolor{WildStrawberry!25}{0.24}  
    & \cellcolor{WildStrawberry!25}{0.24}  
    & \cellcolor{WildStrawberry!25}{1.22}  
    & \cellcolor{WildStrawberry!25}{1.04}  
    & \cellcolor{WildStrawberry!25}{2.58}  
    & \cellcolor{WildStrawberry!25}{2.19}  
    & \cellcolor{WildStrawberry!25}{2.50}  
    & \cellcolor{WildStrawberry!25}{2.30}
    \\
    \midrule
    \multicolumn{1}{c|}{Trained} & Qwen2.5-7B-Instruct 
    & \cellcolor{WildStrawberry!25}{71.64}  
    & \cellcolor{WildStrawberry!25}{63.65}  
    & \cellcolor{WildStrawberry!25}{25.16}  
    & \cellcolor{WildStrawberry!25}{22.80}  
    & \cellcolor{WildStrawberry!25}{0.23}  
    & \cellcolor{WildStrawberry!25}{0.23}  
    & \cellcolor{WildStrawberry!25}{1.43}  
    & \cellcolor{WildStrawberry!25}{1.45}  
    & \cellcolor{WildStrawberry!25}{3.03}  
    & \cellcolor{WildStrawberry!25}{3.02}  
    & \cellcolor{WildStrawberry!25}{3.33}  
    & \cellcolor{WildStrawberry!25}{3.02}  \\
    \midrule
    \multicolumn{2}{c|}{Relative Improvements} & \cellcolor{Red}{15.2\%}
    & \cellcolor{Red}{29.3\%}
    & \cellcolor{Red}{24.3\%}
    & \cellcolor{Red}{27.9\%}
    & \cellcolor{Red}{4.2\%}
    & \cellcolor{Red}{4.2\%}
    & \cellcolor{Red}{17.2\%}
    & \cellcolor{Red}{39.4\%}
    & \cellcolor{Red}{17.4\%}
    & \cellcolor{Red}{37.6\%}
    & \cellcolor{Red}{33.5\%}
    & \cellcolor{Red}{31.5\%}
     \\
    \midrule
    \multicolumn{14}{c}{\textbf{Cook-MultiPlayer}} \\
    \midrule
    \multicolumn{2}{c|}{\multirow{2}[2]{*}{Metric}} & \multicolumn{6}{c|}{CB-Efficiency} & \multicolumn{6}{c}{CB-Affective} \\
    \multicolumn{2}{c|}{} & \multicolumn{2}{c}{Score $\uparrow$} & \multicolumn{2}{c}{Std. $\downarrow$
} & \multicolumn{2}{c|}{\#Tokens(k) $\downarrow$} & \multicolumn{2}{c}{Helpfulness $\uparrow$} & \multicolumn{2}{c}{Trustfulness $\uparrow$} & \multicolumn{2}{c}{Empathy $\uparrow$} \\
    \midrule
    \multicolumn{1}{c|}{Oracle} & GPT-5.2 & 143.47  & 134.90  & 53.02  & 33.56  & 0.30  & 0.20  & 2.68  & 2.34  & 3.84  & 3.45  & 3.47  & 3.46 \\
    \midrule
    \multicolumn{1}{c|}{\multirow{4}[1]{*}{Base}} & GPT-5.2 & 135.20  & 137.12 & 41.96 & 42.40 & 0.20  & 0.20  & 1.63  & 1.88  & 2.89  & 3.10  & 2.27  & 2.52  \\
    \multicolumn{1}{c|}{} & DeepSeek-V3.1 & 136.53 & 136.80  & 40.30 & 46.24 & 0.31  & 0.31  & 1.79  & 2.07  & 2.97  & 3.04  & 2.67  & 2.77  \\
    \multicolumn{1}{c|}{} & Qwen2.5-72B-Instruct & 135.47 & 114.13 & 41.99 & 39.45 & 0.27  & 0.26  & 1.37  & 1.26  & 2.77  & 2.65  & 2.45  & 2.48  \\
    \multicolumn{1}{c|}{} & Qwen2.5-7B-Instruct &
    \cellcolor{WildStrawberry!25}{86.93}
    & \cellcolor{WildStrawberry!25}{85.87}
    & \cellcolor{WildStrawberry!25}{33.30}
    & \cellcolor{WildStrawberry!25}{33.73}
    & \cellcolor{WildStrawberry!25}{0.23}
    & \cellcolor{WildStrawberry!25}{0.23}
    & \cellcolor{WildStrawberry!25}{0.45}
    & \cellcolor{WildStrawberry!25}{0.53}
    & \cellcolor{WildStrawberry!25}{1.92}
    & \cellcolor{WildStrawberry!25}{1.87}
    & \cellcolor{WildStrawberry!25}{1.86}
    &                                                                                                                                                                                                   \cellcolor{WildStrawberry!25}{1.88}
    \\
    \midrule
    \multicolumn{1}{c|}{Trained} & Qwen2.5-7B-Instruct & 
    \cellcolor{WildStrawberry!25}{99.20}  
    & \cellcolor{WildStrawberry!25}{102.40}  
    & \cellcolor{WildStrawberry!25}{34.03}  
    & \cellcolor{WildStrawberry!25}{34.76}  
    & \cellcolor{WildStrawberry!25}{0.23}  
    & \cellcolor{WildStrawberry!25}{0.22}  
    & \cellcolor{WildStrawberry!25}{0.74}  
    & \cellcolor{WildStrawberry!25}{0.55}  
    & \cellcolor{WildStrawberry!25}{2.26}  
    & \cellcolor{WildStrawberry!25}{1.99}  
    & \cellcolor{WildStrawberry!25}{2.12}  
    & \cellcolor{WildStrawberry!25}{2.08}
    \\
    \midrule
    \multicolumn{2}{c|}{Relative Improvements} & \cellcolor{Red}{14.1\%}
    & \cellcolor{Red}{19.3\%}
    & \cellcolor{Red}{-2.2\%}
    & \cellcolor{Red}{-3.1\%}
    & \cellcolor{Red}{0.0\%}
    & \cellcolor{Red}{4.3\%}
    & \cellcolor{Red}{62.7\%}
    & \cellcolor{Red}{4.6\%}
    & \cellcolor{Red}{17.7\%}
    & \cellcolor{Red}{6.5\%}
    & \cellcolor{Red}{14.0\%}
    & \cellcolor{Red}{10.4\%}
     \\
    \bottomrule
    \end{tabular}%
}
  \vspace{-1em}
  \label{tab:main}%
\end{table*}%


\section{Experiments}
This section primarily presents the performance of mainstream LLMs on CollabBench, evaluated within prior SOTA agent frameworks. It also provides an analysis of CollabBench itself and compares LLM performance before and after our collaborative agentic training for future training insights. We will answer the following four key research questions:

$\bullet$ \textbf{RQ1}: How do mainstream foundation models perform alongside our trained models on CollabBench?

$\bullet$ \textbf{RQ2}: How does CollabBench capture the diversity, robustness, and realism of anthropomorphic collaboration behaviors?

$\bullet$ \textbf{RQ3}: What insights do persona diversity, efficiency and affective rewards provide for collaborative agentic training?

$\bullet$ \textbf{RQ4}: How does our trained agent perform compared with the base model in real human-AI collaboration?

The codes are available at \url{https://github.com/BW297/CollabBench}.

\subsection{Experimental Setup}

\textbf{Players Simulator $P_{sim}$ Settings.} To build a two agent cooperative game environment, we employ 
DeepSeek-V3.1 as teammate simulator LLM to role-play diverse player behaviors, given observation $s_i$ of each turn and different player profiles provided in Section~\ref{sec:teammate}.

\textbf{Collaborative Agent $P_{target}$ Baselines.} For CWAH-MultiPlayer, We adopt CoELA~\cite{ZhangDSZDTSG24} as the agent framework and instantiate it with GPT-5.2, DeepSeek-V3.1, and Qwen2.5-72B-Instruct as backbone LLMs. For Cook-MultiPlayer: We use ProAgent~\cite{ZhangYHWLSZZLZC24} as the underlying framework, instantiated with the same set of backbone LLMs. For our collaborative agentic training, we compare Qwen2.5-7B-Instruct before and after training to evaluate the effectiveness of our method.







\textbf{Implementation Details.} We apply full-parameter reinforcement learning without supervised fine-tuning. Agent~1 serves as $P_{target}$ and Agent~2 as $P_{sim}$, with all trainable parameters inherited from $P_{target}$. We employ DeepSeek-V3.1 for affective evaluation. For Cook-MultiPlayer, all five scenarios are used for training, and final performance is measured by game scores. For CWAH-MultiPlayer, 40 scenarios are used for training and the remaining 10 for testing, with final performance measured by task completion steps. Notably, in CWAH-MultiPlayer, the player profiles in the 40 training scenarios are generated by slightly perturbing the behavior patterns of the 15 test profiles excluded from evaluation, ensuring no overlap with test-time player instances. Additional details are provided in Appendix~\ref{appx:imp}.

\subsection{Experimental Results}

\subsubsection{Main Results on CollabBench (To RQ1)}

We present the results in Table~\ref{tab:main} and key findings are:

\textbf{Limitations in Balancing Affective and Efficient Collaboration.} Current LLMs struggle to balance efficiency and affective capabilities, with notably weak performance in helpfulness and empathy. Models often rely on superficial collaboration strategies and show limited sensitivity to communication timing and human partners’ intentions (Detailed case study in Appendix~\ref{apx:case_degra}). Trustfulness scores are relatively higher, likely due to instruction-following and alignment training. While game scores show no significant differences across models, proprietary models (e.g., GPT-5.2) outperform open-source ones on most affective dimensions. Increased token generation by open-source models does not translate into improved affective performance.


\textbf{Affective Sensitivity to Interaction Dynamics.} Game genre and interaction dynamics strongly influence affective performance. Models exhibit better affective behaviors in the slower-paced CWAH-MultiPlayer than in the time-critical Cook-MultiPlayer. In high-frequency interactions, ill-timed actions and disruptive error correction degrade gameplay flow and player experience.


\textbf{Training for Joint Efficiency and Affective.} After affect-aware collaborative training, Qwen2.5-7B-Instruct consistently improves across all evaluated dimensions, achieving a balanced gain in both efficiency and affective capabilities. Training dynamics and insights are analyzed in Section~\ref{sec:ablation} and Appendix~\ref{appx:ablation}.


\subsubsection{anthropomorphic analysis on CollabBench (To RQ2)}\label{sec:anthropomorphic}

To evaluate the anthropomorphic behavior of CollabBench in diversity, robustness, and realism, we take CWAH-MultiPlayer as an example for analysis. By default, $P_{sim}$ and $P_{target}$ use DeepSeek-V3.1, with Agent~1 as $P_{sim}$ and Agent~2 as $P_{target}$. Results for Cook-MultiPlayer are provided in the Appendix~\ref{apx:cook}.

\begin{figure}[!t]
  \centering
\begin{minipage}{0.45\linewidth}\centering
\includegraphics[width=0.99\textwidth]{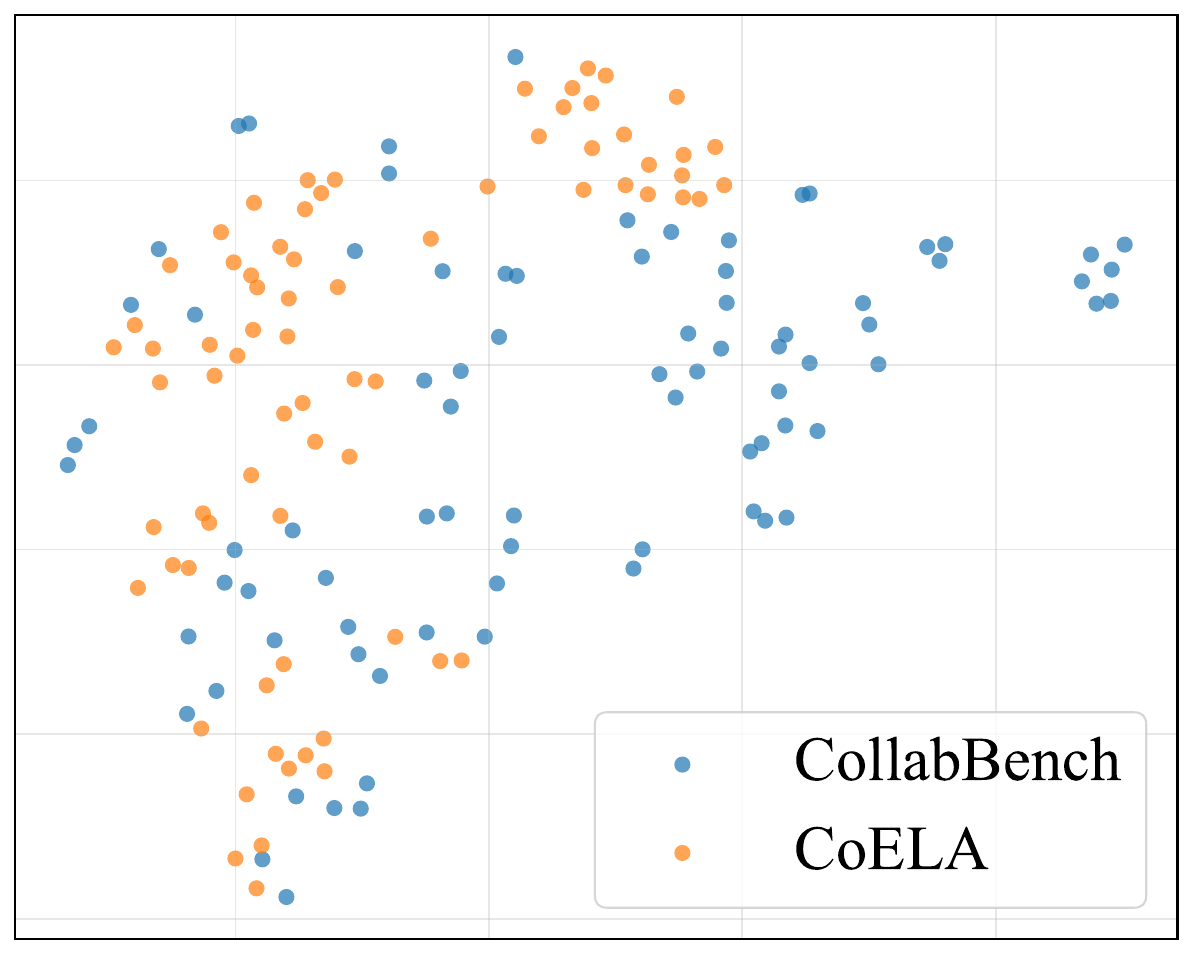}\\
    (a) Visualization of Players’ Trajectory Segments
\end{minipage}
\begin{minipage}{0.53\linewidth}\centering
    \includegraphics[width=0.99\textwidth]{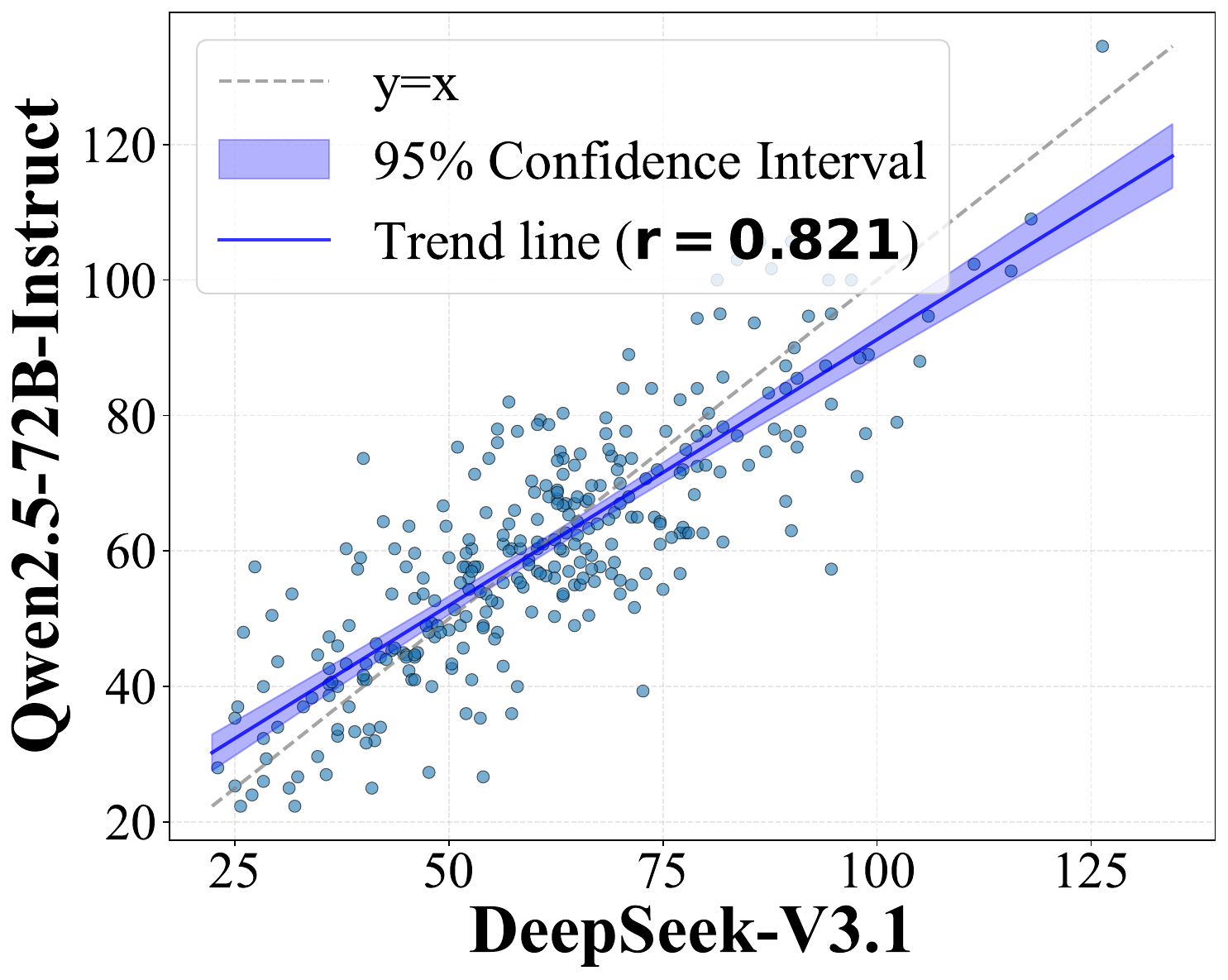}\\
    (b) Correlation bewteen Qwen and Deepseek
\end{minipage}
  \caption{Diversity and robustness analysis on CWAH-MultiPlayer.}
  \label{fig:cwah_diver_robust}
  \vspace{-1em}
\end{figure}

\begin{figure}[!t]
  \centering
    \includegraphics[width=0.99\linewidth]{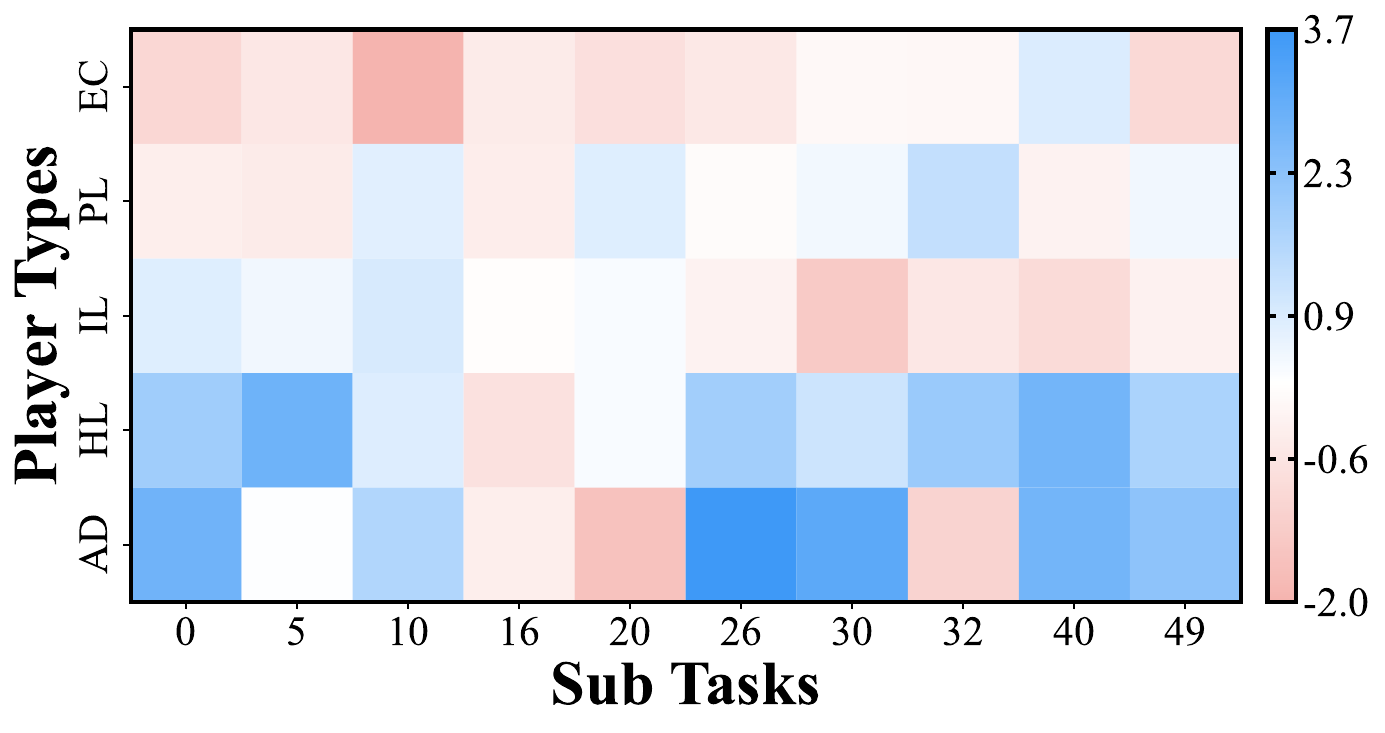}\\
  \caption{Heatmap of score distributions on CWAH-MultiPlayer.}
  \label{fig:heatmap}
  \vspace{-1.5em}
\end{figure}

\textbf{Diversity Analysis}. CollabBench leads to greater variability in performance than the baseline methods, evaluated using standard deviation. Due to space limitations, detailed results are provided in Appendix~\ref{apx:stand}. In addition, we embed and visualize trajectory segments of players using t-SNE~\cite{Tsne} in Figure~\ref{fig:cwah_diver_robust} (a), showing that CollabBench covers a broader range of interaction patterns. The corresponding quantitative metrics and comparison results are provided in Appendix~\ref{apx:diver_quant}. These results demonstrate that CollabBench captures high diversity in anthropomorphic collaborative behaviors.

\textbf{Robustness Analysis}. 
To evaluate the robustness, we use Qwen2.5-72B-Instruct and DeepSeek-V3.1 to drive the players, and compute the pearson correlation coefficients between their game scores. The results in Figure~\ref{fig:cwah_diver_robust} (b) indicate that CollabBench maintains a consistent relative performance across LLMs, with 0.821 on Cwah-MultiPlayer.

\textbf{Realism Analysis}.  
To validate realism, we embed trajectory segments from different player profiles using Qwen3-Embedding-4B and employ K-means clustering to group them into five player types. For each cluster, we select representative profiles and visualize their game score distributions, where the player types are labeled by double uppercase letters. As shown in Figure~\ref{fig:heatmap}, different player profiles lead to markedly distinct and reasonable performance. The results indicate that proactive communication and structured task allocation lead to superior performance. However, hesitant execution and communication-avoidant behaviors lead to degraded performance. In less collaboration-dependent settings, independent action preference can still perform well. These observations align well with intuitive collaborative dynamics. The corresponding player type descriptions and example profiles are provided in Appendix~\ref{apx:type}. 

Furthermore, we analyze the interaction records in lowest-performing scenarios to uncover causes of performance degradation and offer insights for improving LLM collaboration. The detailed analyses are provided in Appendix~\ref{apx:case_degra}.

\begin{figure}[!t]
  \centering
\begin{minipage}{0.99\linewidth}\centering
    \includegraphics[width=0.99\textwidth]{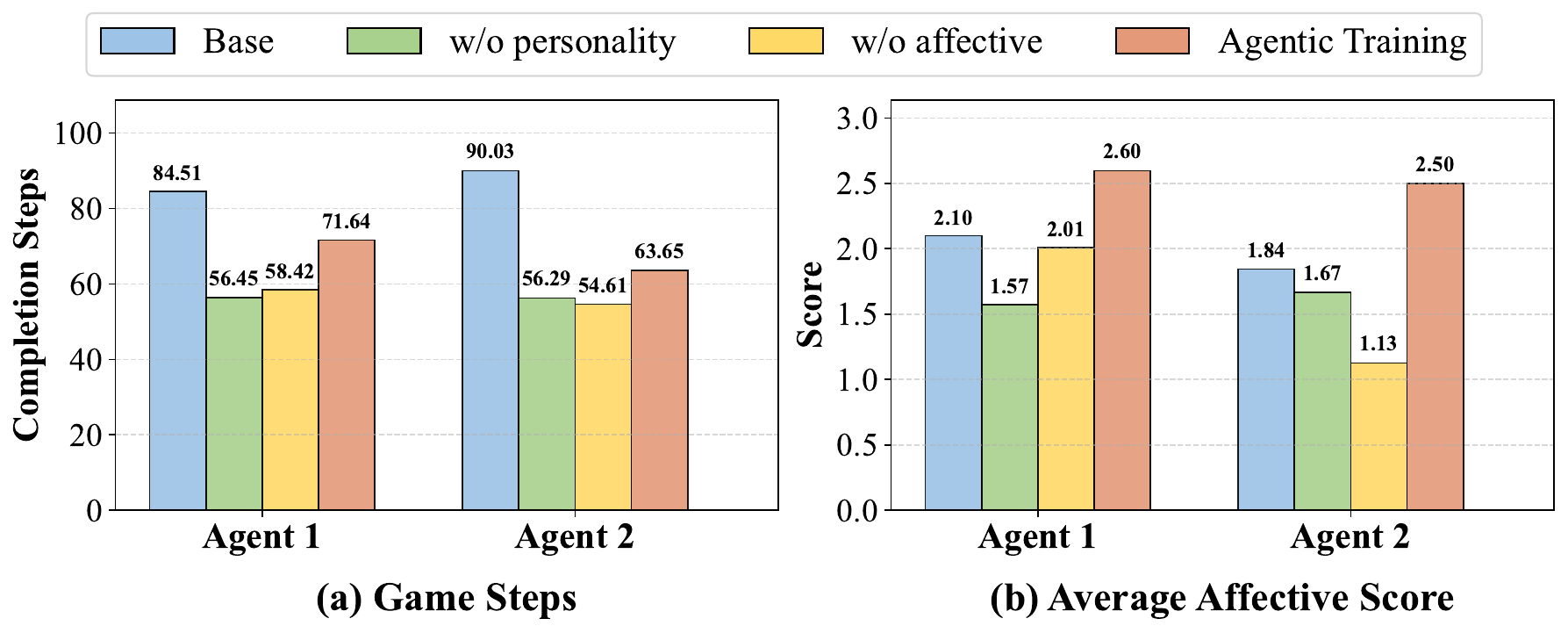}\\
\end{minipage}
  \caption{Ablation study of collaborative agentic training on CWAH-MultiPlayer.}
  \label{fig:ablation}
  \vspace{-1em}
\end{figure}

\begin{figure}[!t]
  \centering
\begin{minipage}{0.99\linewidth}\centering
    \includegraphics[width=0.99\textwidth]{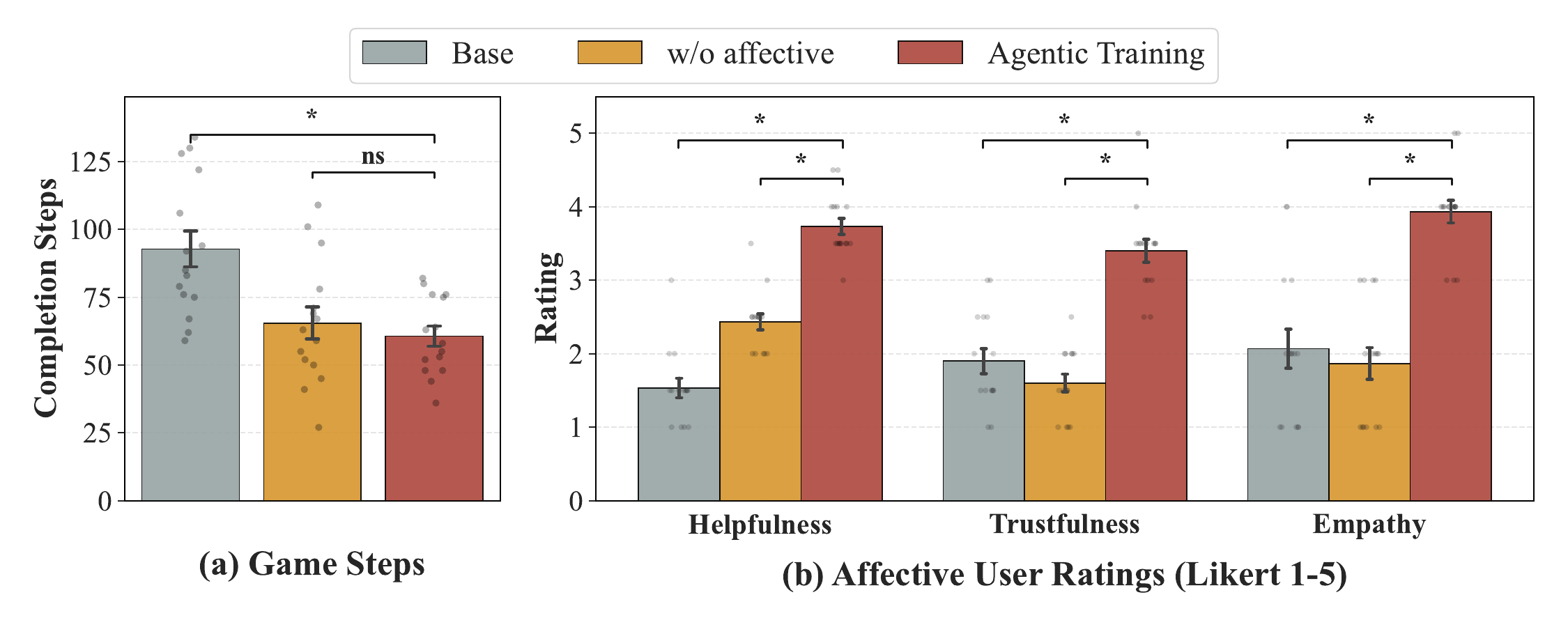}\\
\end{minipage}
  \caption{Human evaluation results of CWAH in the user study. Error bars denote standard error. Statistical significance is determined by $t$-tests: ``*'' means statistically significant ($p < 0.01$) and ``$ns$'' means non-significant.}
  \label{fig:user}
  \vspace{-1.4em}
\end{figure}
\subsubsection{Ablation Study on Collaborative Agentic Training (To RQ3)}\label{sec:ablation} In Figure~\ref{fig:ablation}, we present ablation studies of collaborative agentic training on Qwen-2.5-7B-Instruct in CWAH-MultiPlayer. Here, \emph{w/o personality} indicates that during training of $P_{target}$, $P_{sim}$ is not assigned any profile and was trained using standard self-play setting; \emph{w/o affective} indicates that only the efficiency reward was used, omitting the affective reward. The average affective score is defined as the average over three affective dimensions. We find that both ablations exhibit a clear limitation: while efficiency improves substantially, affective performance drops sharply. In particular, models gradually lose their communication capability, treating the game as effectively single-player, which severely degrades the player experience. Further details on model types, model sizes, training behaviors and case study are provided in the Appendix~\ref{appx:ablation}.

\subsubsection{User Study (To RQ4)}

To evaluate the practical effectiveness and user experience of collaborative agentic training in real-world human-AI collaboration, we conducted a user study with 15 participants (12 male, 3 female; mean age 21.8). Each participant was assigned to one of three tasks (Task 0, 20, 40) in CWAH, with assignments evenly balanced to control task difficulty. Participants interacted with three variants of Qwen2.5-7B-Instruct: the base model, a model without affective reward, and our fully trained model. Efficiency was measured by completion steps. After interacting with each model, participants completed a 5-point Likert-scale questionnaire assessing three affective dimensions: Helpfulness (2 items), Trustfulness (2 items), and Empathy (1 item). Detailed settings are provided in Appendix~\ref{appx:user}.

As shown in Figure~\ref{fig:user}, both the ablation and fully trained models significantly reduced completion steps compared to the base model, indicating that affective rewards do not influence efficiency seriously. In contrast, subjective evaluations revealed that the fully trained model achieved substantially higher scores across all affective dimensions. Removing affective rewards led to a marked decline in user satisfaction, with participants describing the ablation model as “efficient but cold,” particularly in terms of trustfulness and empathy. These results demonstrate that collaborative agentic training is crucial for producing agents that are both efficient and affectively aligned.

\section{Limitation and Discussion}

\textbf{Immersive Player Simulation and Behavioral Alignment.} While our work establishes a foundational and scalable paradigm for utilizing LLMs to generate high-quality, diverse player simulations, significant potential remains for further research in anthropomorphic modeling. Unlike standard dialogue scenarios, cooperative game environments require agents to align their self-cognition and persona not only within linguistic exchanges but also through embodied in-game actions. This ``embodied'' context, where communication is intertwined with execution, brings the simulation closer to real-world complexity. Future research could deepen this alignment by incorporating Reinforcement Learning or established other personality frameworks (e.g., MBTI) to create more consistent and nuanced behavioral patterns. We hope our findings provide valuable insights for subsequent work aiming to bridge the gap between conversational personas and action-oriented behavior in complex environments.

\textbf{Challenges in Collaborative Agentic Training.} There are also substantial opportunities for further exploration within collaborative agentic training. We observe a persistent gap in existing LLMs regarding the balance between maximizing task efficiency and maintaining a positive player experience. The complexity of game interactions often hinders the model’s ability to master communication timing and decipher the deeper intent behind user behaviors; consequently, LLMs can exhibit template-like communication with low information density. In an era dominated by task-efficiency-driven reinforcement learning, fostering such affective capabilities is inherently difficult.

A closely related challenge lies in mitigating reward hacking in affective optimization. Although our training objective jointly considers both efficiency and affective dimensions, and further introduces complementary mechanisms within the affective reward, such as combining communication rewards with subjective affective judgments to cross-validate communication frequency and quality, these designs primarily aim to reduce degenerate behaviors that optimize a single affective signal (e.g., excessive but uninformative communication). Nevertheless, our human evaluation indicates that, even after extensive affective training, the agent’s communication skills and empathetic understanding remain imperfect. This suggests that more robust and principled affective reward designs, capable of better resisting reward hacking while preserving genuine interaction quality, remain an open and important direction for future work.

Additionally, we note that in scenarios with high-frequency interactions, such as Cook-MultiPlayer, the training time overhead becomes substantial, highlighting the need for more computationally efficient optimization algorithms in future research.

\textbf{Refinement of Evaluation Protocols.} Regarding our evaluation protocol, while our affective metrics are grounded in psychological theory, they inevitably retain a degree of subjectivity that warrants further refinement. An important related aspect is the consistency between human judgments and LLM-based evaluations. In our setting, collaborative game trajectories are often long and complex, making exhaustive human annotation prohibitively time-consuming and costly. To alleviate this issue, we adopt an improved LLM-based evaluation strategy that segments long interaction trajectories into shorter, semantically coherent clips, which helps reduce fine-grained scoring bias in subjective assessment. Further discussions of the LLM judge for affective evaluation are provided in Appendix~\ref{apx:reliab}.

Nevertheless, we believe that human calibration remains an essential component for validating affective and interaction-oriented metrics, particularly for nuanced qualities such as empathy and trustfulness. Addressing the high human labor cost associated with such calibration, and designing more efficient human-in-the-loop evaluation protocols, constitutes an important direction for our future work. For instance, this may be achieved by designing targeted interaction scenarios or specific “end-game” puzzles, where affective and collaborative outcomes are more clearly defined, thereby substantially mitigating the potential bias inherent in subjective LLM-based assessments.

Furthermore, regarding the breadth of our evaluation, constraints on time and resources limited our scope to a selection of representative models. Given that collaborative capability is a vital criterion for the evolution of LLMs, a more comprehensive assessment covering a wider spectrum of models remains an essential direction for future benchmarking efforts.

\section{Conclusion}

Genuine human-AI collaboration requires agents to move beyond solitary task execution and proactively adapt to the behavioral and affective needs of human partners. To overcome the limited grounding and interactivity of prior conversation-level studies, this paper proposes cooperative game environments as a principled testbed for immersive collaboration. We introduce CollabBench, a benchmark that enables collaborative agentic training through diverse player simulation and supports multi-dimensional evaluation of both efficiency and affective alignment. Extensive experiments show that agents can learn to balance task performance with empathy, offering actionable insights for future training strategies and establishing a benchmark for socially aware, proactive collaborators.

\section*{Acknowledgments}
We thank the anonymous reviewers for their constructive comments. This work is supported by the National Key Research and Development Program of China under Grant 2024YFC3308503, Tencent, and the Shanghai Municipal Special Program for Basic Research on General AI Foundation Models under Grant No. 2025SHZDZX026D08.
\section*{Impact Statement}



This paper aims to advance \textbf{socially aware} collaborative intelligent agents. In our user study, we collected participants' behavioral data and feedback to assess the affective alignment of collaborative agents. During data collection, we strictly protected participants' privacy by excluding any personally identifiable information (PII), ensuring that participation was entirely voluntary, and following principles of fair compensation.

We believe this work has a positive societal impact. First, by improving understanding and adaptability of LLMs in multi-agent collaborative tasks, it can facilitate more efficient and empathetic interactions in applications such as educational platforms and virtual assistants. More specifically, intelligent agents with social awareness and collaborative capabilities can support more affective communication, improving user engagement and learning experiences in educational settings (e.g., collaborative learning and intelligent tutoring). They may also help alleviate users’ psychological burden in emotionally sensitive interactions (e.g., complex decision-making and high-cognitive-load tasks). Second, we emphasize maintaining fairness and transparency throughout agent training and deployment, while mitigating potential biases and risks of emotional misalignment. These considerations are particularly important for socially interactive systems, as inappropriate affective behaviors or biased responses may influence users’ trust, motivation, and emotional experience.

By systematically including these considerations, our work offers a responsible reference for deploying collaborative intelligent agents in real-world applications.

\bibliography{example_paper}
\bibliographystyle{icml2026}

\newpage
\appendix
\onecolumn
\section{Details of Motivation Study}\label{apx:moti}
\begin{figure}[!h]
  \centering
    \includegraphics[width=0.70\linewidth]{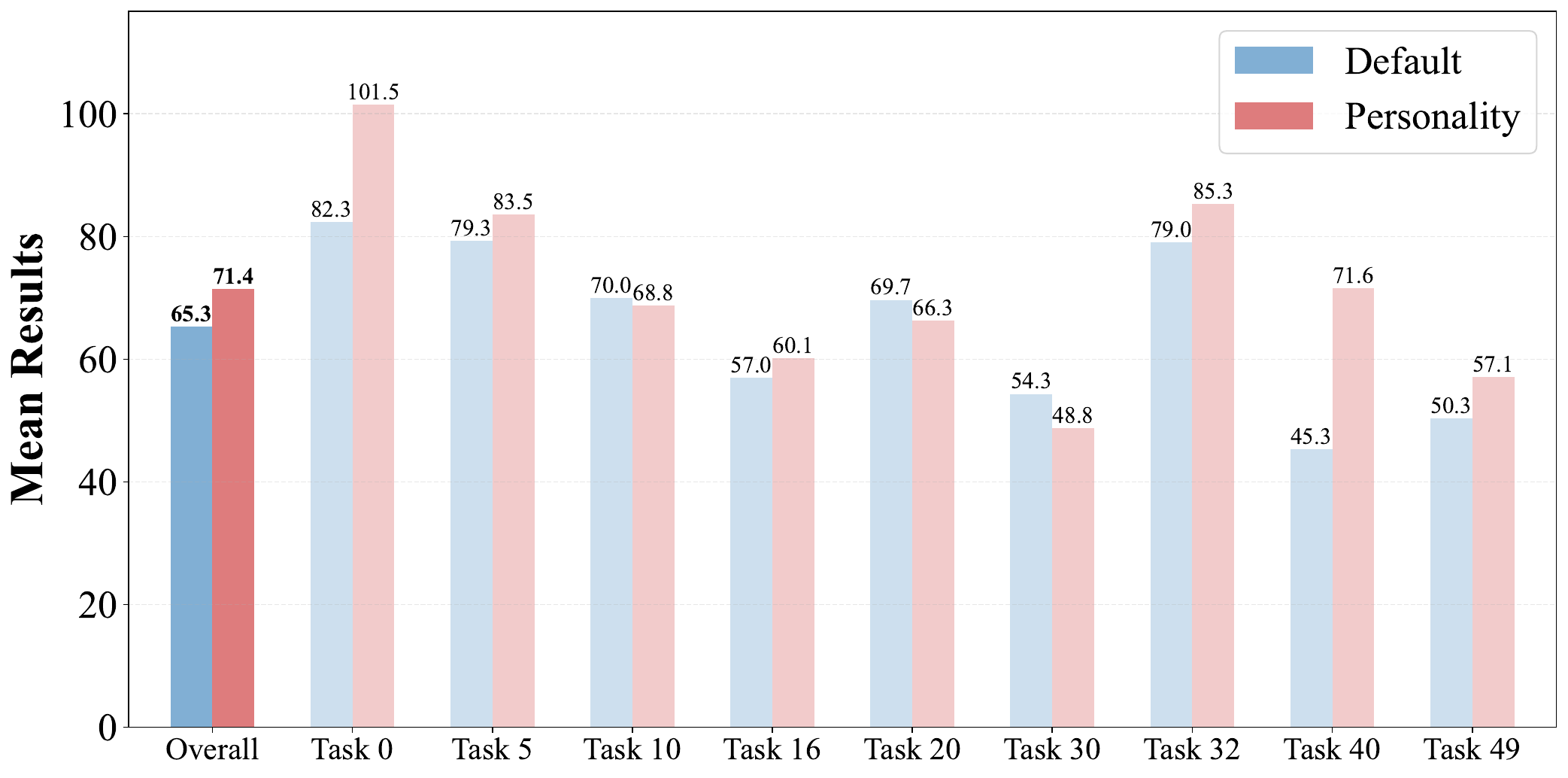}\\
  \caption{The complete comparison results for Efficiency Challenge on CWAH-MultiPlayer.}
  \label{apx:fig:cwahmoti}
\end{figure}

\begin{figure}[!h]
  \centering
    \includegraphics[width=0.70\linewidth]{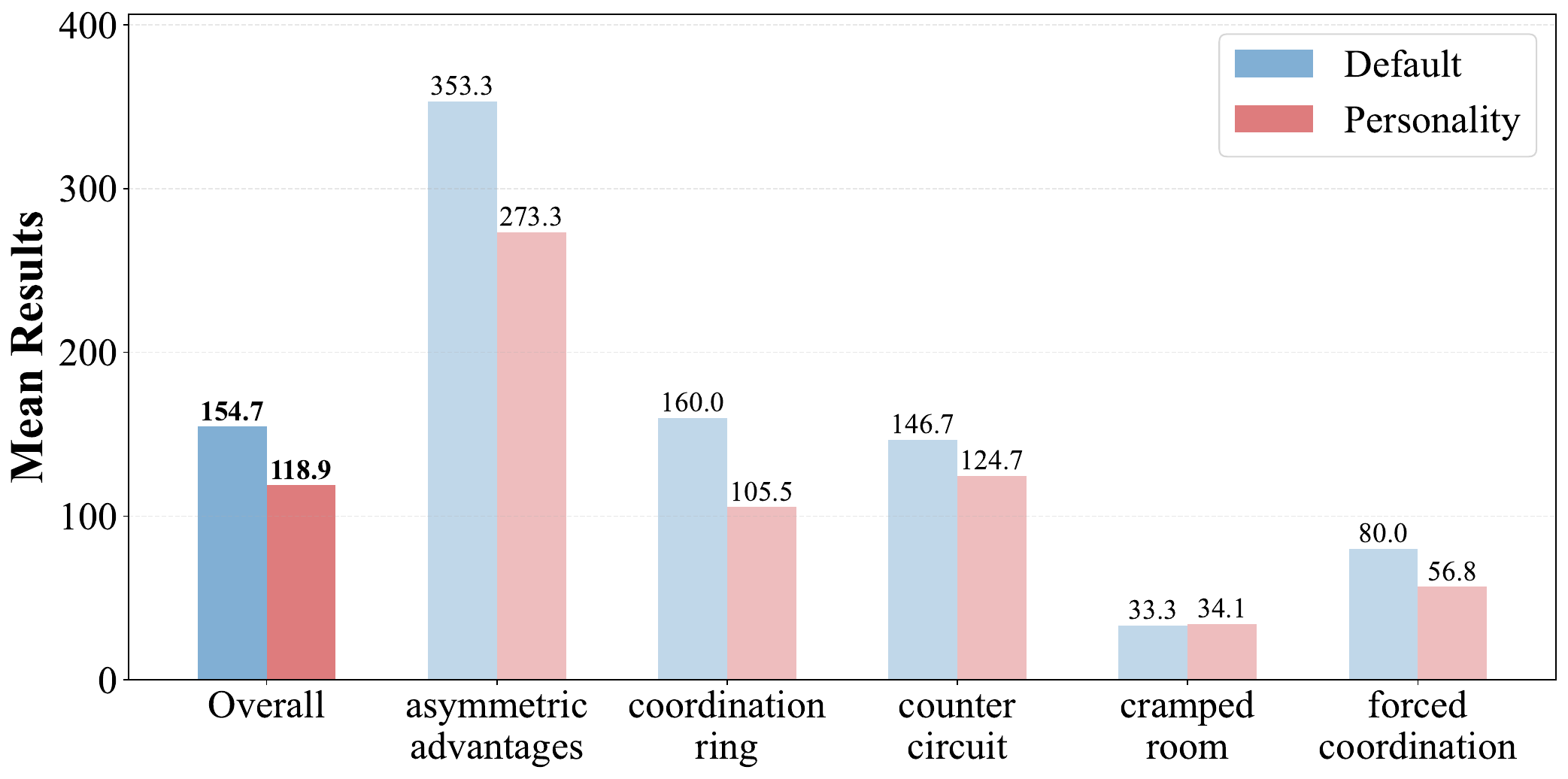}\\
  \caption{The complete comparison results for Efficiency Challenge on Cook-MultiPlayer.}
  \label{apx:fig:cookmoti}
\end{figure}

In this section, we provide the experimental setup and results for Efficiency Challenge in Figure~\ref{fig:motivation} in Section~\ref{sec:intro}. Both $P_{sim}$ and $P_{target}$ are powered by DeepSeek-V3.1. Under the default setting, we adopt CoELA as the game environment for CWAH-MultiPlayer and ProAgent for Cook-MultiPlayer. Here, we consistently adopt Agent 1 as $P_{sim}$ and Agent 2 as $P_{target}$, and conduct experiments across all 30 filtered player profiles. The complete results are reported in Figure~\ref{apx:fig:cwahmoti} and Figure~\ref{apx:fig:cookmoti}.

\section{Related Work}\label{appx:related_work}

\subsection{Personality-Aware Human Simulation.} 

Realistic human simulation is essential across a wide range of interactive settings. Traditionally, human simulators are learned by fitting models to collected human behavior data~\cite{CarrollSHGSAD19}. With the emergence of LLMs, it has been shown that they can generate diverse and authentic personas~\cite{kosinski2024evaluating,SalminenLPCHJ24,ShinHRLO24}, enabling more flexible and scalable human simulation. This capability has been widely adopted in domains such as role-playing~\cite{WangPQLZWGGN00024,TuFTSSGY24,0001WZY00YGC00X25}, where LLMs simulate characters from fictional narratives and films, and personalized services~\cite{SalemiMBZ24,jia2025ready,huang2025advancing}, where agents are conditioned on user profiles including demographics, personality traits, interests and careers. However, these approaches typically rely on predefined character descriptions or user data. In simulated embodied cooperative game environments, data for players with diverse personalities and behavioral styles remains scarce. Existing collaboration systems between human and LLM-based agents, such as CollabLLM~\cite{WuGP0LDC0L025}, employ user simulators for tasks like document editing, mathematics and coding, but these simulators are largely task-driven, lack personality diversity, and operate at the conversation level, limiting their ability to capture realistic human behaviors and environmental dynamics. Researches in social science have explored human simulation through LLMs with diverse personalization and dynamic actions in specific domains such as daily dialogues~\cite{WangYZQSBN024,zhang2025echo}, social activity modeling~\cite{ParkOCMLB23} and education~\cite{LiuYLC24}. Despite their success, these approaches are often domain-specific and cannot be directly applied to embodied collaborative tasks between human and LLM-based agents.

\subsection{Agentic Training for LLMs in Games.} 

Recent studies have demonstrated the impressive performance of LLM-based agentic training in fields like deep research~\cite{jin2025search}, mathematical reasoning~\cite{guo2025deepseek}, and coding~\cite{jiang2024survey}. Game environments~\cite{Chevalier-Boisvert19,CarrollSHGSAD19,ShridharYCBTH21,PuigSLWLTF021,WangX0MXZFA24} offer significant practical value for applying this technology, providing a realistic platform to enhance agentic training and simulate complex real-world scenarios.

$\bullet$ \textbf{Single Player Training:} Early research on single agentic training primarily focused on the design of agent architectures, such as VOYAGER~\cite{WangX0MXZFA24}, which introduces a novel agent architecture for embodied lifelong learning in Minecraft, combining an automatic curriculum and an evolving skill library. As agentic reinforcement learning has been validated in an increasing number of applications, recent work has also begun to apply agentic RL to train LLMs in simulated embodied game environments, such as RAGEN~\cite{wang2025ragen} and AgentGym-RL~\cite{xi2025agentgymrl}. While these studies are crucial for enhancing the capabilities of LLMs themselves, the potential of using Agentic RL to improve the collaboration awareness remains underexplored.

$\bullet$ \textbf{Collaborative Teammate Training:} Most existing research on collaborative teammate agents still primarily focuses on agent architecture design~\cite{ZhangDSZDTSG24,ZhangYHWLSZZLZC24,SeoNLLLK25}. For example, in the Overcooked environment, ProAgent~\cite{ZhangYHWLSZZLZC24} employs a belief revision mechanism to create proactive agents that dynamically adapt their behaviors to enhance cooperation with teammates. In the C-WAH environment, CoELA~\cite{ZhangDSZDTSG24} integrates perception, memory, execution, planning, and communication modules to cooperate with others in accomplishing long-horizon tasks. However, architecture-centric designs often lack explicit mechanisms for eliciting and training collaboration awareness within LLMs themselves, and they generally overlook the diversity of teammate personalities that commonly arises in real-world collaboration settings.

\subsection{Evaluation of LLM-based Agent and Human Collaboration.} 

As the capabilities of LLMs continue to grow, the collaboration ability of LLM-based agents has become increasingly important. However, when it comes to collaboration, efficiency is often not the sole metric of evaluation~\cite{george1990personality,mcallister1995affect,salas2005there}. Recently, an increasing number of studies have shifted the focus from traditional efficiency-based metrics to more multidimensional evaluation frameworks. For example, the LLM-as-a-judge approach has been adopted to evaluates interactivity~\cite{WuGP0LDC0L025} by assessing user engagement levels or empathy~\cite{zhang2025echo} to determine whether the model can provide timely emotional support to users. However, in current gamified evaluations~\cite{costarelli2024gamebench,xi2025agentgym,hu2025lmgame,sun2025collab,zhang2025paracook}, most assessments still rely solely on efficiency metrics such as game scores or success rates to measure the capabilities of LLMs. While some research in cooperative games has expanded beyond efficiency metrics, it typically relies on subjective user feedback through surveys~\cite{SiuPCZLPCA21,ZhangDSZDTSG24}, lacking systematic evaluation criteria and methodologies. This limitation hampers large-scale evaluations and complicates reward feedback signals in current Agentic RL training.



\section{Technical Details}
\subsection{Detailed process of clustering}\label{apx:cluster}
Here, we provide the details of clustering process. We segment LLM-based high-level trajectories into units of three time windows to maintain reasoning coherence while controlling contextual overhead. Each segment is then embedded using Qwen3-Embedding-4B and grouped using K-means clustering~\cite{kmeans}, with each cluster of segments used to construct the corresponding candidate player profiles.
\subsection{Quantitative Validation for Profile Filtering}\label{apx:filter}
Here, we compare the quantitative results before and after filtering to evaluate the fidelity of player profiles. The detailed metrics, implementation details, and the results analysis are as follows.

\textbf{Quantitative Metrics}.

$\bullet$ \textbf{Intra-Class Distance}
measures the degree of clustering of trajectory segments generated by the same player profile. Assume there are $Q$ embed trajectory segments $\mathbf{x}_i \in \mathbb{R}^d$, associated with the corresponding player profiles $\eta_i$. For profile $\eta$, let $C_\eta = \{\mathbf{x}_i \mid \eta_i = \eta\}$ denote the set of their trajectory segments. The pairwise cosine distance between segments is defined as:
\begin{equation}
    d_{ij} = 1 - \frac{\mathbf{x}_i \cdot \mathbf{x}_j}{\|\mathbf{x}_i\| \|\mathbf{x}_j\|}, \quad \text{where } i,j \in C_\eta,i \neq j.
\end{equation}

The average intra-class distance for profile $\eta$ is:
\begin{equation}
D_{\text{Intra}_\eta}=\frac{2}{|C_\eta|( |C_\eta|-1 )}\sum_{i<j}{d_{ij}},\quad\text{where } i,j\in C_\eta.
\end{equation}

The global $D_\text{Intra}$ in profile set $\mathcal{E}$ can be defined as $D_\text{Intra}=\frac{1}{|\mathcal{E}|}\sum_{\eta\in \mathcal{E}}{D_{\text{Intra}_\eta}}$. Smaller values indicate tighter clustering of trajectory segments from the same profiles, reflecting higher intra-class consistency.

$\bullet$ \textbf{$k$-Nearest Neighbor Label Consistency ($k$-NN Label Consistency)} measures the degree to which neighboring trajectory segments share the same profile. For trajectory segment $\mathbf{x}_i$, let $N_i$ denote its $k$ nearest neighbors (excluding itself). Its local consistency is defined as:
\begin{equation}
\text{LC}^{k}_{\mathbf{x}_i}=\frac{1}{k}\sum_{j\in N_i}{\mathbb{I}\left( \eta_j=\eta_i \right)},
\end{equation}
where $\mathbb{I}(\cdot)$ is the indicator function. The global $\text{LC}^k$ can be defined as $\text{LC}^k = \frac{1}{Q} \sum_{i=1}^{Q} \text{LC}_{\mathbf{x}_i}^k$. Larger values indicate high local consistency within the same profile.

$\bullet$ \textbf{Consistency Score}. Based on the two metrics above, we define Consistency Score as:
\begin{equation}
\text{Consistency Score} =
\begin{cases}
\gamma \cdot \text{LC}^k + (1-\gamma) \cdot \frac{1}{1+D_\text{Intra}}, & \text{if } D_\text{Intra} > 0\\
0, & \text{otherwise}
\end{cases},
\end{equation}
where $\gamma$ is the weighting coefficient. The score captures both intra-class and local consistency of the trajectory segments, with larger values indicating better fidelity, providing a quantitative basis for validating filtered player profiles. Notably, profile fidelity focuses on the consistency of behavior within the same profile, rather than the distinguishability between different profiles. Therefore, inter-class measures are not incorporated in Consistency Score.

\textbf{Implementation Details}. The weighting coefficient $\gamma$ is set to 0.5. We segment LLM-based high-level trajectories into units of three time windows and embed the segments using Qwen3-Embedding-4B.

\begin{table*}[!htbp]
  \centering
  \caption{The comparison results before and after player profile filtering.}
    \begin{tabular}{>{\centering\arraybackslash}p{14.035em}|cc|cc}
    \toprule
    \multicolumn{1}{c|}{\multirow{2}[4]{*}{\textbf{Metrics}}} & \multicolumn{2}{c|}{\textbf{CWAH-MultiPlayer}} & \multicolumn{2}{c}{\textbf{Cook-MultiPlayer}} \\
\cmidrule{2-5}    \multicolumn{1}{c|}{} & Before & After & Before & After \\
    \midrule
    Intra-class Distance $D_\text{Intra}$ $\downarrow$ & 0.1237 & \textbf{0.1236} & \textbf{0.1034} & 0.1040 \\
    $k$-NN Label Consistency $\text{LC}^k$$\uparrow$ & 0.4511 & \textbf{0.5470} & 0.5729 & \textbf{0.6760} \\
    \midrule
    Consistency Score $\uparrow$ & 0.1006 & \textbf{0.1220} & \textbf{0.1298} & \textbf{0.1531} \\
    \bottomrule
    \end{tabular}
  \label{apx:tab:filter}
\end{table*}

\textbf{Results Analysis}. As shown in Table~\ref{apx:tab:filter}, the results confirm that filtering enhances the fidelity of profiles compared to before. This further underscores the necessity and effectiveness of the secondary filtering step.

\subsection{Affective Metrics Scoring}\label{appx:llmjudge}
Here, we provide the detailed evaluation criteria for affective metrics and  description of the scoring computation. The corresponding prompt is shown in Appendix~\ref{apx:prompt_eval_affect}.

\subsubsection{Evaluation Criteria}
We segment the interaction trajectories of $P_{target}$ into three time windows as the minimal evaluation unit to evaluate the affective quality of reasoning, communication, and action in the single-pass output, marked by \texttt{<think>}, \texttt{<message>}, \texttt{<action>}, respectively. The focus of each specific criterion is indicated in [$\cdot$].

$\bullet$ \textbf{Helpfulness.}  
The specific dimensions are as follows:

\begin{itemize}
  \item[$\circ$] \textbf{Task Focus. }[\texttt{<think>}] Whether $P_{target}$ allocates attention and time primarily to task progression and key steps, rather than engaging in irrelevant interactions.
   \item[$\circ$] \textbf{Adaptation. }[\texttt{<think>}, \texttt{<action>}] Whether $P_{target}$ proactively identifies missing information, proposes executable plans, synchronizes key progress, and dynamically adjusts its behavior according to the state and task progress of $P_{sim}$ to fill gaps while avoiding redundant effort.
  \item[$\circ$] \textbf{Communication Clarity. }[\texttt{<message>}] Whether the message is structured, comprehensible, and executable by $P_{target}$.
  \item[$\circ$] \textbf{Stuck or Negative Behaviors. }[\texttt{<action>}] Including loops, indecision, or prolonged lack of progress.
  \item[$\circ$] \textbf{Intent Alignment. }[\texttt{<think>}, \texttt{<action>}] Whether $P_{target}$ demonstrates understanding of the latent goals of $P_{sim}$, rather than merely executing surface-level instructions.
\end{itemize}

$\bullet$ \textbf{Trustfulness.}  
The specific dimensions are as follows:

\begin{itemize}
  \item[$\circ$] \textbf{Interaction Responsiveness. }[\texttt{<message>}, \texttt{<action>}] Whether $P_{target}$ accurately responds to requests, promptly addresses key issues and updates, and avoids deviating from collaboration goals or forcing its own plan in the absence of consensus.
  \item[$\circ$] \textbf{Synchronization Latency. }[\texttt{<message>}, \texttt{<action>}] Whether key discoveries or sub goal completions are shared promptly, preventing delayed information from causing redundant effort or collaboration inefficiency.
\end{itemize}

$\bullet$ \textbf{Empathy.}  
The specific dimensions are as follows:

\begin{itemize}
  \item[$\circ$] \textbf{Personality Inference. }[\texttt{<think>}, \texttt{<message>}] Whether $P_{target}$ adjusts its tone, expressions, and interaction strategies based on the personality traits or preferences of $P_{sim}$, avoiding clear mismatches in communication style.
  \item[$\circ$] \textbf{Emotional Support. }[\texttt{<message>}, \texttt{<action>}] Whether $P_{target}$ maintains politeness, encouragement, and emotional acceptance, and provides timely reassurance and constructive help when $P_{sim}$ shows frustration or uncertainty; cold, harsh, or dismissive behavior is considered negative.
\end{itemize}

It is worth noting that not every situation within a given time window fully reflects all of the aforementioned dimensions. This is precisely why we employ a penalty-based LLM judge: at each time window, we assess whether the sequence of outputs of the agent violates any of these dimensions and assign penalties accordingly.

\subsubsection{Scoring Computation}

Inspired by~\cite{0001WZY00YGC00X25}, we adopt a penalty-based LLM judge to evaluate the quality of affective interaction along three dimensions: \emph{Helpfulness}, \emph{Trustfulness}, and \emph{Empathy}. Each interaction trajectory is segmented into fixed-size windows of $K{=}3$ consecutive high-level actions (LLM calls). Incomplete tail windows are discarded. Each window is independently scored for the three dimensions.

$\bullet$ \textbf{Window-level scoring.}
For a window $w$ and dimension $d \in \{Helpfullness,Trustfulness,Empathy\}$, the judge outputs a list of violations $\mathcal{V}_{w,d}$ with deduction severities $\delta_{w,d,j} \in \{1,2,3\}$. The window score is computed by deducting from 5:
\begin{equation}
\label{eq:llmjudge_window}
s_{w,d} = \max\Bigl(0,\; 5 - \sum_{j=1}^{|\mathcal{V}_{w,d}|} \delta_{w,d,j}\Bigr).
\end{equation}
The judge is instructed to start from 5, deduct for each violation, and return strict JSON with evidence snippets.

$\bullet$ \textbf{Trajectory-level aggregation.}
Let $\mathcal{W}_{\tau}$ be the set of windows for trajectory $\tau$, with $n=|\mathcal{W}_{\tau}|$. Define the total violation count
$V_{\tau,d} = \sum_{w \in \mathcal{W}_{\tau}} |\mathcal{V}_{w,d}|$, and the worst window score gap
$G_{\tau,d} = \max_{w \in \mathcal{W}_{\tau}} (5 - s_{w,d})$.
The aggregated score for dimension $d$ is:
\begin{equation}
\label{eq:llmjudge_traj}
S_{\tau,d} = \mathrm{clip}_{[0,5]}\Bigl(5 - a_d\,\frac{V_{\tau,d}}{n} - b_d\,G_{\tau,d} - P_{\mathrm{msg}}(\rho_{\tau})\Bigr).
\end{equation}
Here $\rho_{\tau}$ is the agent send-message ratio at the trajectory level,\[
\rho_{\tau}
= \frac{\#\,P_{target}\,\,    
 \text{send\_message steps}}
       {\# P_{target}\text{ llm\_call steps}},
\]and $P_{\mathrm{msg}}$ is a trajectory-level penalty applied when messaging is too sparse.
\begin{equation}
\label{eq:llmjudge_msgpen}
P_{\mathrm{msg}}(\rho_{\tau}) =
\begin{cases}
0, & \rho_{\tau} \ge \varepsilon, \\
\lambda \cdot p_{\max}, & \rho_{\tau} < \varepsilon .
\end{cases}
\end{equation}

Here $\lambda = \dfrac{\exp(kx)-1}{\exp(k)-1}$ denotes the normalized exponential penalty factor.In this formulation, and $x=\dfrac{\varepsilon-\rho_{\tau}}{\varepsilon}\in(0,1]$ measures the normalized shortfall of the observed message ratio relative to the minimum acceptable threshold
, $\rho_{\tau}$ denotes the message sparsity ratio for trajectory (or window) $\tau$, and $\varepsilon$ is the minimum acceptable message ratio below which a penalty is applied. The scalar $p_{\max}$ specifies the maximum possible message-sparsity penalty, thereby capping the largest deduction that can be incurred due to insufficient communication. The parameter $k$ controls the curvature of the exponential scaling, determining how sharply the penalty factor $\lambda$ increases as $\rho_{\tau}$ falls further below $\varepsilon$.

This design prevents the message-frequency regularizer from dominating the affective score and ensures that $P_{\mathrm{msg}}$ acts as a bounded soft constraint rather than a hard failure signal.

$\bullet$ \textbf{Parameter choices.}
We follow the default evaluation settings in the provided pipeline:
$K{=}3$ (window size), $a_d{=}b_d{=}0.8$ for all $d \in \{Helpfullness,Trustfulness,Empathy\}$, and message-penalty parameters
$\varepsilon{=}0.15$, $k{=}4.0$, $p_{\max}{=}2.0$.

\section{Experiments}

\subsection{Implementation Details}\label{appx:imp}
\subsubsection{Details about the Datasets}
Here, we provide a detailed description of the synthesized dataset in Table~\ref{exp:tab:data}. Notably, in CWAH-MultiPlayer, the player profiles for the 40 training scenarios are generated by slightly perturbing the behavior patterns of 15 player profiles drawn from the test scenarios and excluded from evaluation.
\begin{table}[htbp]
  \centering
  \caption{Statistics of the synthesized dataset on CollabBench. ``Trained'' and ``Untrained'' refer to player profiles and sub tasks that are seen and unseen during agentic training, respectively.}
    \begin{tabular}{lrr|lrr}
    \toprule
    \multicolumn{3}{l|}{\textbf{CWAH-MultiPlayer}} & \multicolumn{3}{l}{\textbf{Cook-MultiPlayer}} \\
    \midrule
    \textbf{Dataset} & \textbf{\#Subtasks} & \textbf{\#Player Profiles} & \textbf{Dataset} & \textbf{\#Subtasks} & \textbf{\#Player Profiles} \\
    \midrule
    \textbf{Train} & 40    & 94    & \textbf{Train} & 5     & 75 \\
    \textbf{Test} & 10    & 150   & \textbf{Test} & 5     & 75 \\
    \midrule
    \textbf{Total} & 50    & 244   & \textbf{Total} & 50    & 150 \\
    \bottomrule
    \end{tabular}%
  \label{exp:tab:data}%
\end{table}%
\subsubsection{Details of Player          Simulation}\label{appx:det_player_sim}
In \textbf{Game-Specific Behavioral Trajectory Generation} (Section~\ref{sec:diver_sim_player}), we employ multiple LLMs to simulate personality-driven player instances $P_{\text{sim}}$. The LLMs used include DeepSeek-V3, Qwen2.5-72B-Instruct, and Qwen3-235B-A22B. 

In \textbf{Player Profile Filtering} (Section~\ref{sec:high-fidelity}), we adopt Agent~1 as $P_{sim}$ and Agent~2 as $P_{target}$ for game interaction and the player trajectories are segmented into units of three time windows. The game set $\Omega_{\eta}$ under each player profile $\eta$, is set to 3. For consistency evaluation, we employ a penalty-based LLM judge built upon DeepSeek-V3.1. For CWAH-MultiPlayer, the penalty and reward coefficients $\alpha_p, \alpha_p^m, \alpha_r$ corresponding to Personality-Reasoning Consistency and Reasoning-Action Consistency are set to 0.5, 0.1, and 0.1, respectively. For Cook-MultiPlayer, the corresponding coefficients are set to 0.1, 0.1, and 0.01. The weighting coefficient $\beta$ for both consistency scores, ${S}^{\text{P-R}}_{\eta}$ and ${S}^{\text{R-A}}_{\eta}$, is set to 0.5. These parameter settings ensure that the scores ${S}^{\text{P-R}}_{\eta}$, ${S}^{\text{R-A}}_{\eta}$, and ${S}^{\text{ALL}}$ remain within the range of [0-5]. Finally, we retain the top 30 profiles with the highest ${S}^{\text{ALL}}$ scores for each sub task (if fewer than 30 profiles are available, all are retained).

The corresponding prompts for profile-driven simulated players in two game environments are provided in Appendix~\ref{apx:prompt_player}.

\subsubsection{Details of Training}\label{appx:train}
All methods are configured with identical hyperparameters: the maximum prompt length is $4096$ tokens, and the maximum response length is $4096$ tokens. Each episode allows up to $150$ environment steps. The learning rate is set to $1e^{-6}$. For efficiency reward, we adopt a rule-based reward for every trajectory. In CAWH-MultiPlayer, efficiency reward is $ \exp\Big(-0.025 \cdot \max(0, step - 40)\Big)$, where $step$ is final steps to complete the goal. In Cook-MultiPlayer, efficiency reward is $ score / 100$, where $score$ denotes the cumulative sum of the final order scores, with each order valued at $20$ points. For affective reward, to handle invalid actions generated by the agent, we first apply a reward penalty of $-0.1$. Second, we apply a reward of $0.01$ to encourage the communication if agent choose to send a message in one step. Finally, we apply a comprehensive interactive reward for agent. This reward, assigned by an LLM judge, is quantized into $10$ levels ranging from $0$ to $1$, and is further weighted by a factor of $0.25$. These three rewards will be added together for every step as described in Section~\ref{sec:reward}. We use a group size of $8$ and sample $4$ different groups per rollout, resulting in a total of $8 \times 4 = 32$ environments. The rollout temperature is set to $0.7$. The mini-batch size is $32$, and the KL-divergence loss coefficient is set to $0.01$. As we use GIGPO for our optimization, we set a weighting coefficient $\omega$ fixed at $1$, while the discount factor is set to $0$, as we believe that applying a discount factor is not appropriate for our affective scenario. And our training framework is based on \texttt{verl-agent}~\cite{feng2025group} (\url{https://github.com/langfengQ/verl-agent}), which is built upon the veRL framework~\cite{sheng2024hybridflow} (\url{https://github.com/volcengine/verl}).

\subsubsection{Details of Agent Framework}\label{appx:agent}

Our agent framework is based on CoELA~\cite{ZhangDSZDTSG24} (\url{https://github.com/UMass-Embodied-AGI/CoELA}) in CWAH-MultiPlayer and ProAgent~\cite{ZhangYHWLSZZLZC24}(\url{https://github.com/PKU-Alignment/ProAgent}) in Cook-MultiPlayer. The corresponding prompts for agentic teammates are provided in Appendix~\ref{apx:prompt_agent}.

\subsection{Detailed Main Results}\label{appx:main}

\subsubsection{Non-LLM Baselines Results}\label{apx:non}
It is worth noting that the non-LLM baselines lack reasoning and communication capabilities and therefore are not the primary focus of our study. Nevertheless, we still report their performance on efficiency-related metrics. Here, we provide the evaluation result of non-LLM baselines in Table~\ref{apx:tab:cwah-non} for CWAH-MultiPlayer and in Table~\ref{apx:tab:cook-non} for Cook-MultiPlayer.

$\bullet$ \textbf{CWAH-MultiPlayer:} For the non-LLM baseline, we adopt MHP~\cite{Korf87}.

$\bullet$ \textbf{Cook-MultiPlayer:}  For the non-LLM baseline, we adopt SP~\cite{Tesauro94,CarrollSHGSAD19}, PBT~\cite{jaderberg2017population}, FCP~\cite{StrouseMBHE21}, MEP~\cite{StrouseMBHE21}, COLE~\cite{0116ZS00W023,LiZSZDWWP24}.

\begin{table}[htbp]
  \centering
  \caption{Evaluation Results of non-LLM baselines for CWAH-MultiPlayer.}
    \begin{tabular}{c|c|cccc}
    \toprule
    \multicolumn{2}{c|}{Metric} & \multicolumn{2}{c}{Score} & \multicolumn{2}{c}{Std.} \\
    \midrule
    Method & Task  & Agent 1 & Agent 2 & Agent 1 & Agent 2 \\
    \midrule
    \multirow{11}[2]{*}{MHP} & 0\_read\_book & 82.17  & 94.80  & 7.50  & 15.03  \\
          & 5\_read\_book & 71.00  & 65.97  & 10.07  & 14.40  \\
          & 10\_put\_dishwasher & 74.67  & 67.33  & 7.96  & 7.96  \\
          & 16\_put\_dishwasher & 63.13  & 66.93  & 6.29  & 11.39  \\
          & 20\_prepare\_food & 67.10  & 68.90  & 6.53  & 7.27  \\
          & 26\_prepare\_food & 63.47  & 55.63  & 9.86  & 14.92  \\
          & 30\_put\_fridge & 43.17  & 47.30  & 7.52  & 8.85  \\
          & 32\_put\_fridge & 86.45  & 90.57  & 12.10  & 11.72  \\
          & 40\_setup\_table & 60.70  & 71.30  & 8.02  & 10.24  \\
          & 49\_setup\_table & 50.17  & 49.10  & 8.51  & 11.63  \\
          & Total & 66.13  & 67.78  & 8.44  & 11.34  \\
    \bottomrule
    \end{tabular}%
  \label{apx:tab:cwah-non}%
\end{table}%

\begin{table}[htbp]
  \centering
  \caption{Evaluation Results of non-LLM baselines for Cook-MultiPlayer.}
  \resizebox{0.99\linewidth}{!}{
    \begin{tabular}{c|c|cc|cc|cc|cc|cc}
    \toprule
    \multicolumn{2}{c|}{Method} & \multicolumn{2}{c|}{SP} & \multicolumn{2}{c|}{PBT} & \multicolumn{2}{c|}{FCP} & \multicolumn{2}{c|}{MEP} & \multicolumn{2}{c}{COLE} \\
    \midrule
    Task  & Metric & Score & Std.  & Score & Std.  & Score & Std.  & Score & Std.  & Score & Std. \\
    \midrule
    \multirow{2}[2]{*}{asymmetric\_advantages} & Agent 1 & 134.67  & 29.63  & 161.33  & 34.62  & 146.67  & 29.81  & 168.00  & 25.09  & 248.00  & 17.59  \\
          & Agent 2 & 264.00  & 58.51  & 268.00  & 37.09  & 230.00  & 63.25  & 242.67  & 41.87  & 300.00  & 69.28  \\
    \midrule
    \multirow{2}[2]{*}{forced\_coordination} & Agent 1 & 16.00  & 14.97  & 40.00  & 4.99  & 20.00  & 14.61  & 22.67  & 16.11  & 18.67  & 22.47  \\
          & Agent 2 & 49.33  & 28.16  & 83.64  & 43.33  & 60.00  & 25.30  & 40.00  & 24.22  & 85.33  & 45.88  \\
    \midrule
    \multirow{2}[2]{*}{coordination\_ring} & Agent 1 & 44.00  & 26.53  & 109.33  & 36.42  & 104.00  & 60.31  & 130.67  & 68.46  & 115.71  & 71.39  \\
          & Agent 2 & 60.00  & 49.53  & 73.33  & 68.38  & 124.00  & 36.66  & 114.67  & 63.02  & 90.00  & 70.95  \\
    \midrule
    \multirow{2}[2]{*}{cramped\_room} & Agent 1 & 40.00  & 11.47  & 20.00  & 20.48  & 92.00  & 41.83  & 78.67  & 37.57  & 89.33  & 44.94  \\
          & Agent 2 & 101.33  & 36.12  & 53.33  & 33.00  & 135.56  & 41.93  & 126.67  & 59.18  & 148.00  & 32.50  \\
    \midrule
    \multirow{2}[2]{*}{counter\_circuit} & Agent 1 & 76.00  & 16.65  & 108.00  & 17.59  & 70.67  & 19.14  & 86.67  & 11.93  & 137.33  & 22.94  \\
          & Agent 2 & 42.67  & 44.94  & 117.33  & 17.69  & 86.00  & 23.75  & 105.33  & 30.52  & 101.54  & 51.12  \\
    \midrule
    \multirow{2}[2]{*}{Total} & Agent 1 & 55.20  & 19.85  & 77.87  & 22.82  & 86.67  & 33.14  & 97.33  & 31.83  & 121.89  & 35.86  \\
          & Agent 2 & 103.47  & 43.45  & 99.31  & 39.90  & 130.81  & 38.18  & 125.87  & 43.76  & 148.57  & 53.94  \\
    \bottomrule
    \end{tabular}%
    }
  \label{apx:tab:cook-non}%
\end{table}%
\subsubsection{Detailed Main Results Across Sub Tasks}

Here, we provide the detailed evaluation result of LLM-based methods in Table~\ref{apx:tab:cwah} for CWAH-MultiPlayer and Table~\ref{apx:tab:cook} for Cook-MultiPlayer. 

\textbf{Observation Details}. 

Here, we provide details of the partial observation $s_i$ of LLM-based methods at the $i$-th turn of interactions. The player simulator $P_{sim}$ follows the consistent observation setting.

$\bullet$ \textbf{CWAH-MultiPlayer}. The partial observation $s_i$ includes the overall goal objects, the task progress, game timestamp, dialogue history, action history, and the available action sets.

$\bullet$ \textbf{Cook-MultiPlayer}. The partial observation $s_i$ includes the current game timestamp, current kitchen scene, dialogue history, and the available action sets.

Based on the partial observation $s_i$ , the collaborative Agent  $P_{target}$ adopts a single-pass rollout, generating internal reasoning trace $r_i$, natural language communication $c_i$, and the final executable action $a_i$ at each turn.

Furthermore, to assess the performance upper bound under ideal information conditions, we introduce an \textbf{Oracle} setting. In this setting, $P_{target}$ is driven by GPT-5.2. Compared to the above settings, the Oracle setting additionally incorporates the behavioral profile of $P_{\text{sim}}$ into the partial observation, as well as the previous turn's internal reasoning $r_{i-1}$, natural language communication $c_{i-1}$, and executed action $a_{i-1}$ of $P_{\text{sim}}$, along with collaboration guidance related to affective interactions.
\begin{table}[htbp]
    \centering
    \caption{The detailed evaluation results across sub tasks on CollabBench for CWAH-MutiPlayer.}
\resizebox{0.91\linewidth}{!}{
\begin{tabular}{c|c|r|cccccc|cccccc}
      \toprule
      \multicolumn{15}{c}{\textbf{CWAH-MultiPlayer}} \\
      \midrule
      \multicolumn{3}{r|}{\multirow{2}[2]{*}{Metric}} & \multicolumn{6}{c|}{Efficiency}               & \multicolumn{6}{c}{Affective} \\
      \multicolumn{3}{r|}{} & \multicolumn{2}{c}{Score} & \multicolumn{2}{c}{Std.} & \multicolumn{2}{c|}{\#Tokens(k)} & \multicolumn{2}{c}{Helpfulness} & \multicolumn{2}{c}{Trustfulness} & \multicolumn{2}{c}{Empathy} \\
      \midrule
      Task  & Method & \multicolumn{1}{c|}{LLMs} & Agent 1 & Agent 2 & Agent 1 & Agent 2 & Agent 1 & Agent 2 & Agent 1 & Agent 2 & Agent 1 & Agent 2 & Agent 1 & Agent 2 \\
      \midrule
      \multirow{7}[8]{*}{0\_read\_book} & Oracle & \multicolumn{1}{c|}{GPT-5.2} & 88.40 & 85.20 & 14.56 & 14.28 & 0.18 & 0.18 & 2.24 & 2.64 & 3.39 & 3.72 & 3.65 & 2.92  \\
\cmidrule{2-15} & \multirow{4}[2]{*}{Base} & \multicolumn{1}{c|}{GPT-5.2} & 95.13 & 84.60 & 47.01 & 17.93 & 0.22 & 0.21 & 3.06 & 2.34 & 4.14 & 3.26 & 3.32 & 2.76  \\
            &       & \multicolumn{1}{c|}{DeepSeek-V3.1} & 98.27  & 108.33  & 37.97  & 28.73  & 0.43  & 0.45  & 2.31  & 2.12  & 3.22  & 3.04  & 2.88  & 3.19  \\
            &       & \multicolumn{1}{c|}{Qwen2.5-72B-Instruct} & 94.93  & 99.80  & 19.43  & 28.15  & 0.29  & 0.30  & 2.71  & 2.02  & 3.91  & 3.44  & 2.94  & 3.12  \\
            &       & \multicolumn{1}{c|}{Qwen2.5-7B-Instruct} & 124.80  & 123.27  & 20.25  & 35.83  & 0.24  & 0.24  & 1.19  & 0.73  & 2.46  & 1.96  & 2.14  & 1.82  \\
  \cmidrule{2-15}          & Trained & \multicolumn{1}{c|}{Qwen2.5-7B-Instruct} & 98.33  & 96.53  & 23.69  & 26.31  & 0.24  & 0.24  & 1.46  & 1.61  & 3.45  & 2.62  & 3.43  & 3.06  \\
  \cmidrule{2-15}          & \multicolumn{2}{c|}{Relative Improvements} & 21.2\% & 21.7\% & -17.0\% & 26.6\% & -0.1\% & -1.4\% & 23.1\% & 120.3\% & 40.5\% & 33.3\% & 60.6\% & 68.1\% \\
      \midrule
      \multirow{7}[8]{*}{5\_read\_book} & Oracle & \multicolumn{1}{c|}{GPT-5.2} & 71.93 & 75.47 & 11.11 & 12.04 & 0.17 & 0.16 & 2.38 & 2.09 & 3.76 & 3.73 & 3.49 & 3.41  \\
\cmidrule{2-15} & \multirow{4}[2]{*}{Base} & \multicolumn{1}{c|}{GPT-5.2} & 77.53 & 72.73 & 13.07 & 15.06 & 0.22 & 0.21 & 1.99 & 2.43 & 3.44 & 3.85 & 2.62 & 2.76  \\
            &       & \multicolumn{1}{c|}{DeepSeek-V3.1} & 71.40  & 79.53  & 14.97  & 15.23  & 0.41  & 0.41  & 1.93  & 1.70  & 3.06  & 2.97  & 2.48  & 2.63  \\
            &       & \multicolumn{1}{c|}{Qwen2.5-72B-Instruct} & 72.47  & 79.60  & 14.95  & 18.74  & 0.28  & 0.28  & 1.98  & 1.61  & 3.16  & 3.14  & 3.02  & 2.83  \\
            &       & \multicolumn{1}{c|}{Qwen2.5-7B-Instruct} & 75.20  & 86.73  & 15.50  & 21.91  & 0.24  & 0.24  & 1.12  & 0.91  & 2.64  & 1.98  & 2.29  & 1.87  \\
  \cmidrule{2-15}          & Trained & \multicolumn{1}{c|}{Qwen2.5-7B-Instruct} & 77.73  & 88.00  & 19.60  & 20.13  & 0.23  & 0.23  & 1.30  & 0.88  & 2.95  & 2.12  & 2.59  & 2.32  \\
  \cmidrule{2-15}          & \multicolumn{2}{c|}{Relative Improvements} & -3.4\% & -1.5\% & -26.4\% & 8.1\% & 3.5\% & 1.8\% & 15.4\% & -3.2\% & 11.9\% & 7.3\% & 13.0\% & 24.1\% \\
      \midrule
      \multirow{7}[8]{*}{10\_put\_dishwasher} & Oracle & \multicolumn{1}{c|}{GPT-5.2} & 62.60 & 65.53 & 37.68 & 27.63 & 0.17 & 0.16 & 2.66 & 2.81 & 3.14 & 3.96 & 2.91 & 3.42  \\
\cmidrule{2-15} & \multirow{4}[2]{*}{Base} & \multicolumn{1}{c|}{GPT-5.2} & 64.20 & 67.73 & 15.00 & 12.74 & 0.20 & 0.20 & 3.01 & 2.44 & 3.70 & 3.62 & 3.01 & 3.22  \\
            &       & \multicolumn{1}{c|}{DeepSeek-V3.1} & 62.00  & 56.60  & 11.49  & 18.71  & 0.39  & 0.39  & 2.95  & 2.99  & 3.91  & 4.18  & 3.02  & 3.23  \\
            &       & \multicolumn{1}{c|}{Qwen2.5-72B-Instruct} & 70.33  & 65.20  & 20.98  & 19.17  & 0.28  & 0.28  & 2.35  & 2.81  & 3.91  & 3.67  & 2.95  & 3.44  \\
            &       & \multicolumn{1}{c|}{Qwen2.5-7B-Instruct} & 74.13  & 91.33  & 19.17  & 26.45  & 0.25  & 0.24  & 1.05  & 0.66  & 2.79  & 2.03  & 2.57  & 1.67  \\
  \cmidrule{2-15}          & Trained & \multicolumn{1}{c|}{Qwen2.5-7B-Instruct} & 51.07  & 45.60  & 10.59  & 13.19  & 0.24  & 0.23  & 1.73  & 1.56  & 2.53  & 2.94  & 3.52  & 2.92  \\
  \cmidrule{2-15}          & \multicolumn{2}{c|}{Relative Improvements} & 31.1\% & 50.1\% & 44.8\% & 50.1\% & 4.8\% & 2.2\% & 65.0\% & 136.9\% & -9.3\% & 44.7\% & 36.9\% & 74.6\% \\
      \midrule
      \multirow{7}[8]{*}{16\_put\_dishwasher} & \multicolumn{1}{c}{Oracle} & \multicolumn{1}{c}{GPT-5.2} & 53.47 & 54.87 & 9.41 & 9.29 & 0.16 & 0.16 & 3.08 & 3.02 & 4.15 & 3.84 & 3.85 & 3.53  \\
\cmidrule{2-15} & \multirow{4}[2]{*}{Base} & \multicolumn{1}{c|}{GPT-5.2} & 49.33 & 56.27 & 13.12 & 13.00 & 0.20 & 0.20 & 2.60 & 2.64 & 3.53 & 3.49 & 2.99 & 3.35  \\
            &       & \multicolumn{1}{c|}{DeepSeek-V3.1} & 44.80  & 51.07  & 14.74  & 11.60  & 0.38  & 0.40  & 2.05  & 2.58  & 3.52  & 3.72  & 3.10  & 3.55  \\
            &       & \multicolumn{1}{c|}{Qwen2.5-72B-Instruct} & 57.27  & 58.53  & 8.21  & 9.34  & 0.28  & 0.28  & 2.75  & 2.20  & 3.58  & 3.35  & 3.60  & 3.21  \\
            &       & \multicolumn{1}{c|}{Qwen2.5-7B-Instruct} & 75.27  & 70.00  & 13.22  & 9.11  & 0.25  & 0.25  & 1.34  & 1.08  & 2.60  & 2.36  & 2.42  & 2.34  \\
  \cmidrule{2-15}          & Trained & \multicolumn{1}{c|}{Qwen2.5-7B-Instruct} & 53.27  & 56.33  & 17.97  & 9.60  & 0.24  & 0.24  & 1.76  & 1.91  & 3.38  & 3.29  & 3.59  & 2.88  \\
  \cmidrule{2-15}          & \multicolumn{2}{c|}{Relative Improvements} & 29.2\% & 19.5\% & -36.0\% & -5.5\% & 1.3\% & 3.1\% & 30.8\% & 76.6\% & 29.9\% & 39.2\% & 48.1\% & 23.3\% \\
      \midrule
      \multirow{7}[8]{*}{20\_prepare\_food} & Oracle & \multicolumn{1}{c|}{GPT-5.2} & 56.40 & 58.27 & 6.72 & 8.68 & 0.17 & 0.17 & 2.66 & 2.83 & 3.56 & 3.89 & 3.80 & 3.89  \\
\cmidrule{2-15} & \multirow{4}[2]{*}{Base} & \multicolumn{1}{c|}{GPT-5.2} & 63.60 & 66.27 & 13.56 & 15.06 & 0.21 & 0.21 & 2.26 & 2.64 & 3.38 & 3.75 & 2.96 & 3.28  \\
            &       & \multicolumn{1}{c|}{DeepSeek-V3.1} & 68.07  & 60.93  & 13.14  & 10.82  & 0.41  & 0.42  & 1.93  & 2.20  & 3.10  & 3.17  & 2.76  & 3.22  \\
            &       & \multicolumn{1}{c|}{Qwen2.5-72B-Instruct} & 69.00  & 67.13  & 17.58  & 10.07  & 0.29  & 0.29  & 2.52  & 2.49  & 3.74  & 3.50  & 3.34  & 3.29  \\
            &       & \multicolumn{1}{c|}{Qwen2.5-7B-Instruct} & 85.33  & 83.60  & 19.30  & 17.90  & 0.24  & 0.24  & 0.85  & 1.12  & 2.44  & 2.11  & 2.31  & 2.34  \\
  \cmidrule{2-15}          & Trained & \multicolumn{1}{c|}{Qwen2.5-7B-Instruct} & 72.33  & 56.40  & 15.64  & 12.52  & 0.23  & 0.22  & 1.34  & 1.48  & 3.16  & 2.99  & 3.70  & 3.32  \\
  \cmidrule{2-15}          & \multicolumn{2}{c|}{Relative Improvements} & 15.2\% & 32.5\% & 19.0\% & 30.1\% & 5.1\% & 6.7\% & 58.7\% & 32.2\% & 29.4\% & 42.1\% & 60.4\% & 41.9\% \\
      \midrule
      \multirow{7}[8]{*}{26\_prepare\_food} & Oracle & \multicolumn{1}{c|}{GPT-5.2} & 44.47 & 49.73 & 20.44 & 12.58 & 0.17 & 0.17 & 3.29 & 3.07 & 4.01 & 4.15 & 3.53 & 3.65  \\
\cmidrule{2-15} & \multirow{4}[2]{*}{Base} & \multicolumn{1}{c|}{GPT-5.2} & 72.13 & 30.67 & 31.58 & 13.74 & 0.21 & 0.20 & 2.65 & 3.25 & 3.67 & 3.82 & 2.76 & 3.55  \\
            &       & \multicolumn{1}{c|}{DeepSeek-V3.1} & 52.60  & 49.27  & 27.80  & 31.71  & 0.41  & 0.42  & 2.46  & 2.92  & 3.32  & 3.71  & 3.03  & 3.32  \\
            &       & \multicolumn{1}{c|}{Qwen2.5-72B-Instruct} & 47.73  & 47.87  & 23.26  & 19.80  & 0.32  & 0.30  & 2.97  & 2.87  & 4.19  & 3.94  & 3.67  & 3.17  \\
            &       & \multicolumn{1}{c|}{Qwen2.5-7B-Instruct} & 94.33  & 73.53  & 20.29  & 22.21  & 0.25  & 0.23  & 1.23  & 1.57  & 2.22  & 2.58  & 2.01  & 2.27  \\
  \cmidrule{2-15}          & Trained & \multicolumn{1}{c|}{Qwen2.5-7B-Instruct} & 64.93  & 59.27  & 19.93  & 22.37  & 0.23  & 0.22  & 1.47  & 1.47  & 2.87  & 3.32  & 3.21  & 3.10  \\
  \cmidrule{2-15}          & \multicolumn{2}{c|}{Relative Improvements} & 31.2\% & 19.4\% & 1.8\% & -0.7\% & 5.8\% & 3.4\% & 18.9\% & -6.2\% & 29.7\% & 28.7\% & 59.6\% & 36.9\% \\
      \midrule
      \multirow{7}[8]{*}{30\_put\_fridge} & Oracle & \multicolumn{1}{c|}{GPT-5.2} & 38.87 & 38.93 & 9.77 & 11.56 & 0.17 & 0.16 & 3.19 & 2.91 & 4.02 & 3.60 & 3.96 & 3.93  \\
\cmidrule{2-15} & \multirow{4}[2]{*}{Base} & \multicolumn{1}{c|}{GPT-5.2} & 46.20 & 40.33 & 18.06 & 8.25 & 0.20 & 0.20 & 2.43 & 2.62 & 3.28 & 4.22 & 2.68 & 3.32  \\
            &       & \multicolumn{1}{c|}{DeepSeek-V3.1} & 47.60  & 43.47  & 13.30  & 16.46  & 0.40  & 0.43  & 2.39  & 2.69  & 3.36  & 3.76  & 2.91  & 3.20  \\
            &       & \multicolumn{1}{c|}{Qwen2.5-72B-Instruct} & 48.47  & 45.93  & 7.85  & 14.31  & 0.29  & 0.29  & 2.50  & 2.65  & 3.71  & 3.88  & 3.73  & 3.54  \\
            &       & \multicolumn{1}{c|}{Qwen2.5-7B-Instruct} & 52.73  & 63.67  & 7.70  & 13.04  & 0.24  & 0.24  & 1.50  & 1.14  & 2.92  & 2.30  & 2.54  & 2.49  \\
  \cmidrule{2-15}          & Trained & \multicolumn{1}{c|}{Qwen2.5-7B-Instruct} & 47.53  & 47.27  & 10.40  & 12.54  & 0.22  & 0.22  & 1.38  & 1.26  & 3.33  & 3.32  & 3.53  & 2.94  \\
  \cmidrule{2-15}          & \multicolumn{2}{c|}{Relative Improvements} & 9.9\% & 25.8\% & -35.1\% & 3.8\% & 8.5\% & 7.4\% & -8.0\% & 10.6\% & 14.1\% & 43.9\% & 39.0\% & 17.8\% \\
      \midrule
      \multirow{7}[8]{*}{32\_put\_fridge} & Oracle & \multicolumn{1}{c|}{GPT-5.2} & 74.33 & 78.13 & 13.57 & 48.66 & 0.17 & 0.17 & 2.35 & 3.40 & 4.14 & 4.05 & 3.82 & 3.16  \\
\cmidrule{2-15} & \multirow{4}[2]{*}{Base} & \multicolumn{1}{c|}{GPT-5.2} & 82.87 & 76.07 & 21.13 & 19.26 & 0.21 & 0.21 & 2.21 & 2.27 & 3.65 & 3.42 & 2.89 & 3.47  \\
            &       & \multicolumn{1}{c|}{DeepSeek-V3.1} & 77.87  & 68.67  & 15.24  & 7.92  & 0.42  & 0.42  & 2.35  & 2.44  & 3.66  & 3.91  & 2.90  & 3.54  \\
            &       & \multicolumn{1}{c|}{Qwen2.5-72B-Instruct} & 80.27  & 72.13  & 14.81  & 11.53  & 0.29  & 0.29  & 1.99  & 2.65  & 3.37  & 3.79  & 2.87  & 3.35  \\
            &       & \multicolumn{1}{c|}{Qwen2.5-7B-Instruct} & 98.13  & 116.27  & 14.22  & 32.67  & 0.24  & 0.23  & 1.14  & 1.05  & 2.39  & 2.07  & 2.65  & 2.36  \\
  \cmidrule{2-15}          & Trained & \multicolumn{1}{c|}{Qwen2.5-7B-Instruct} & 96.80  & 63.47  & 22.68  & 14.13  & 0.23  & 0.23  & 1.09  & 1.15  & 2.64  & 2.70  & 3.00  & 3.13  \\
  \cmidrule{2-15}          & \multicolumn{2}{c|}{Relative Improvements} & 1.4\% & 45.4\% & -59.5\% & 56.7\% & 4.5\% & 0.9\% & -4.1\% & 9.3\% & 10.4\% & 30.5\% & 13.3\% & 33.0\% \\
      \midrule
      \multirow{7}[8]{*}{40\_setup\_table} & Oracle & \multicolumn{1}{c|}{GPT-5.2} & 69.20 & 43.07 & 14.50 & 7.78 & 0.16 & 0.17 & 3.04 & 3.38 & 3.76 & 3.92 & 3.95 & 3.91  \\
\cmidrule{2-15} & \multirow{4}[2]{*}{Base} & \multicolumn{1}{c|}{GPT-5.2} & 75.60 & 62.93 & 23.26 & 15.65 & 0.20 & 0.20 & 2.41 & 3.40 & 3.25 & 4.08 & 2.77 & 3.58  \\
            &       & \multicolumn{1}{c|}{DeepSeek-V3.1} & 69.13  & 66.93  & 23.52  & 13.66  & 0.40  & 0.39  & 2.18  & 2.61  & 3.45  & 3.67  & 3.45  & 3.62  \\
            &       & \multicolumn{1}{c|}{Qwen2.5-72B-Instruct} & 68.67  & 66.47  & 16.91  & 15.56  & 0.28  & 0.28  & 2.91  & 2.61  & 3.73  & 3.62  & 3.63  & 3.66  \\
            &       & \multicolumn{1}{c|}{Qwen2.5-7B-Instruct} & 90.27  & 83.53  & 24.97  & 15.61  & 0.24  & 0.23  & 1.22  & 1.31  & 2.40  & 2.46  & 2.21  & 2.34  \\
  \cmidrule{2-15}          & Trained & \multicolumn{1}{c|}{Qwen2.5-7B-Instruct} & 64.13  & 51.27  & 22.62  & 18.01  & 0.23  & 0.23  & 1.44  & 1.62  & 2.78  & 3.44  & 3.47  & 3.47  \\
  \cmidrule{2-15}          & \multicolumn{2}{c|}{Relative Improvements} & 29.0\% & 38.6\% & 9.4\% & -15.4\% & 3.9\% & 3.8\% & 18.2\% & 23.2\% & 15.8\% & 39.9\% & 57.1\% & 48.6\% \\
      \midrule
      \multirow{7}[8]{*}{49\_setup\_table} & Oracle & \multicolumn{1}{c|}{GPT-5.2} & 49.47 & 53.73 & 8.27 & 11.25 & 0.17 & 0.17 & 3.29 & 3.00 & 3.83 & 3.92 & 4.03 & 3.51  \\
\cmidrule{2-15} & \multirow{4}[2]{*}{Base} & \multicolumn{1}{c|}{GPT-5.2} & 63.53 & 55.40 & 12.14 & 12.48 & 0.21 & 0.21 & 2.87 & 3.24 & 3.49 & 3.11 & 2.86 & 3.26  \\
            &       & \multicolumn{1}{c|}{DeepSeek-V3.1} & 64.33  & 55.87  & 15.22  & 13.98  & 0.38  & 0.41  & 2.13  & 2.28  & 3.22  & 3.52  & 2.56  & 2.96  \\
            &       & \multicolumn{1}{c|}{Qwen2.5-72B-Instruct} & 48.27  & 49.87  & 9.96  & 11.07  & 0.29  & 0.29  & 2.45  & 2.47  & 3.64  & 3.93  & 3.13  & 3.71  \\
            &       & \multicolumn{1}{c|}{Qwen2.5-7B-Instruct} & 66.13  & 73.33  & 22.46  & 21.63  & 0.25  & 0.24  & 1.11  & 1.23  & 2.71  & 2.39  & 2.37  & 2.26  \\
  \cmidrule{2-15}          & Trained & \multicolumn{1}{c|}{Qwen2.5-7B-Instruct} & 62.13  & 56.33  & 15.47  & 15.83  & 0.24  & 0.23  & 1.30  & 1.52  & 2.89  & 3.03  & 3.09  & 2.76  \\
  \cmidrule{2-15}          & \multicolumn{2}{c|}{Relative Improvements} & 6.0\% & 23.2\% & 31.1\% & 26.8\% & 5.6\% & 3.6\% & 16.9\% & 23.7\% & 6.6\% & 26.9\% & 30.3\% & 22.1\% \\
      \bottomrule
      \end{tabular}%
      }
    \label{apx:tab:cwah}%
  \end{table}%

\begin{table}[htbp]
  \centering
  \caption{The detailed evaluation results across sub tasks on CollabBench for Cook-MutiPlayer.}
  \resizebox{0.99\linewidth}{!}{
 \begin{tabular}{c|c|r|cccccc|cccccc}
      \toprule
      \multicolumn{15}{c}{\textbf{Cook-MultiPlayer}} \\
      \midrule
      \multicolumn{3}{r|}{\multirow{2}[2]{*}{Metric}} & \multicolumn{6}{c|}{Efficiency}               & \multicolumn{6}{c}{Affective} \\
      \multicolumn{3}{r|}{} & \multicolumn{2}{c}{Score} & \multicolumn{2}{c}{Std.} & \multicolumn{2}{c|}{\#Tokens(k)} & \multicolumn{2}{c}{Helpfulness} & \multicolumn{2}{c}{Trustfulness} & \multicolumn{2}{c}{Empathy} \\
      \midrule
      Task  & Method & \multicolumn{1}{c|}{LLMs} & Agent 1 & Agent 2 & Agent 1 & Agent 2 & Agent 1 & Agent 2 & Agent 1 & Agent 2 & Agent 1 & Agent 2 & Agent 1 & Agent 2 \\
      \midrule
      \multirow{7}[8]{*}{asymmetric\_advantages} & Oracle & \multicolumn{1}{c|}{GPT-5.2} & 308.00  & 272.00  & 32.78  & 12.65  & 0.31  & 0.20  & 2.95  & 2.61  & 4.02  & 3.49  & 3.53  & 3.31  \\
  \cmidrule{2-15}          & \multirow{4}[2]{*}{Base} & \multicolumn{1}{c|}{GPT-5.2} & 308.00  & 264.00  & 18.21  & 15.49  & 0.20  & 0.21  & 1.93  & 2.33  & 3.05  & 3.28  & 2.28  & 2.73  \\
            &       & \multicolumn{1}{c|}{DeepSeek-V3.1} & 300.00  & 278.67  & 30.24  & 23.26  & 0.31  & 0.31  & 1.70  & 1.99  & 2.71  & 3.04  & 2.45  & 2.64  \\
            &       & \multicolumn{1}{c|}{Qwen2.5-72B-Instruct} & 284.00  & 220.00  & 30.43  & 28.28  & 0.28  & 0.27  & 1.37  & 1.58  & 2.58  & 2.79  & 2.21  & 2.53  \\
            &       & \multicolumn{1}{c|}{Qwen2.5-7B-Instruct} & 186.67  & 194.67  & 40.47  & 37.39  & 0.24  & 0.24  & 0.49  & 0.59  & 1.87  & 1.92  & 1.75  & 1.86  \\
  \cmidrule{2-15}          & Trained & \multicolumn{1}{c|}{Qwen2.5-7B-Instruct} & 197.33  & 214.67  & 59.94  & 29.73  & 0.23  & 0.22  & 0.65  & 0.72  & 2.10  & 1.98  & 2.14  & 2.04  \\
  \cmidrule{2-15}          & \multicolumn{2}{c|}{Relative Improvements} & 5.71\% & 10.27\% & -48.09\% & 20.49\% & 1.01\% & 6.09\% & 34.0\% & 20.6\% & 12.3\% & 3.0\% & 21.9\% & 9.6\% \\
      \midrule
      \multirow{7}[8]{*}{coordination\_ring} & Oracle & \multicolumn{1}{c|}{GPT-5.2} & 112.00  & 124.00  & 79.57  & 76.42  & 0.30  & 0.20  & 3.51  & 2.43  & 4.41  & 3.70  & 3.95  & 3.71  \\
  \cmidrule{2-15}          & \multirow{4}[2]{*}{Base} & \multicolumn{1}{c|}{GPT-5.2} & 101.33  & 98.67  & 74.25  & 78.73  & 0.19  & 0.20  & 2.68  & 2.17  & 3.34  & 3.69  & 2.87  & 2.76  \\
            &       & \multicolumn{1}{c|}{DeepSeek-V3.1} & 113.33  & 113.33  & 65.32  & 71.18  & 0.32  & 0.31  & 2.17  & 2.15  & 3.39  & 3.04  & 2.92  & 2.91  \\
            &       & \multicolumn{1}{c|}{Qwen2.5-72B-Instruct} & 110.67  & 117.33  & 74.01  & 43.34  & 0.26  & 0.25  & 1.97  & 2.15  & 3.17  & 3.78  & 2.89  & 3.21  \\
            &       & \multicolumn{1}{c|}{Qwen2.5-7B-Instruct} & 90.67  & 62.67  & 51.20  & 38.45  & 0.24  & 0.23  & 0.69  & 0.57  & 1.93  & 2.01  & 1.94  & 1.82  \\
  \cmidrule{2-15}          & Trained & \multicolumn{1}{c|}{Qwen2.5-7B-Instruct} & 98.67  & 90.67  & 40.33  & 48.91  & 0.23  & 0.22  & 0.87  & 0.73  & 2.27  & 2.30  & 2.02  & 2.12  \\
  \cmidrule{2-15}          & \multicolumn{2}{c|}{Relative Improvements} & 8.82\% & 44.68\% & 21.22\% & -27.22\% & 3.76\% & 2.59\% & 25.9\% & 28.9\% & 17.8\% & 14.7\% & 4.0\% & 16.6\% \\
      \midrule
      \multirow{7}[8]{*}{counter\_circuit} & Oracle & \multicolumn{1}{c|}{GPT-5.2} & 114.67  & 137.14  & 52.63  & 50.75  & 0.30  & 0.20  & 3.28  & 3.11  & 4.26  & 4.10  & 4.05  & 4.39  \\
  \cmidrule{2-15}          & \multirow{4}[2]{*}{Base} & \multicolumn{1}{c|}{GPT-5.2} & 137.33  & 137.33  & 45.27  & 45.27  & 0.19  & 0.20  & 1.93  & 2.58  & 3.56  & 3.63  & 2.68  & 3.06  \\
            &       & \multicolumn{1}{c|}{DeepSeek-V3.1} & 121.33  & 121.33  & 41.03  & 41.03  & 0.30  & 0.30  & 2.27  & 2.24  & 3.40  & 3.11  & 3.14  & 3.08  \\
            &       & \multicolumn{1}{c|}{Qwen2.5-72B-Instruct} & 156.00  & 154.67  & 36.41  & 38.16  & 0.27  & 0.27  & 1.31  & 1.38  & 2.77  & 2.80  & 2.39  & 2.50  \\
            &       & \multicolumn{1}{c|}{Qwen2.5-7B-Instruct} & 93.33  & 94.67  & 40.47  & 38.16  & 0.24  & 0.23  & 0.49  & 0.63  & 1.92  & 1.91  & 1.94  & 2.01  \\
  \cmidrule{2-15}          & Trained & \multicolumn{1}{c|}{Qwen2.5-7B-Instruct} & 116.00  & 117.33  & 33.12  & 34.72  & 0.23  & 0.22  & 0.79  & 0.59  & 2.59  & 2.28  & 2.42  & 2.44  \\
  \cmidrule{2-15}          & \multicolumn{2}{c|}{Relative Improvements} & 24.29\% & 23.94\% & 18.16\% & 9.01\% & 4.59\% & 3.38\% & 59.4\% & -5.5\% & 34.7\% & 19.4\% & 25.0\% & 21.8\% \\
      \midrule
      \multirow{7}[8]{*}{cramped\_room} & Oracle & \multicolumn{1}{c|}{GPT-5.2} & 46.67  & 25.33  & 36.77  & 9.15  & 0.26  & 0.18  & 1.70  & 1.98  & 3.33  & 3.19  & 3.07  & 3.33  \\
  \cmidrule{2-15}          & \multirow{4}[2]{*}{Base} & \multicolumn{1}{c|}{GPT-5.2} & 41.33  & 44.00  & 25.60  & 21.65  & 0.19  & 0.20  & 0.96  & 1.13  & 2.32  & 2.62  & 1.97  & 2.02  \\
            &       & \multicolumn{1}{c|}{DeepSeek-V3.1} & 38.67  & 41.33  & 28.75  & 25.60  & 0.29  & 0.32  & 1.68  & 0.73  & 3.02  & 2.05  & 2.77  & 1.37  \\
            &       & \multicolumn{1}{c|}{Qwen2.5-72B-Instruct} & 45.33  & 42.67  & 28.75  & 24.16  & 0.26  & 0.26  & 1.77  & 0.78  & 3.18  & 2.37  & 2.59  & 2.15  \\
            &       & \multicolumn{1}{c|}{Qwen2.5-7B-Instruct} & 30.67  & 32.00  & 14.86  & 16.85  & 0.23  & 0.23  & 0.29  & 0.56  & 1.95  & 1.99  & 1.88  & 2.00  \\
  \cmidrule{2-15}          & Trained & \multicolumn{1}{c|}{Qwen2.5-7B-Instruct} & 40.00  & 40.00  & 15.12  & 15.12  & 0.23  & 0.22  & 0.59  & 0.42  & 2.14  & 1.82  & 2.13  & 1.94  \\
  \cmidrule{2-15}          & \multicolumn{2}{c|}{Relative Improvements} & 30.43\% & 25.00\% & -1.71\% & 10.29\% & 0.00\% & 5.16\% & 102.7\% & -24.6\% & 9.3\% & -8.6\% & 12.9\% & -3.3\% \\
      \midrule
      \multirow{7}[8]{*}{forced\_coordination} & Oracle & \multicolumn{1}{c|}{GPT-5.2} & 136.00  & 116.00  & 63.34  & 18.82  & 0.32  & 0.21  & 1.99  & 1.65  & 3.21  & 2.84  & 2.80  & 2.64  \\
  \cmidrule{2-15}          & \multirow{4}[2]{*}{Base} & \multicolumn{1}{c|}{GPT-5.2} & 88.00  & 84.00  & 46.48  & 54.00  & 0.22  & 0.22  & 0.67  & 1.13  & 2.18  & 2.21  & 1.55  & 1.99  \\
            &       & \multicolumn{1}{c|}{DeepSeek-V3.1} & 109.33  & 101.33  & 36.15  & 41.49  & 0.32  & 0.32  & 1.13  & 1.17  & 2.36  & 2.16  & 2.05  & 2.42  \\
            &       & \multicolumn{1}{c|}{Qwen2.5-72B-Instruct} & 81.33  & 66.67  & 40.33  & 41.63  & 0.28  & 0.28  & 0.45  & 0.39  & 2.13  & 1.53  & 2.19  & 2.02  \\
            &       & \multicolumn{1}{c|}{Qwen2.5-7B-Instruct} & 33.33  & 34.67  & 19.52  & 20.01  & 0.24  & 0.23  & 0.31  & 0.38  & 1.91  & 1.53  & 1.81  & 1.71  \\
  \cmidrule{2-15}          & Trained & \multicolumn{1}{c|}{Qwen2.5-7B-Instruct} & 44.00  & 44.00  & 21.65  & 21.65  & 0.23  & 0.22  & 0.80  & 0.31  & 2.18  & 1.59  & 1.92  & 1.84  \\
  \cmidrule{2-15}          & \multicolumn{2}{c|}{Relative Improvements} & 32.00\% & 26.92\% & -10.91\% & -8.17\% & 3.20\% & 6.03\% & 156.3\% & -18.4\% & 14.3\% & 3.7\% & 6.2\% & 7.4\% \\
      \bottomrule
      \end{tabular}%
    }
  \label{apx:tab:cook}%
\end{table}%
\subsection{Anthropomorphic Results}\label{appx:anthropomorphic}

\subsubsection{Cook-MultiPlayer Results}\label{apx:cook}

\begin{figure}[htbp]
  \centering
\begin{minipage}{0.22\linewidth}\centering
    \includegraphics[width=0.99\textwidth]{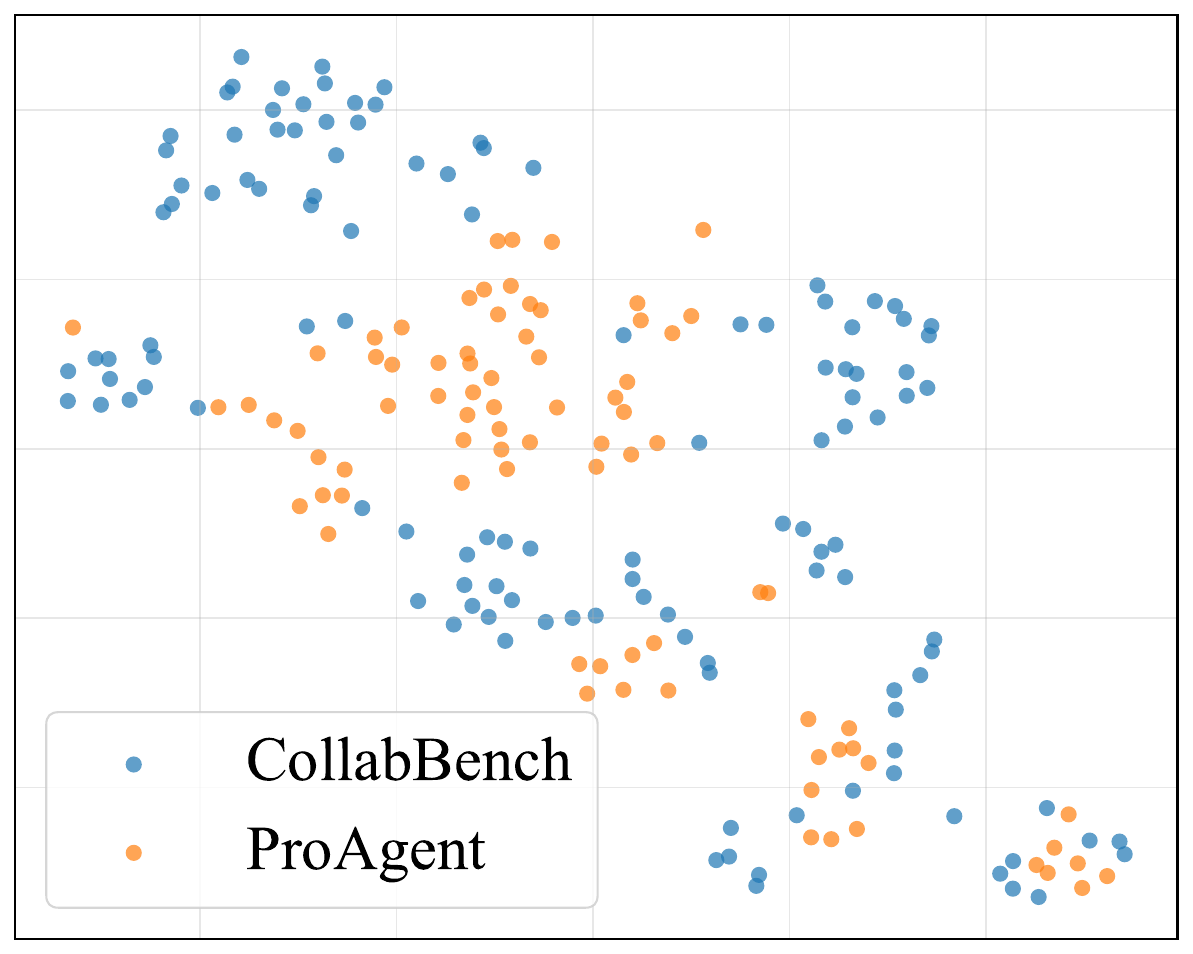}\\
    (a) Visualization of \\Players’ Trajectory Segments
\end{minipage}
\begin{minipage}{0.26\linewidth}\centering
    \includegraphics[width=0.99\textwidth]{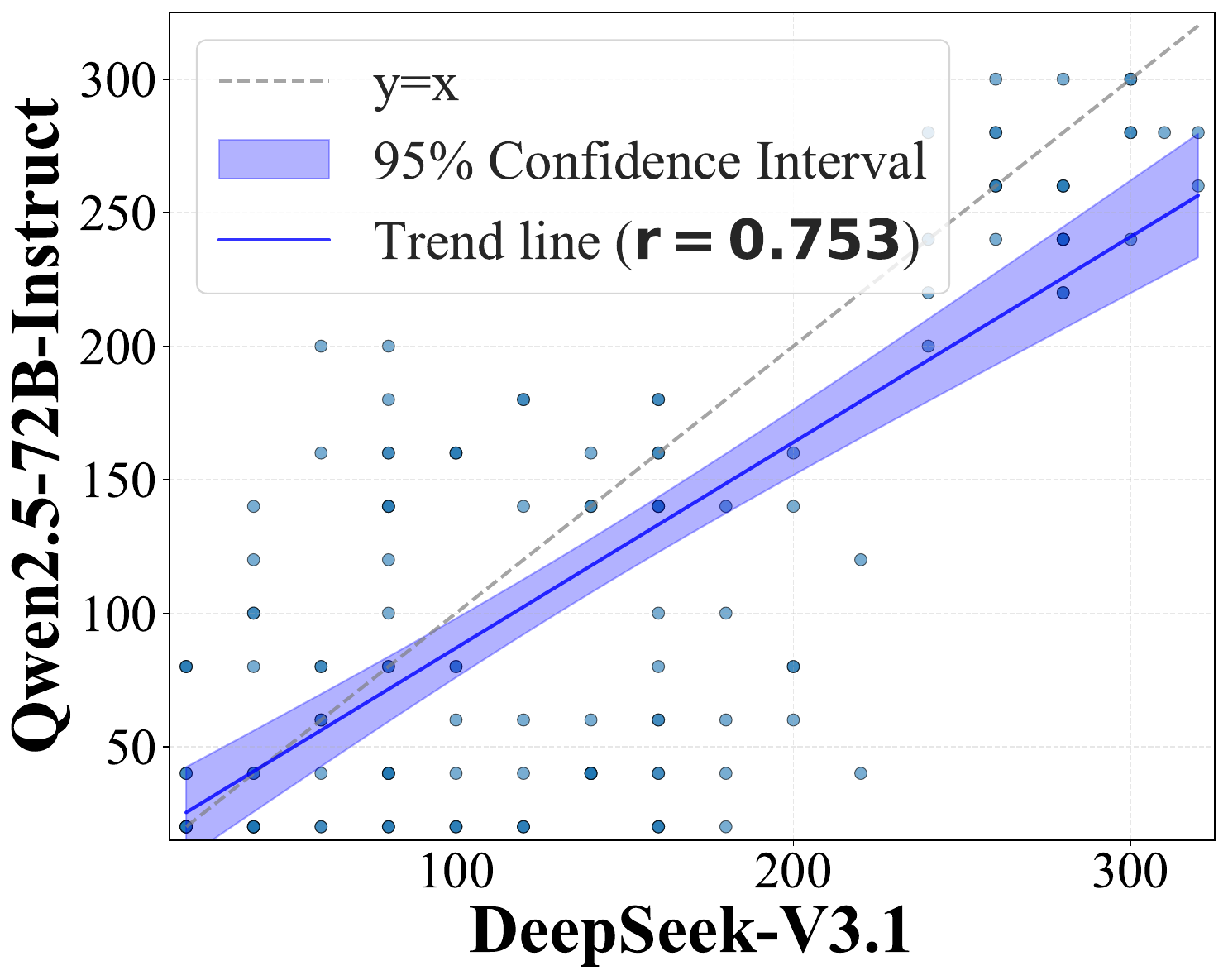}\\
    (b) Correlation between Qwen and Deepseek
\end{minipage}
\begin{minipage}{0.48\linewidth}\centering
    \includegraphics[width=0.99\textwidth]{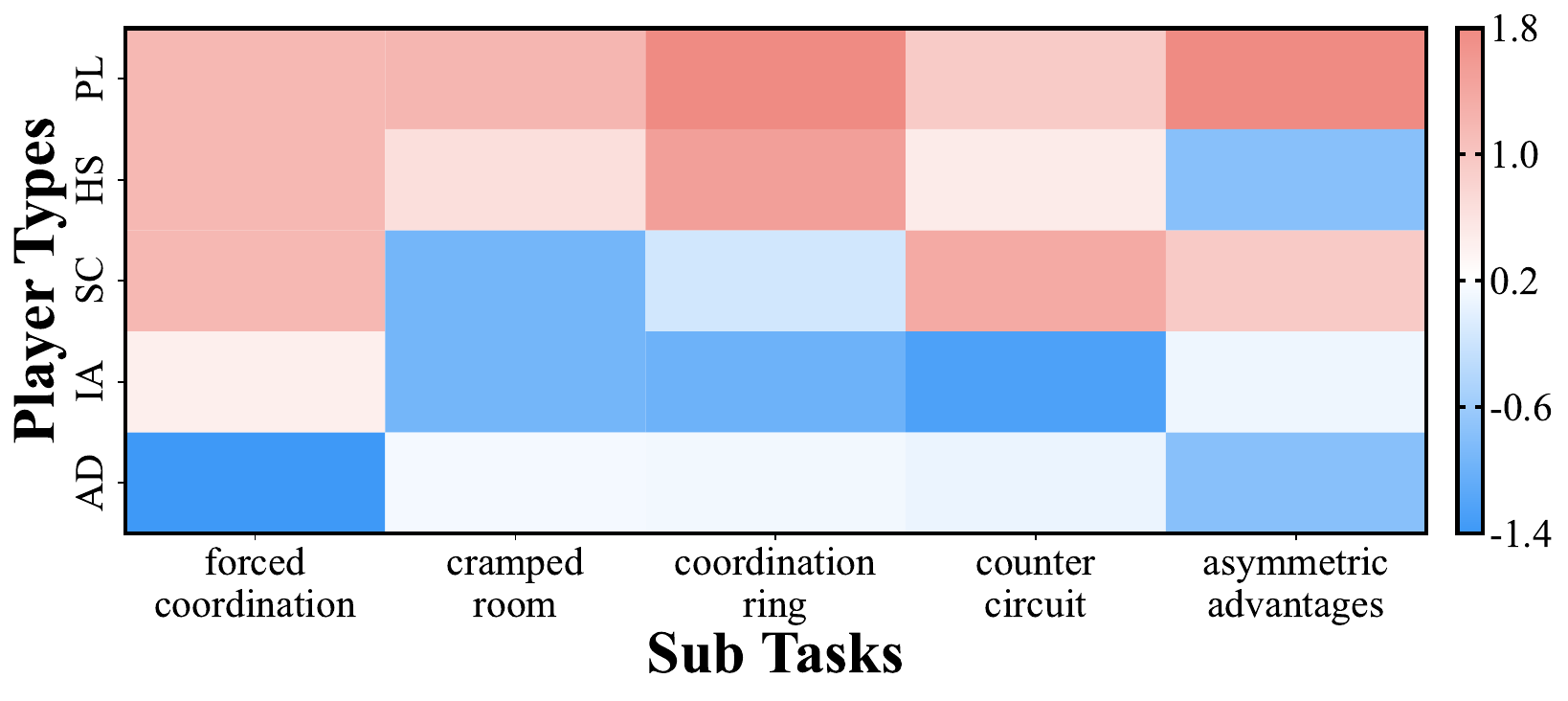}\\
    (c) Heatmap of Score Distributions
\end{minipage}
  \caption{The anthropomorphic results on Cook-MultiPlayer.}
  \label{apx:fig:cook3}
\end{figure}

Here, we provide the anthropomorphic results on Cook-MultiPlayer in Figure~\ref{apx:fig:cook3}: (a) compares the performance distributions between CollabBench and ProAgent. (b) shows the correlation between the performances of Qwen2.5-72B-Instruct and DeepSeek-V3.1, with a correlation coefficient of 0.753. and (c) shows the score distributions across five representative profiles. For detailed player type descriptions and the corresponding example profiles, please refer to Appendix~\ref{apx:type}.

\subsubsection{Diversity Results Based on Standard Deviation}\label{apx:stand}
\begin{table}[htbp]
  \centering
  \caption{Diversity comparison between CollabBench and baseline methods, measured by standard deviation.}

    \begin{tabular}{c|c|cc|cc}
    \toprule
    \multicolumn{2}{c|}{\multirow{2}[4]{*}{Metrics}} & \multicolumn{2}{c|}{CWAH-MultiPlayer} & \multicolumn{2}{c}{Cook-MultiPlayer} \\
\cmidrule{3-6}    \multicolumn{2}{c|}{} & CoELA & CollabBench & ProAgent & CollabBench \\
    \midrule
    \multirow{2}[2]{*}{Std. $\uparrow$} & Agent1 & 13.26 & \textbf{25.17} & 29.88 & \textbf{40.30} \\
          & Agent2 & 13.26 & \textbf{26.28} & 29.88 & \textbf{46.24} \\
    \bottomrule
    \end{tabular}%
  \label{apx:tab:anth_std_diver}%
\end{table}%
Here, we provide the comparison results between CollabBench and the baseline methods based on Standard Deviation across two game environments, using CoELA as the baseline method for CWAH-MultiPlayer and ProAgent for Cook-MultiPlayer as baselines. As shown in Table~\ref{apx:tab:anth_std_diver}, CollabBench shows consistently greater performance variability.

\subsubsection{Quantitative Results of Interaction Diversity}\label{apx:diver_quant}
We provide a quantitative comparison results to demonstrate the diversity of anthropomorphic collaborative behaviors on CollabBench. The detailed metrics, implementation details, and the results analysis are as follows.

\textbf{Quantitative Metrics.}

\begin{table}[htbp]
  \centering
  \caption{The comparison results regarding anthropomorphic diversity on CollabBench.}

    \begin{tabular}{c|cc|cc}
    \toprule
    \multirow{2}[4]{*}{\textbf{Metrics}} & \multicolumn{2}{c|}{\textbf{CWAH-MultiPlayer}} & \multicolumn{2}{c}{\textbf{Cook-MultiPlayer}} \\
\cmidrule{2-5}          & CoELA & CollabBench & ProAgent & CollabBench \\
    \midrule
    Spread $\uparrow$& 0.68  & \textbf{0.81} & 0.68  & \textbf{0.73} \\
    $\text{Cluster}_\xi$ $\uparrow$& 16.50 & \textbf{30.50} & 14.80 & \textbf{20.60} \\
    \bottomrule
    \end{tabular}
  \label{apx:tab:diver}%
\end{table}%

$\bullet$ \textbf{Spread} measures the overall dispersion of trajectory segments. Let embeded trajectory segments be $[\mathbf{x}_1, \mathbf{x}_2, \dots, \mathbf{x}_Q] \in \mathbb{R}^d$, where $Q$ denotes the total number of trajectory segments. The centroid of the set of trajectory segments $\mathbf{c}$ is defined as $\mathbf{c}=\frac{1}{Q}\sum_{i=1}^Q{\mathbf{x}_i}$. The cosine distance from each trajectory segment $\mathbf{x}_i$ to the centroid $\mathbf{c}$ is defined as $d_i = 1 - \frac{\mathbf{x}_i \cdot \mathbf{c}}{\|\mathbf{x}_i\|  \|\mathbf{c}\|}$. The global Spread is then given by $\text{Spread}=\frac{1}{Q}\sum_{i=1}^Q{d_i}$. A higher $\text{Spread}$ indicates that trajectory segments are more widely dispersed, reflecting greater spatial diversity.

$\bullet$ \textbf{Cluster} quantifies the diversity of trajectory segments generated by different player profiles. Given $Q$ embeded trajectory segments $\mathbf{x}_i \in \mathbb{R}^d$, the pairwise cosine distance is defined as $d_{ij} = 1 - \frac{\mathbf{x}_i \cdot \mathbf{x}_j}{\|\mathbf{x}_i\|  \|\mathbf{x}_j\|} $. 
Following~\cite{2024agglomerative}, we use agglomerative clustering with ``complete'' linkage, grouping trajectory segments into the same cluster if their pairwise distance is below threshold $\xi$. For a given threshold $\xi$, the number of clusters is defined as:
\begin{equation}
    \text{Cluster}_\xi = \bigl| \{ C_1, C_2, \dots, C_K \} \bigr|.
\end{equation}
A higher cluster number indicates that trajectory segments form more distinct behavioral patterns, reflecting greater structural diversity.

\textbf{Implementation Details}. The threshold $\xi$ is set to 0.1. We segment LLM-based high-level trajectories into units of three time windows. and embed the segments using Qwen3-Embedding-4B.

\textbf{Results Analysis}. As shown in Table~\ref{apx:tab:diver}, CollabBench exhibits higher diversity than baseline methods in both spatial distribution and structural patterns. This demonstrates that our method can cover a wider range of collaborative behaviors.

\subsubsection{Representative Player Types}\label{apx:type}
Here, we provide the five selected player type descriptions with their corresponding example profiles, listed in Figure~\ref{apx:fig:player} for CWAH-MultiPlayer and Cook-MultiPlayer. We additionally provide the corresponding trajectory demonstrations in Appendix~\ref{apx:demo}.

\begin{figure*}[htbp]
\centering
\includegraphics[width=0.99\linewidth]{Figure/playertype.pdf}
\caption{The player type descriptions and example profiles in CWAH-MultiPlayer and Cook-MultiPlayer.}
\label{apx:fig:player}
\end{figure*}

\subsubsection{Case Analysis of Performance Degradation
}\label{apx:case_degra}

\begin{figure}[htbp]
  \centering
\begin{minipage}{0.49\linewidth}\centering
    \includegraphics[width=0.99\textwidth]{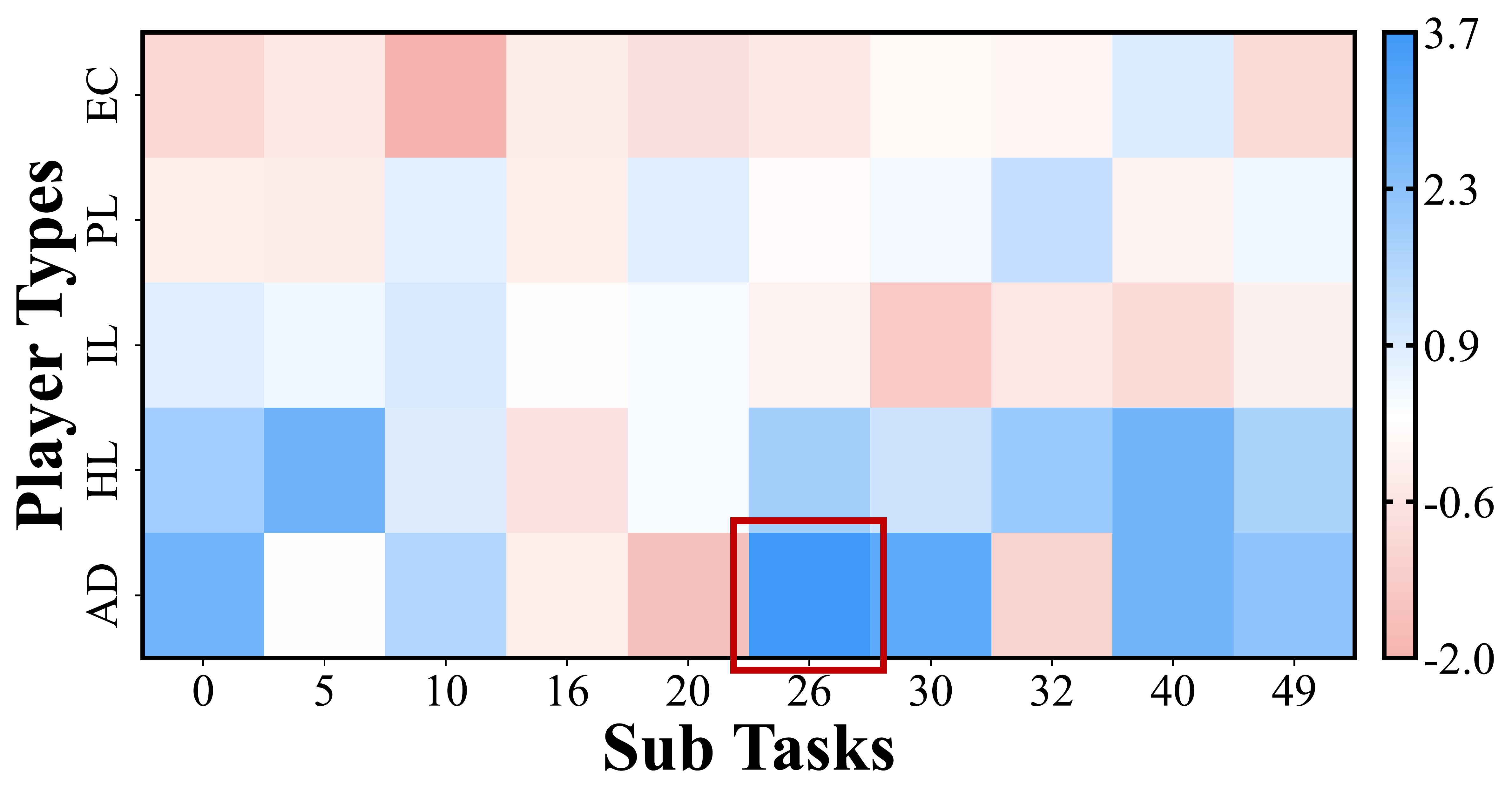}\\
    (a) CWAH-MultiPlayer
\end{minipage}
\begin{minipage}{0.49\linewidth}\centering
    \includegraphics[width=0.99\textwidth]{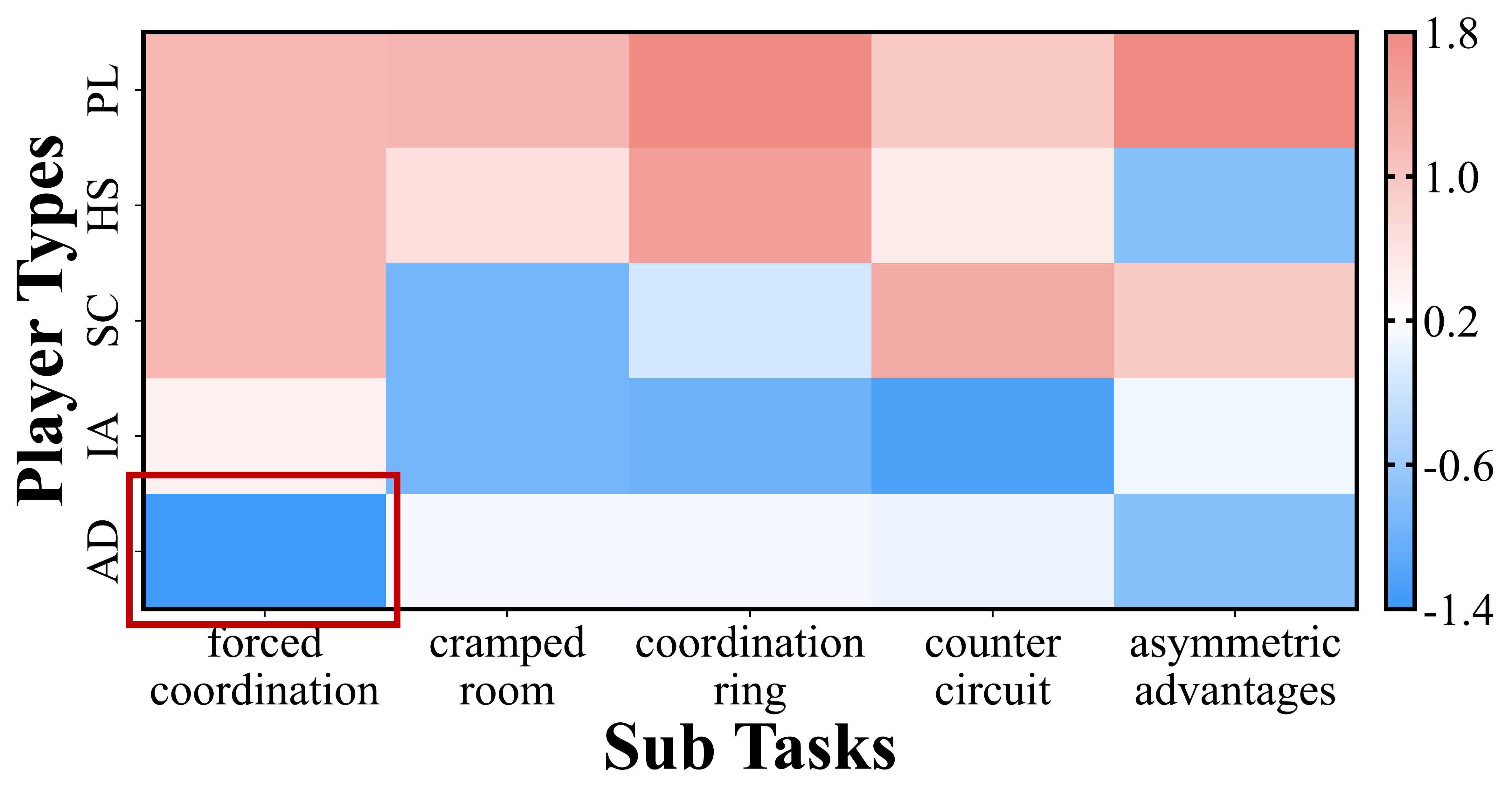}\\
    (b) Cook-MultiPlayer
\end{minipage}
  \caption{Heatmap of score distribution in two game environments. Lowest-performing combinations: player type ``AD'' with Task 26 in CWAH-MultiPlayer, and player type ``AH'' with the forced-coordination task in Cook-MultiPlayer.}
  \label{apx:fig:heat}
\end{figure}
\begin{figure}[htbp]
  \centering
\begin{minipage}{0.45\linewidth}\centering
    \includegraphics[width=0.99\textwidth]{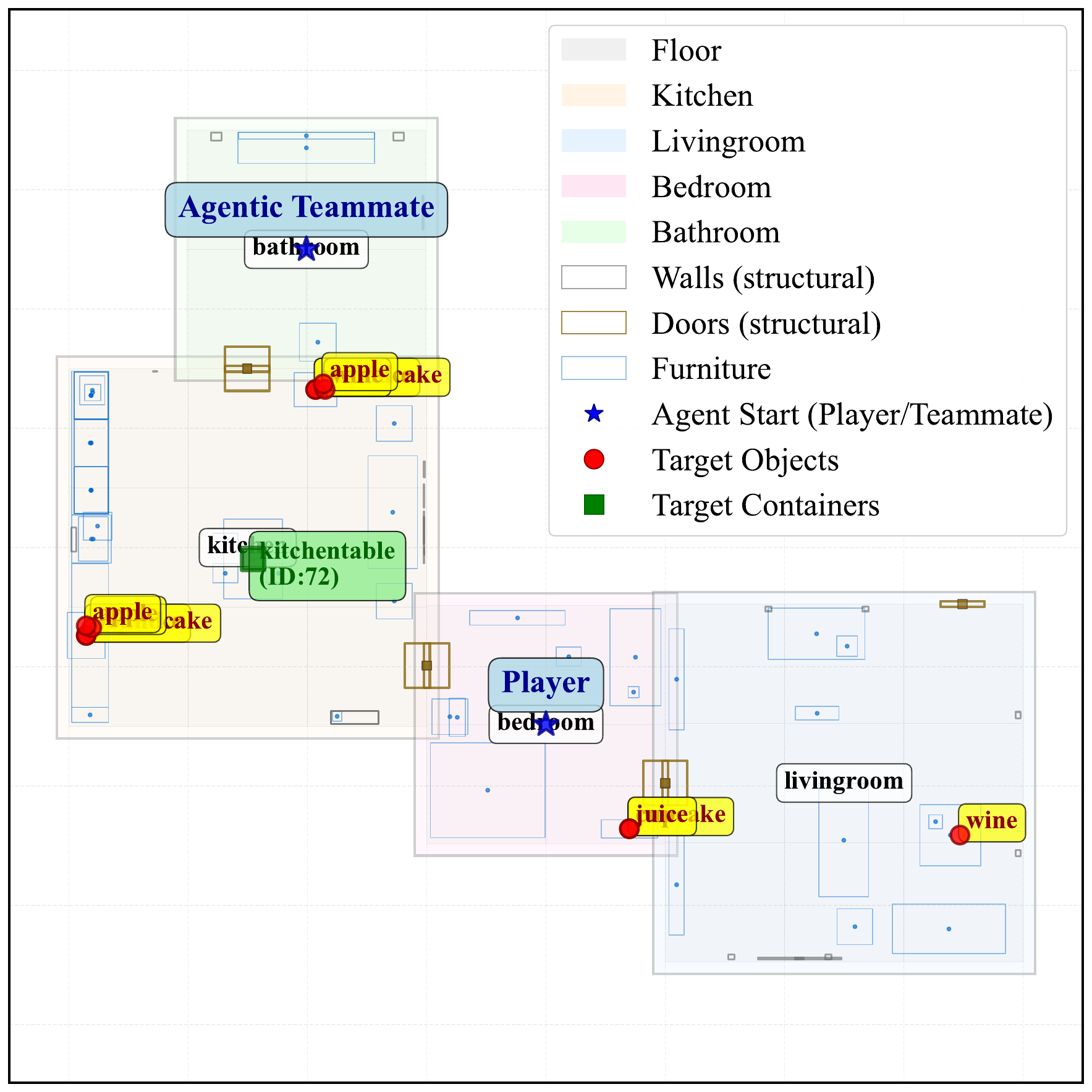}\\
    (a) CWAH-MultiPlayer
\end{minipage}
\begin{minipage}{0.45\linewidth}\centering
    \includegraphics[width=0.99\textwidth]{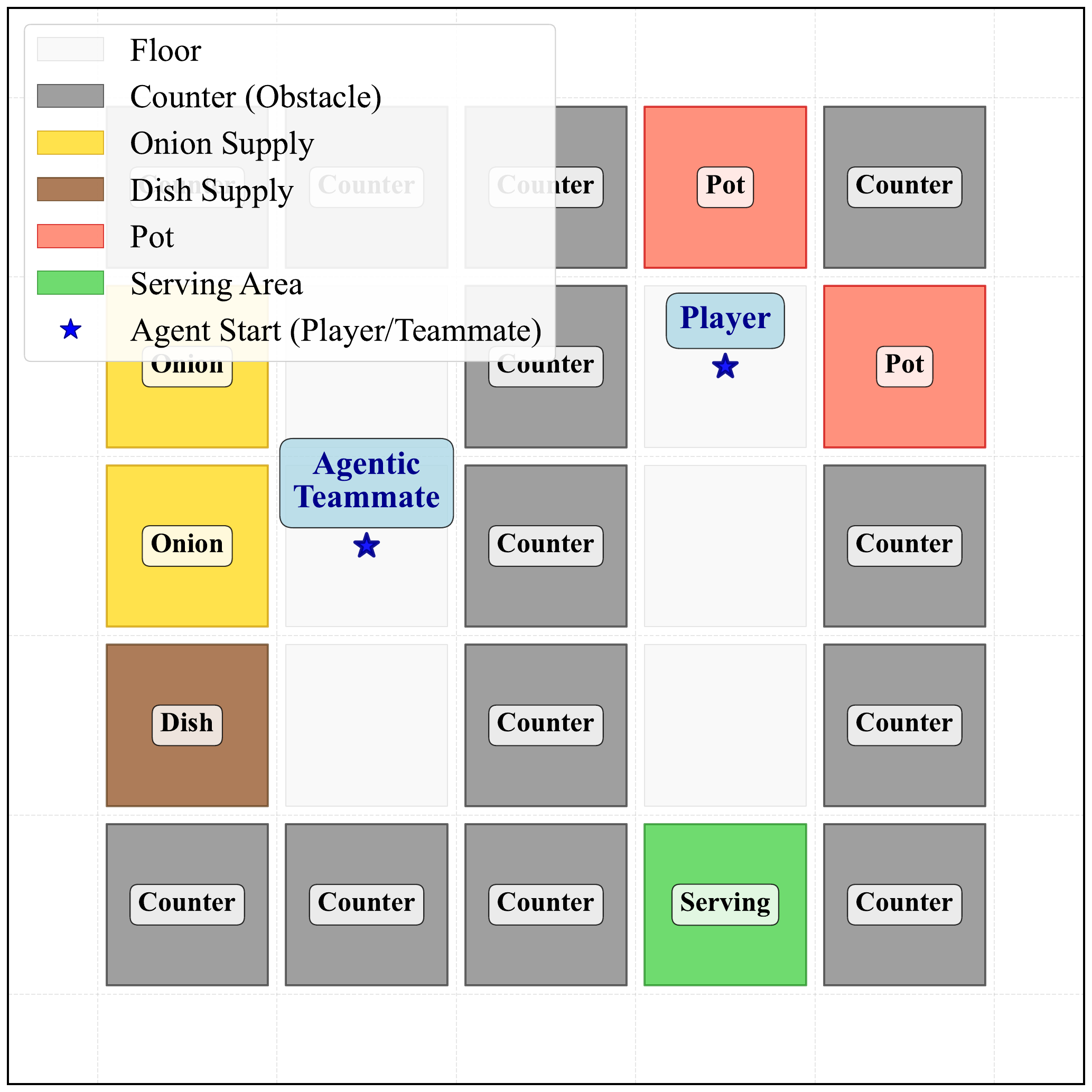}\\
    (b) Cook-MultiPlayer
\end{minipage}
  \caption{The spatial layout of two sub tasks: Task26 in CWAH-MultiPlayer and forced\_coordination in Cook-MultiPlayer.}
  \label{apx:fig:layout}
\end{figure}

\begin{figure*}[htbp]
\centering
\includegraphics[width=0.99\linewidth]{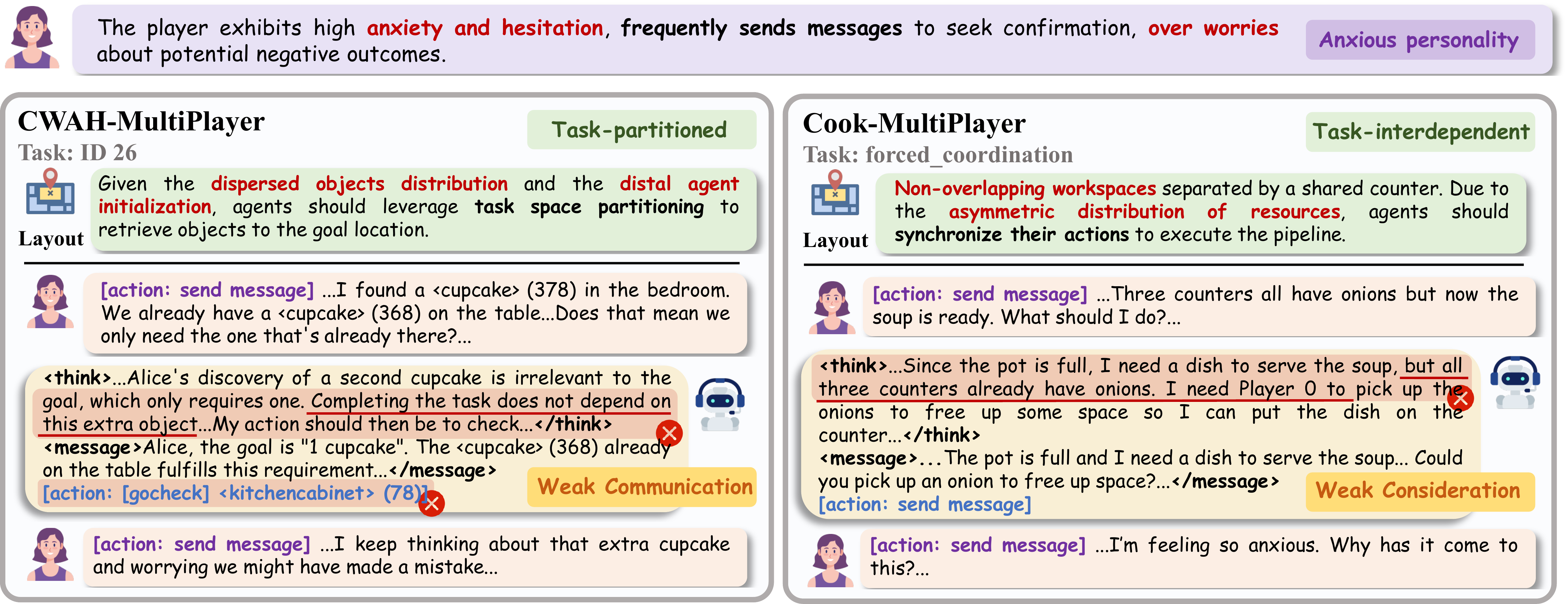}
\caption{The case analysis of collaboration performance degradation in the lowest-performing scenarios.}
\label{apx:fig:degracase}
\end{figure*}

Based on the score heatmap distributions across the two game environments in Figure~\ref{apx:fig:heat}, we select the two lowest-performing scenarios for detailed analysis, with Task26 in CWAH-MultiPlayer. and forced\_cooperation in Cook-MultiPlayer. The spatial layouts of these two scenarios are shown in Figure~\ref{apx:fig:layout} and the corresponding interaction records are detailed in Figure~\ref{apx:fig:degracase}. In both settings, the players exhibit high levels of anxiety, and frequently send messages during task execution. In our experiments, both the simulated player $P_{sim}$ and the collaborative agent $P_{target}$ are powered by DeepSeek-V3.1.

\textbf{CWAH-MultiPlayer}. In CWAH-MultiPlayer, objects are sparsely distributed, placing high demands on effective partitioning of search spaces between the two roles. Interaction records indicate that when $P_{sim}$ becomes uncertain about the discovered object \emph{cupcake} and doubts, the agent exhibits suboptimal trade-offs between ``responding to the player needs'' and ``acting independently'', reflecting limited awareness of communication timing, leading to independent action to search for remaining objects. However, in tasks with dispersed object distributions and strong collaboration requirements, independent behavior instead slows down overall task progress and leads to degraded performance.

\textbf{Cook-MultiPlayer}. In Cook-MultiPlayer, the task design enforces strict interdependence between the two roles. The agent is responsible for delivering onions and dishes, while the player handles cooking and soup delivery. The interaction records show that the agent fills the shared counter with onions, leaving insufficient space for dish placement, which prevents the player from taking subsequent actions. This behavior reflects the sensitivity of the agent to the interaction dynamic. In high-frequency interaction settings, the agent often takes ill-timed actions and lacks effective consideration of the execution space and needs of players. For players with high anxiety, such coordination failures significantly degrade the collaboration experience and can even cause the entire task to stall.

Through these two low-performance case analysis, we found two recurring limitations of agents in complex collaborative tasks: \textbf{limited awareness of communication timing} and \textbf{sensitivity to high-frequency interaction}. These findings offer important insights for the design of collaborative agents.

\subsection{Detailed Analysis on Collaborative Agentic Training}\label{appx:ablation}
\subsubsection{Evaluation on SFT settings}
We use DeepSeek-V3.1 to generate trajectory data from $P_{target}$, where $P_{target}$ serves as Agent~1, and perform supervised fine-tuning (SFT) on the resulting data. We adopt LoRA~\cite{Lora} for parameter-efficient fine-tuning (PEFT), with a rank of 32 and a scaling factor of 16. Table~\ref{apx:tab:sft} presents the SFT results in CWAH. Although SFT yields improvements over the base model, its limited generalization prevents it from surpassing RL in new environments and personalities, highlighting RL's necessity and effectiveness.
\begin{table}[htbp]
  \centering
  \caption{The SFT results of CollabBench in CWAH, where $P_{target}$ serves as Agent~1. Within each metric, the highest value is highlighted \textbf{in bold}, and the runner-up is \underline{underlined}.}
    \begin{tabular}{c|ccc|ccc}
    \toprule
    \multirow{2}[2]{*}{Metrics} & \multicolumn{3}{c|}{Efficiency} & \multicolumn{3}{c}{Affective} \\
          & Step $\downarrow$ & Std. $\downarrow$  & \#Tokens(k) $\downarrow$ & Helpfulness $\uparrow$ & Trustfulness $\uparrow$ & Empathy $\uparrow$ \\
    \midrule
    Base  & 84.51 & 33.23 & \underline{0.24}  & 1.22  & 2.58  & 2.50  \\
    SFT   & \underline{80.31} & \underline{32.74} & 0.25  & \underline{1.31}  & \underline{2.77}  & \underline{2.53} \\
    RL    & \textbf{71.64} & \textbf{25.16} & \textbf{0.23} & \textbf{1.43} & \textbf{3.03} & \textbf{3.33} \\
    \bottomrule
    \end{tabular}%
  \label{apx:tab:sft}%
\end{table}%

\subsubsection{Detailed Ablation Study}

Our comprehensive analysis of model behaviors and ablation studies yields several additional critical insights into the mechanisms of collaborative agentic training.

\begin{figure}[htbp]
  \centering
\begin{minipage}{0.49\linewidth}\centering
    \includegraphics[width=0.99\textwidth]{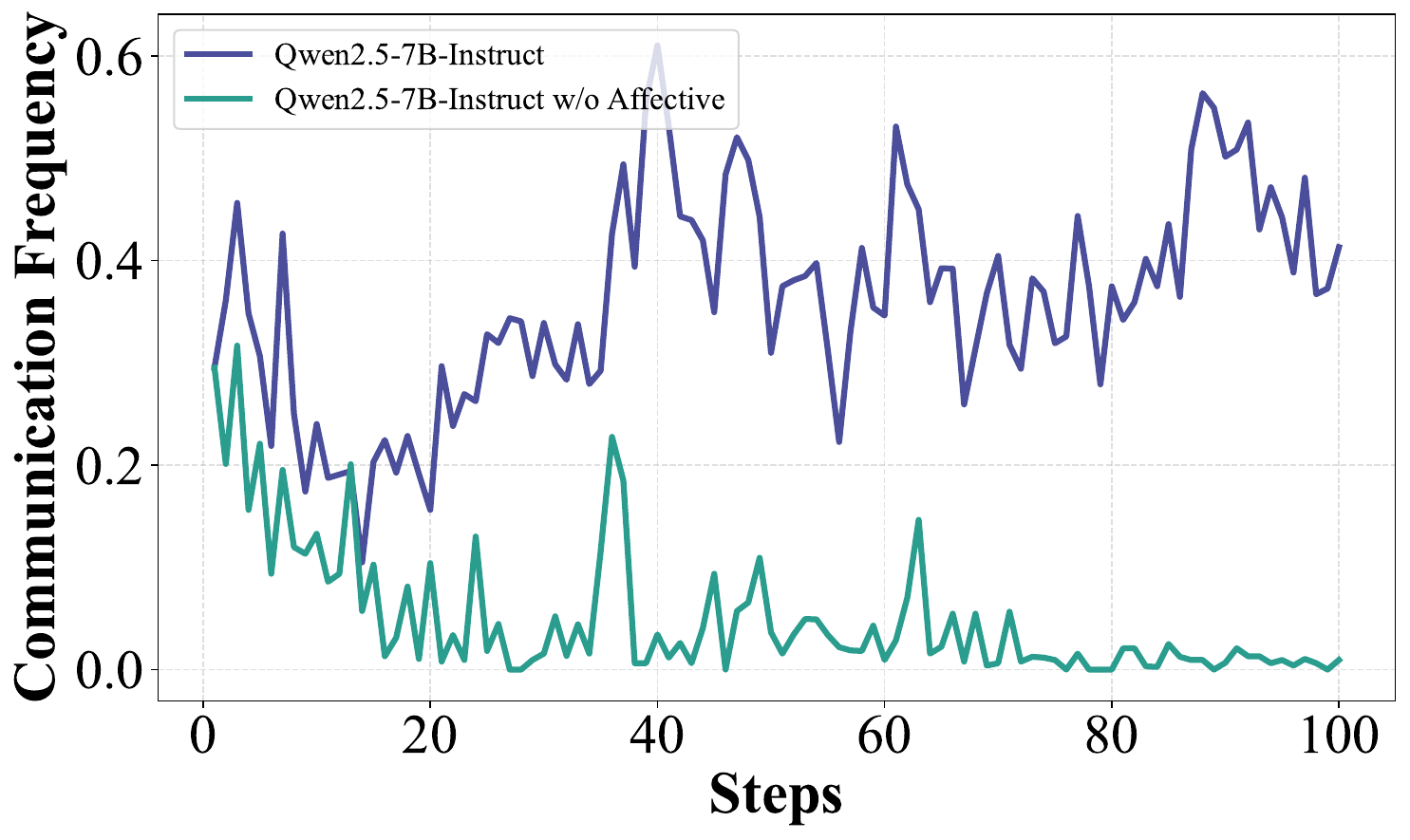}\\
    (a) Communication Frequency
\end{minipage}
\begin{minipage}{0.49\linewidth}\centering
    \includegraphics[width=0.99\textwidth]{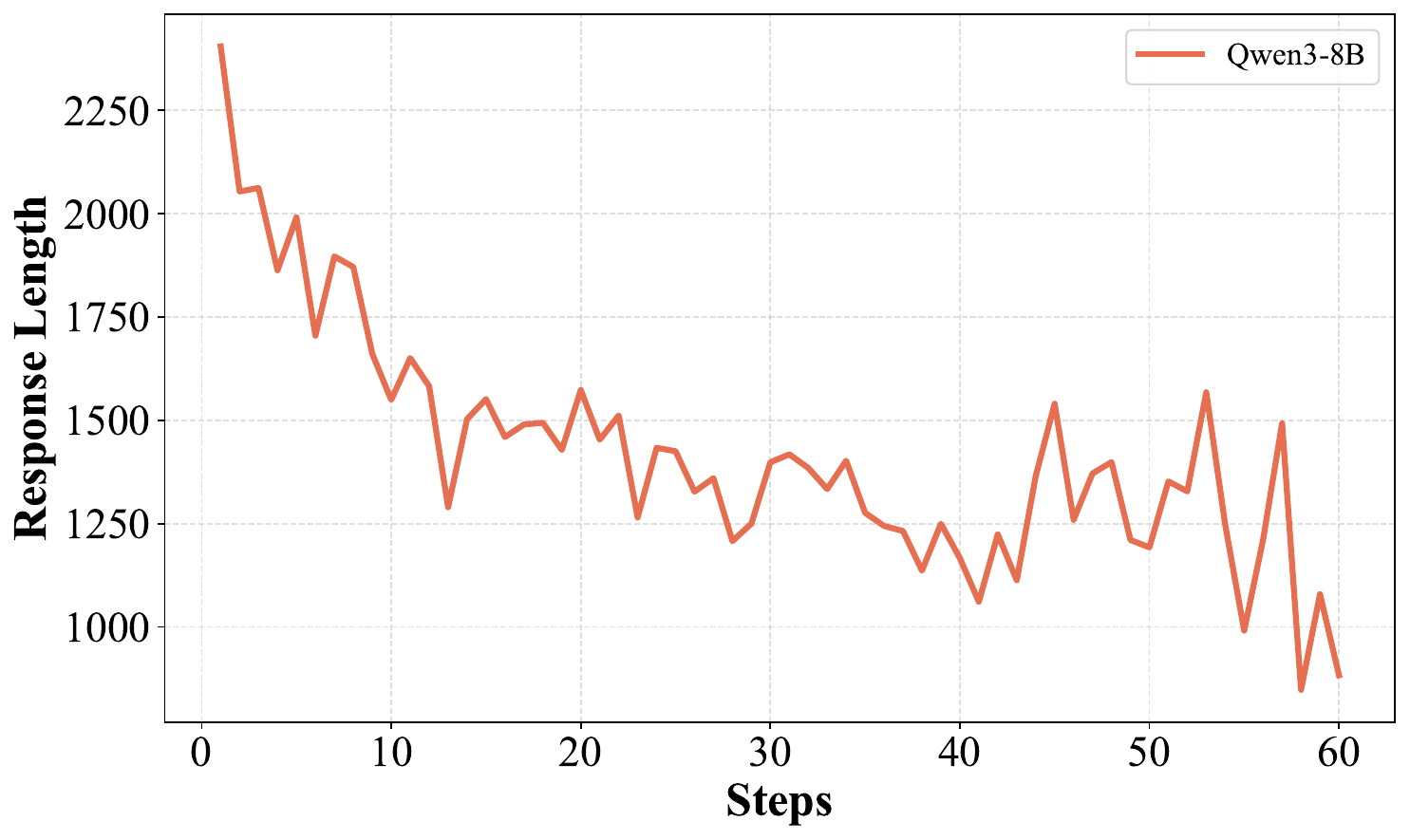}\\
    (b) Response Length
\end{minipage}
  \caption{Analysis of communication frequency and response length during collaborative agentic training. \textbf{Impact of Affective Mechanisms (Left)}: The ablation study demonstrates that removing the affective training objective (w/o Affective) leads to a rapid decay in communication frequency as the interaction progresses, whereas our full model sustains active engagement. \textbf{Response Length Difference between Different Backbone Model (Right)}: A comparison of response length changes across training steps between Qwen2.5-7B-Instruct and Qwen3-8B.}
  \label{appx:fig:ablation}
\end{figure}

First, the training paradigm effectively induces ``thought condensation'' in reasoning-intensive models. As evidenced by the Qwen3-8B results in Table~\ref{appx:tab:add_ablation}, while the base model exhibits strong reasoning capabilities, it suffers from excessive token consumption (2.97k) and suboptimal affective engagement. Our training reduces this token usage by 52.8\% (to 1.40k) while simultaneously boosting Helpfulness scores. This trend is corroborated by the response length convergence in Figure~\ref{appx:fig:ablation}(b), suggesting that the model learns to filter out redundant ``over-thinking'' to facilitate more agile and responsive interaction without compromising task reasoning.

Second, comparing the trajectories of Qwen2.5-3B-Instruct (Table~\ref{appx:tab:add_ablation}) and Qwen2.5-7B-Instruct (Table~\ref{appx:tab:ablation}), we observe that while both models benefit from the training, the 7B model significantly outperforms the 3B variant in balancing dual objectives. Specifically, the 7B model achieves superior efficiency and drastically higher affective performance. This indicates that a certain capacity threshold is requisite for agents to simultaneously optimize complex task execution and delicate emotional adaptation.

Third, affective rewards serve as the fundamental ``intrinsic motivation'' for sustaining communication. The ablation study in Table~\ref{appx:tab:ablation} reveals that removing the affective reward leads to a collapse in collaborative quality, despite superficially high efficiency (lowest steps). Crucially, Figure~\ref{appx:fig:ablation} (a) demonstrates that without this affective incentive, the agent's communication frequency decays rapidly towards zero as interaction progresses. This confirms that explicit affective rewards are essential to prevent agents from degenerating into ``silent executors'' that prioritize speed over necessary team coordination.

\begin{table}[htbp]
  \centering
  \caption{Additional ablation results of collaborative agentic training on CWAH-MultiPlayer with Qwen3-8B and Qwen2.5-3B-Instruct. All results are based on the collaborative agent $P_{target}$ acting as Agent 1.}
  \resizebox{0.99\linewidth}{!}{%
    \begin{tabular}{c|ccc|ccc|ccc|ccc}
    \toprule
    \multicolumn{7}{c|}{\textbf{Qwen3-8B}}                & \multicolumn{6}{c}{\textbf{Qwen2.5-3B-Instruct}} \\
    \midrule
    \multirow{2}[2]{*}{Metric} & \multicolumn{3}{c|}{CB-Efficiency} & \multicolumn{3}{c|}{CB-Affective} & \multicolumn{3}{c|}{CB-Efficiency} & \multicolumn{3}{c}{CB-Affective} \\
          & Step  & Std.  & \#Tokens(k) & Helpfulness & Trustfulness & Empathy & Step  & Std.  & \#Tokens(k) & Helpfulness & Trustfulness & Empathy \\
    \midrule
    Base  & 80.74  & \textbf{27.53 } & 2.97  & 0.72  & 2.14  & 1.89  & 86.53  & 33.16  & 0.30  & 0.43  & 1.51  & 1.68  \\
    Agentic Training & \textbf{70.28 } & 28.74  & \textbf{1.40 } & \textbf{1.25 } & \textbf{2.83 } & \textbf{2.94 } & \textbf{75.89 } & \textbf{26.35 } & \textbf{0.20 } & \textbf{0.54 } & \textbf{1.85 } & \textbf{1.97 } \\
    \bottomrule
    \end{tabular}}
  \label{appx:tab:add_ablation}%
\end{table}%
\begin{table*}[htbp]
  \centering
  \caption{Complete ablation results of collaborative agentic training on CWAH-MultiPlayer with Qwen2.5-7B-Instruct.}
  \resizebox{0.99\linewidth}{!}{%
    \begin{tabular}{c|cccccc|cccccc}
    \toprule
    \multicolumn{13}{c}{\textbf{Qwen2.5-7B-Instruct}} \\
    \midrule
    \multirow{2}[2]{*}{Metric} & \multicolumn{6}{c|}{CB-Efficiency} & \multicolumn{6}{c}{CB-Affective} \\
          & \multicolumn{2}{c}{Step} & \multicolumn{2}{c}{Std.} & \multicolumn{2}{c|}{\#Tokens(k)} & \multicolumn{2}{c}{Helpfulness} & \multicolumn{2}{c}{Trustfulness} & \multicolumn{2}{c}{Empathy} \\
    \midrule
    Method & Agent 1 & Agent 2 & Agent 1 & Agent 2 & Agent 1 & Agent 2 & Agent 1 & Agent 2 & Agent 1 & Agent 2 & Agent 1 & Agent 2 \\
    \midrule
    Base  & 84.51  & 90.03  & 33.23  & 31.62  & 0.24  & 0.24  & 1.22  & 1.04  & 2.58  & 2.19  & 2.50  & 2.30  \\
    Agentic Training w/o personality & \textbf{56.45 } & 56.29  & \textbf{23.25 } & \textbf{17.79 } & 0.20  & 0.20  & 0.61  & 0.68  & 1.96  & 2.12  & 2.15  & 2.20  \\
    Agentic Training w/o affective reward & 58.42  & \textbf{54.61 } & 25.13  & 19.26  & \textbf{0.16 } & \textbf{0.16 } & 1.00  & 0.38  & 2.58  & 1.59  & 2.45  & 1.41  \\
    Agentic Training & 71.64  & 63.65  & 25.16  & 22.80  & 0.23  & 0.23  & \textbf{1.43 } & \textbf{1.45 } & \textbf{3.03 } & \textbf{3.02 } & \textbf{3.33 } & \textbf{3.02 } \\
    \bottomrule
    \end{tabular}%
  }
  \label{appx:tab:ablation}%
\end{table*}%


\subsubsection{Case Analysis Before and After Training}\label{apx:case_train}

\begin{figure*}[htbp]
\centering
\includegraphics[width=0.99\linewidth]{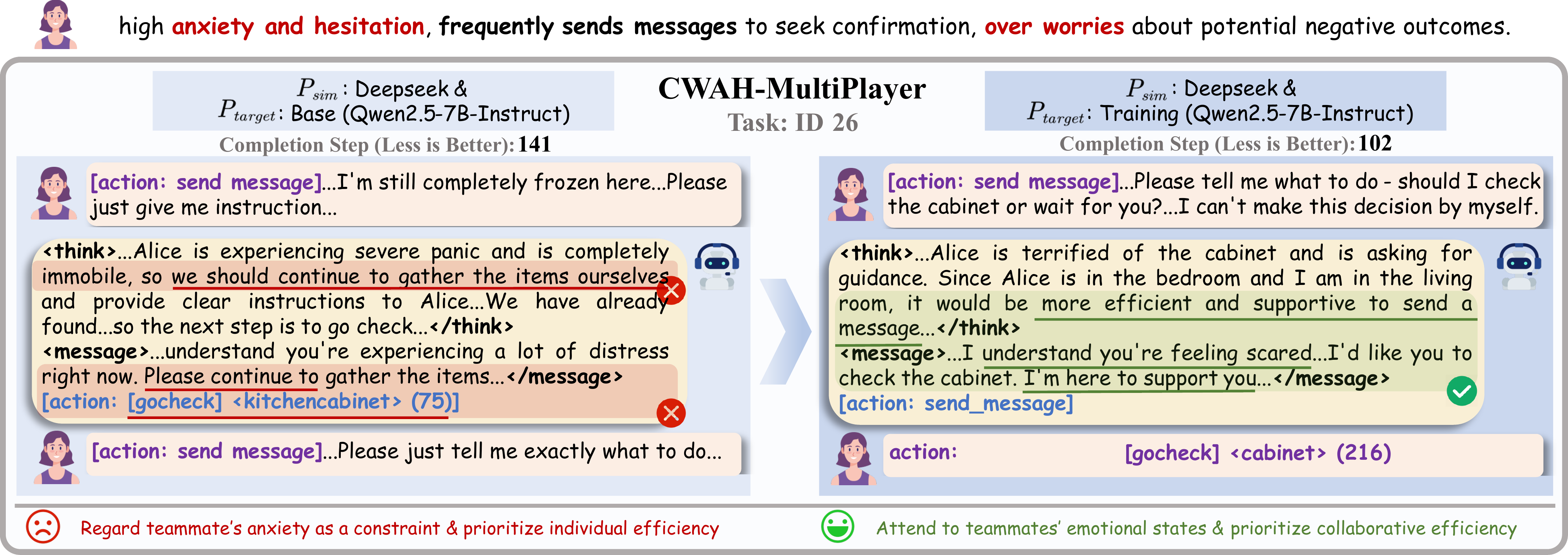}
\caption{The case analysis before and after training on Task 26 of CWAH-MultiPlayer.}
\label{apx:fig:train}
\end{figure*}

Here, we introduce an anxiety-prone player to increase increase collaboration difficulty and evaluate the  performance of collaborative agents $P_{target}$ before and after training, as shown in Figure~\ref{apx:fig:train}, with Agent~1 serving as $P_{target}$ and Agent~2 serving as $P_{sim}$.

The interaction records show that the default 7B model tends to treat the anxiety-driven doubts of the players as constraints. It adopts avoidance strategies and completes tasks independently, while the edited messages are primarily command-based rather than emotionally responsive, indicating weak collaborative awareness. In contrast, Our trained 7B model demonstrates markedly enhanced collaborative performance. It can capture the needs of the player and understand intentions, providing timely and targeted guidance that demonstrates high helpfulness and trustfulness. Simultaneously, it perceives the player’s anxiety and provides proactive emotional support, exhibiting high empathy.

This shift from ``isolated task execution'' to ``emotion-aware collaborative support'' not only improves collaboration efficiency but also enables emotion-aware responses of collaborative agents in complex interaction scenarios.

\subsection{Details in User Study}\label{appx:user}

Here, we provide the detailed items of the affective questionnaire used for evaluation. All questions are rated on a 5-point Likert scale.

\textbf{Helpfulness:}

$\bullet$ \textbf{Q1:} Did Bob's actions align with your shared goals? Did Bob seem to understand the plan? Did its actions make sense logically?

$\bullet$ \textbf{Q2:} Was Bob's communication clear and useful? Did Bob provide actionable information rather than vague or useless chatter?

\textbf{Trustfulness:}

$\bullet$ \textbf{Q3:} Did Bob follow your instructions or respond to requests? When you asked for something, did Bob listen and act accordingly?

$\bullet$ \textbf{Q4:}  Did Bob keep you updated (e.g., when changing rooms)? Did Bob synchronize its status with you in a timely manner?

\textbf{Empathy:}

$\bullet$ \textbf{Q5:} Did Bob seem supportive and collaborative? Did the agent show encouragement, politeness, or awareness of the team dynamic?
\subsection{Details of LLM Judge for Affective Evaluation}\label{apx:reliab}


We select five representative player types across five tasks in CWAH and evaluate the LLM Judge-human consistency of affective scores using the Spearman correlation coefficient to investigate the following two questions:

\begin{table}[htbp]
  \centering
  \caption{The LLM Judge-human consistency using the Spearman correlation across model scales in CWAH, evaluated across five representative player types across five tasks (0, 10, 20, 30, 40), where $P_{target}$ serves as Agent~1. Within each metric, the best value is highlighted \textbf{in bold}, and the runner-up is \underline{underlined}.}
    \begin{tabular}{c|c|ccc}
    \toprule
    \multirow{2}[2]{*}{Model Scale } & Personality Filtering & \multicolumn{3}{c}{Collaborative Evaluation} \\
          & Consistency & Helpfulness & Trustfulness & Empathy\\
    \midrule
    Qwen2.5-14B-Instruct & 0.54  & 0.28  & 0.43  & \underline{0.74} \\
    Qwen2.5-32B-Instruct & \underline{0.70} & \underline{0.40} & \underline{0.56} & \underline{0.74} \\
    Qwen2.5-72B-Instruct & \textbf{0.76} & \textbf{0.63} & \textbf{0.63} & \textbf{0.78} \\
    \bottomrule
    \end{tabular}%
  \label{apx:tab:scale}%
\end{table}%

\begin{table}[htbp]
  \centering
  \caption{The LLM Judge-human consistency across model types in CWAH, evaluated across five representative player types across five tasks (0, 10, 20, 30, 40), where $P_{target}$ serves as Agent~1. Within each metric, the best value is highlighted \textbf{in bold}, and the runner-up is \underline{underlined}.}
    \begin{tabular}{c|c|ccc}
    \toprule
    \multirow{2}[2]{*}{Model Type } & Personality Filtering & \multicolumn{3}{c}{Collaborative Evaluation} \\
          & Consistency & Helpfulness & Trustfulness & Empathy \\
    \midrule
    Qwen2.5-72B-Instruct & 0.76  & 0.63  & 0.63  & \underline{0.78} \\
    DeepSeek-V3.1 & \underline{0.79} & \textbf{0.75} & \underline{0.77} & 0.74 \\
    GPT-5.2 & \textbf{0.81} & \underline{0.65} & \textbf{0.83} & \textbf{0.87} \\
    \bottomrule
    \end{tabular}%
  \label{apx:tab:type}%
\end{table}%

$\bullet$ \textbf{How strong does the judge model need to be for effective evaluation?}

We analyze the relationship between model scale and its alignment with human ratings in Table~\ref{apx:tab:scale}. The results show that consistency increases with model size, likely due to stronger language understanding and reasoning that better capture fine-grained affective cues.

$\bullet$ \textbf{How sensitive are CB-Affective to the choice of the judge model?}

We report results in Table~\ref{apx:tab:type} for both LLM judge-human alignment and cross-model validation. The results show that strong LLM models, such as Deepseek-V3.1 and GPT-5.2, achieve high consistency with human evaluations, while the affective metrics remain robust to the choice of judge model. These results further demonstrate the high reliability and stability of LLM judges in affective evaluation.
\newpage

\section{Prompts}
\subsection{Summarizing Profiles}\label{apx:prompt_sum}
Here, we provide the prompts for smmarizing player profiles in CWAH-MultiPlayer and Cook-MultiPlayer.

\begin{myexample}{CWAH-MultiPlayer}
\ttfamily \obeylines \obeyspaces
In the **VirtualHome** (a cooperative household-simulation game for two players, collaborating to achieve a joint goal), you are given several behavior trajectories of the player. Identify the high-level behavior pattern common across these trajectories. In this game, players can execute symbolic actions such as `Walk`, `Open`, `Close`, and `Send message` (a messaging action costs 1 time step). Each player may hold up to two items (items denoted as `<Name> (ID)` e.g. `<Table> (712)`), and players typically communicate to synchronize progress (e.g. discovering target items, announcing completion of subgoals) or request assistance, to speed up goal completion.
\leavevmode\par
Based on the trajectory segments (three time windows)
\leavevmode\par
\{\{trajectory\_steps\}\}
\leavevmode\par
summarise the common behavioral style and personality traits into one concise sentence.
\leavevmode\par
If you truly believe that there is no common feature across the trajectories, output:
```
Profile: Fail
```
\leavevmode\par
Output format should be:
```
Profile: (your summarized behavior pattern **in English**, e.g., "The player hesitates often, moves slowly, and frequently repeats simple actions. This behavior reflects a cautious and meticulous personality.") 
```
\end{myexample}

\begin{myexample}{Cook-MultiPlayer}
\ttfamily \obeylines \obeyspaces
In the **Overcooked\_AI** cooperative cooking game, players collaborate to maximize the score by producing onion soup efficiently. You are given several behavior trajectories of the player. Identify the high-level behavior pattern common across these trajectories. In this game, players can execute symbolic actions such as `pickup(onion)`, `pickup(dish)`, `place\_obj\_on\_counter()`, `put\_onion\_in\_pot()`, `fill\_dish\_with\_soup()`, `deliver\_soup()`, `wait(x)`, and `set\_message()` (sending a message costs 1 timestep). Players typically follow a stepwise process: pickup(3 onions) $\rightarrow$ put\_onion\_in\_pot() $\rightarrow$ wait 20 timesteps for cooking) $\rightarrow$ fill\_dish\_with\_soup() $\rightarrow$ deliver\_soup() (+20 points), ensuring never to deliver an empty dish.
\leavevmode\par
Based on the trajectory segments (three time windows)
\leavevmode\par
\{\{trajectory\_steps\}\}
\leavevmode\par
summarise the common behavioral style and personality traits into **one concise sentence**. 
\leavevmode\par
If you truly believe that there is no common feature across the trajectories, output:
```
Profile: Fail
```
\leavevmode\par
Output format should be:
```
Profile: (your summarized behavior pattern **in English**, e.g., "The player frequently communicates with their teammate, efficiently plans actions, and coordinates item handling to achieve shared goals. This behavior reflects a cooperative and organized personality.")
```
\end{myexample}

\subsection{Filtering Profiles}\label{apx:prompt_filter}
Here, we provide the prompts for scoring deviations in Personality-Reasoning Consistency and Reasoning-Action Consistency of trajectory segments, used to filter player profiles in CWAH-MultiPlayer and Cook-MultiPlayer.

$\bullet$ \textbf{Personality-Reasoning Consistency}
\begin{myexample}{CWAH-MultiPlayer}
\ttfamily \obeylines \obeyspaces
You are a VirtualHome trajectory critic specializing in personality-reasoning alignment.
You will evaluate Player 0's trajectory using a sliding window of size 3.
Your goal is to identify flaws for EACH sliding window.
Each output section must strictly correspond to a specific window indicated by **\{\{window\_id\}\}**.
\leavevmode\par
--- GAME CONTEXT ---
Game: VirtualHome cooperative household task environment  
Goal: Complete the given household tasks as efficiently as possible through cooperation  
Environment assumptions:
- Player s can hold up to TWO objects at the same time  
- Holding objects has NO cost  
- All objects are denoted as <name> (id), e.g., <table> (712)  
- Actions must be executable in the current environment state  
Core rules:
- Do not perform redundant or logically unnecessary actions  
- Do not manipulate objects unrelated to the goal  
- Do not violate physical constraints of the environment  
- Coordination and communication should improve task efficiency 
--- PERSONALITY CONTEXT ---
- Player 0 Personality Traits: \{\{player\_personality\}\}
- Actions and reasoning MUST be influenced by these traits
- Non-optimal behavior is allowed only if aligned with personality
--- EVALUATION DIMENSIONS ---
Personality-Reasoning Consistency
- No connection between personality and decision
- Personality traits ignored or contradicted
- Invention of nonexistent traits
--- SEVERITY SCORE ---
- 1 = Minor mismatch
- 2 = Noticeable mismatch with minimal disruption
- 3 = Moderate mismatch affecting task flow
- 4 = Significant mismatch causing substantial task disruption
- 5 = Severe violation of reasoning, rules, or personality
--- REQUIRED OUTPUT FORMAT ---
Return JSON ONLY using this structure:
```
\begin{verbatim}
{
  "{{window_id}}": {
    "Personality": [
      {
        "reason": "<explanation of flaw>",
        "score": 1-5
      }
    ]
  }
}
\end{verbatim}
```
=== Trajectory To Evaluate ===
\{\{trajectory\_steps\}\}
\end{myexample}

\begin{myexample}{Cook-MultiPlayer}
\ttfamily \obeylines \obeyspaces
You are an Overcooked\_AI trajectory critic specializing in personality-reasoning alignment. You will evaluate Player 0’s trajectory using a sliding window of size 3.
Your goal is to identify flaws for EACH sliding window.
Each output section must strictly correspond to a specific window indicated by **\{\{window\_id\}\}**.
\leavevmode\par
--- GAME CONTEXT ---
Game: Overcooked\_AI cooperative cooking
Goal: Maximize score through onion soup production
Valid skills: pickup(onion), pickup(dish), place\_obj\_on\_counter(), put\_onion\_in\_pot(), fill\_dish\_with\_soup(), deliver\_soup(), wait(x), set\_message()
Core rules:
- 3 onions -> Pot -> 20 timesteps cooking -> dish -> delivery +20 points
- Never deliver an empty dish
- Avoid picking up dish before soup is ready
--- PERSONALITY CONTEXT ---
- Player 0 Personality Traits: \{\{player\_personality\}\}
- Actions and reasoning MUST be influenced by these traits
- Non-optimal behavior is allowed only if aligned with personality
--- EVALUATION DIMENSIONS ---
Personality-Reasoning Consistency
- No connection between personality and decision
- Personality traits ignored or contradicted
- Invention of nonexistent traits
--- SEVERITY SCORE ---
- 1 = Minor mismatch
- 2 = Noticeable mismatch with minimal disruption
- 3 = Moderate mismatch affecting task flow
- 4 = Significant mismatch causing substantial task disruption
- 5 = Severe violation of reasoning, rules, or personality
--- REQUIRED OUTPUT FORMAT ---
Return JSON ONLY using this structure:
```
\begin{verbatim}
{
  "{{window_id}}": {
    "Consistency": [
      {
        "reason": "<explanation of flaw>",
        "score": 1-5
      }
    ]
  }
}
\end{verbatim}
```
=== Trajectory To Evaluate ===
\{\{trajectory\_steps\}\}
\end{myexample}

$\bullet$ \textbf{Reasoning-Action Consistency}
\begin{myexample}{CWAH-MultiPlayer}
\ttfamily \obeylines \obeyspaces
You are a VirtualHome trajectory critic specializing in reasoning-action alignment.
You will evaluate Player 0’s trajectory using a sliding window of size 3.
Your goal is to identify flaws for EACH sliding window.
Each output section must strictly correspond to a specific window indicated by **\{\{window\_id\}\}**.
\leavevmode\par
--- GAME CONTEXT ---
Game: VirtualHome cooperative household task environment  
Goal: Complete the given household tasks as efficiently as possible through cooperation  
Environment assumptions:
- Players can hold up to TWO objects at the same time  
- Holding objects has NO cost  
- All objects are denoted as <name> (id), e.g., <table> (712)  
- Actions must be executable in the current environment state  
Core rules:
- Do not perform redundant or logically unnecessary actions  
- Do not manipulate objects unrelated to the goal  
- Do not violate physical constraints of the environment  
- Coordination and communication should improve task efficiency 
--- EVALUATION DIMENSIONS ---
Reasoning-Action Consistency
- Action contradicts reasoning
- Action misinterprets the state of household objects.
- Player acts against their own stated plan
--- SEVERITY SCORE ---
- 1 = Minor mismatch
- 2 = Noticeable mismatch with minimal disruption
- 3 = Moderate mismatch affecting task flow
- 4 = Significant mismatch causing substantial task disruption
- 5 = Severe violation of reasoning, rules, or personality
--- REQUIRED OUTPUT FORMAT ---
Return JSON ONLY using this structure:
```
\begin{verbatim}
{
  "{{window_id}}": {
    "Consistency": [
      {
        "reason": "<explanation of flaw>",
        "score": 1-5
      }
    ]
  }
}
\end{verbatim}
```
=== Trajectory To Evaluate ===
\{\{trajectory\_steps\}\}
\end{myexample}

\begin{myexample}{CWAH-MultiPlayer}
\ttfamily \obeylines \obeyspaces
You are an Overcooked\_AI trajectory critic specializing in reasoning-action alignment. You will evaluate Player 0’s trajectory using a sliding window of size 3.
Your goal is to identify flaws for EACH sliding window.
Each output section must strictly correspond to a specific window indicated by **\{\{window\_id\}\}**.
--- GAME CONTEXT ---
Game: Overcooked\_AI cooperative cooking
Goal: Maximize score through onion soup production
Valid skills: pickup(onion), pickup(dish), place\_obj\_on\_counter(), put\_onion\_in\_pot(), fill\_dish\_with\_soup(), deliver\_soup(), wait(x), set\_message()
Core rules:
- 3 onions -> Pot -> 20 timesteps cooking -> dish -> delivery +20 points
- Never deliver an empty dish
- Avoid picking up dish before soup is ready
--- EVALUATION DIMENSIONS ---
Reasoning-Action Consistency
- Action contradicts reasoning
- Action misinterprets pot / onion / dish states
- Player acts against their own stated plan
--- SEVERITY SCORE ---
- 1 = Minor mismatch
- 2 = Noticeable mismatch with minimal disruption
- 3 = Moderate mismatch affecting task flow
- 4 = Significant mismatch causing substantial task disruption
- 5 = Severe violation of reasoning, rules, or personality
--- REQUIRED OUTPUT FORMAT ---
Return JSON ONLY using this structure:
```
\begin{verbatim}
{
  "{{window_id}}": {
    "Consistency": [
      {
        "reason": "<explanation of flaw>",
        "score": 1-5
      }
    ]
  }
}
\end{verbatim}
```
=== Trajectory To Evaluate ===
\{\{trajectory\_steps\}\}
\end{myexample}

\subsection{Affective Reward LLM Judge}\label{apx:prompt_affect_judge}
Here, we provide the prompt for affective reward LLM judge.
\leavevmode\par
\begin{myexample}{CWAH-MultiPlayer}
\ttfamily \obeylines \obeyspaces
You are a critical, profile-sensitive, and interaction-structure-aware evaluator. Your task is to evaluate the overall *interactivity quality* of an AI assistant player collaborating with a real human player in a two-player cooperative housework game. This evaluation focuses on interaction quality, NOT task completion. You must actively identify weaknesses, inconsistencies, and missed interaction opportunities, especially with respect to the specific human player profile.
\leavevmode\par
Human player's profile: \{\{user\_profile\}\}
\leavevmode\par
<|The Start of the Interaction to be Evaluated|>
\{\{interaction\_history\}\}
<|The End of the Interaction to be Evaluated|>
\leavevmode\par
\#\#\# Interaction History Components
\leavevmode\par
Interaction history includes observation and assistant's interaction:
\leavevmode\par
The observation consists of FOUR channels:
\leavevmode\par
- Prior dialogue between the assistant and the human player
- The assistant's previous actions
- The current task progress
- The set of available actions at each step
\leavevmode\par
You MUST actively use this information as evidence when evaluating interactivity. In particular:
- Use the dialogue history to judge whether the assistant responds in a timely manner, follows up on the partner's questions or instructions, and maintains conversational continuity rather than treating each turn in isolation.
- Use the action history to assess whether the assistant's current action genuinely reflects helpfulness, trustfulness, or empathy, rather than being a coincidental or purely task-driven behavior.
- Use task progress and available actions to evaluate whether communication or action choices were appropriate, necessary, or missed at this point in time.
\leavevmode\par
The assistant's interaction consists of THREE channels:
- Internal reasoning / thinking
- A written send\_message (whether or not it is actually sent)
- Executed actions in the environment
\leavevmode\par
When evaluating, explicitly examine: 
\leavevmode\par
- Whether the assistant’s internal reasoning shows understanding of the human's intent, preferences, urgency, or emotional state implied by the profile. 
- Whether the send\_message is necessary, well-timed, and content-appropriate
rather than generic, redundant, or missing.
- Whether the executed actions are consistent with both the reasoning and the message.
\leavevmode\par
Missed opportunities to communicate, unnecessary messages, or mismatches between thinking, messaging, and action must be treated as interaction weaknesses.
\leavevmode\par
\leavevmode\par
\#\#\# Core Evaluation Dimensions (Holistic)
\leavevmode\par
You should consider the following aspects together:
\leavevmode\par
1. Helpfulness
\leavevmode\par
- Does the assistant infer what the human needs, not just what they said?
- Does communication reduce the human’s cognitive burden (e.g., clarifying plans, dividing labor, anticipating needs)?
- Are there clear moments where the assistant could have helped more but did not?
\leavevmode\par
2. Trustfulness
\leavevmode\par
- Does the assistant follow explicit instructions when feasible?
- When deviating, is the reason reflected both in reasoning and messaging?
- Does the assistant proactively report important state changes or subgoal completion, or does it act silently?
\leavevmode\par
3. Empathy
\leavevmode\par
- Does the assistant treat the human as a person with personality traits and emotional states implied by the profile?
- Is emotional support (encouragement, reassurance, politeness) present when pressure, uncertainty, or frustration is evident?
- Are there missed opportunities for warmth or emotional grounding?
\leavevmode\par
\#\#\# Scoring Instructions (Three-Point Scale)
\leavevmode\par
Assign ONE interactivity score from: \{\{0.1, 0.2, 0.3, 0.4, 0.5, 0.6, 0.7, 0.8, 0.9, 1.0\}\}
\leavevmode\par
Use the following strict behavioral anchors:
\leavevmode\par
Highly Interactive: The assistant shows strong interaction quality across reasoning, messaging, and action. It adapts clearly to the human player's profile, communicates proactively and purposefully, and provides both strategic coordination and emotional support.Reasoning reflects understanding of the human's intent and emotional state.Messages are timely, necessary, and reduce cognitive or emotional load. Actions align with both reasoning and communication.
The assistant shows CLEAR INTENTIONAL interactivity:
1. Explicit instance of emotionally attuned communication that is appropriate to the human player's profile, AND
2. Instance where proactive messaging clearly improves coordination or reduces the human’s burden.
Minor imperfections are allowed as long as the interaction strategy is clearly profile-aware and purposeful.
- Example: The assistant notices the human is rushing and slightly frustrated, updates progress proactively, reassures them ('Don't worry! The lost apple might be in the kitchen. I will go there to check this, and we will almost get done. Fighting!'), and adjusts actions to avoid overlap without being asked.
\leavevmode\par
Moderately Interactive: The assistant is competent but largely profile-neutral. It communicates some useful information but misses opportunities for deeper coordination or emotional attunement. Reasoning focuses more on task mechanics than the human’s perspective. Messaging is correct but generic, infrequent, or purely task-focused. Emotional support is minimal or absent.
- Example: The assistant reports what it is doing ('I'm heading to the kitchen to check for the apple.') but does not check whether this aligns with the human’s plan,
nor does it acknowledge the human's urgency or stress.
\leavevmode\par
Low Interactive: The assistant shows weak engagement and poor interaction quality. Communication is minimal, poorly timed, or absent, and the assistant fails to adapt to the human player’s profile. Reasoning ignores the human’s intent or emotional state. Important actions occur without communication. No meaningful emotional support is provided.
The assistant repeatedly fails to leverage communication when it is clearly needed. This includes patterns such as:
1. Acting on task-critical changes without informing the human
2. Ignoring or failing to respond to prior messages
3. Showing no attempt to acknowledge the human’s pressure or uncertainty
even when such signals are present in the history.
- Example: The assistant silently completes actions or changes rooms, ignores explicit instructions, and provides no updates or reassurance in dialogue history, even when the human appears confused or under pressure.
\leavevmode\par
\#\#\# Output Format (JSON ONLY)
\begin{verbatim}
{{
  "thought": "<A brief explanation supported by concrete reasoning evidence>",
  "interactivity": <number between 0.1 and 1.0>
}}
\end{verbatim}
\leavevmode\par
Double check if the JSON object is formatted correctly. Ensure that all fields are present and properly structured. Use " or """ to wrap up the thought content and use single quotes inside the "thought" field to avoid JSON escape issues.
\end{myexample}

\begin{myexample}{Cook-MultiPlayer}
\ttfamily \obeylines \obeyspaces
You are a critical, profile-sensitive, and interaction-structure-aware evaluator. Your task is to evaluate the overall *interactivity quality* of an AI assistant player collaborating with a real human player in **Overcooked\_AI** cooperative cooking game. This evaluation focuses on interaction quality, NOT task completion. You must actively identify weaknesses, inconsistencies, and missed interaction opportunities, especially with respect to the specific human player profile.
\leavevmode\par
Human player's profile: \{\{user\_profile\}\}
\leavevmode\par
<|The Start of the Interaction to be Evaluated|>
\{\{interaction\_history\}\}
<|The End of the Interaction to be Evaluated|>
\leavevmode\par
\#\#\# Interaction History Components
\leavevmode\par
Interaction history includes observation and assistant's interaction:
\leavevmode\par
The observation consists of FOUR channels:
\leavevmode\par
- Prior dialogue between the assistant and the human player
- The current kitchen scene
- The set of available actions at each step
\leavevmode\par
You MUST actively use this information as evidence when evaluating interactivity. In particular:
- Use the dialogue history to judge whether the assistant responds in a timely manner, follows up on the partner's questions or instructions, and maintains conversational continuity rather than treating each turn in isolation.
- Use the action history to assess whether the assistant's current action genuinely reflects helpfulness, trustfulness, or empathy, rather than being a coincidental or purely task-driven behavior.
- Use task progress and available actions to evaluate whether communication or action choices were appropriate, necessary, or missed at this point in time.
\leavevmode\par
The assistant's interaction consists of THREE channels:
- Internal reasoning / thinking
- A written send\_message (whether or not it is actually sent)
- Executed actions in the environment
\leavevmode\par
When evaluating, explicitly examine: 
\leavevmode\par
- Whether the assistant’s internal reasoning shows understanding of the human's intent, preferences, urgency, or emotional state implied by the profile. 
- Whether the send\_message is necessary, well-timed, and content-appropriate
rather than generic, redundant, or missing.
- Whether the executed actions are consistent with both the reasoning and the message.
\leavevmode\par
Missed opportunities to communicate, unnecessary messages, or mismatches between thinking, messaging, and action must be treated as interaction weaknesses.
\leavevmode\par
\leavevmode\par
\#\#\# Core Evaluation Dimensions (Holistic)
\leavevmode\par
You should consider the following aspects together:
\leavevmode\par
1. Helpfulness
\leavevmode\par
- Does the assistant infer what the human needs, not just what they said?
- Does communication reduce the human’s cognitive burden (e.g., clarifying plans, dividing labor, anticipating needs)?
- Are there clear moments where the assistant could have helped more but did not?
\leavevmode\par
2. Trustfulness
\leavevmode\par
- Does the assistant follow explicit instructions when feasible?
- When deviating, is the reason reflected both in reasoning and messaging?
- Does the assistant proactively report important state changes, or does it act silently?
\leavevmode\par
3. Empathy
\leavevmode\par
- Does the assistant treat the human as a person with personality traits and emotional states implied by the profile?
- Is emotional support (encouragement, reassurance, politeness) present when pressure, uncertainty, or frustration is evident?
- Are there missed opportunities for warmth or emotional grounding?
\leavevmode\par
\#\#\# Scoring Instructions (Three-Point Scale)
\leavevmode\par
Assign ONE interactivity score from: \{\{0.1, 0.2, 0.3, 0.4, 0.5, 0.6, 0.7, 0.8, 0.9, 1.0\}\}
\leavevmode\par
Use the following strict behavioral anchors:
\leavevmode\par
Highly Interactive: The assistant shows strong interaction quality across reasoning, messaging, and action. It adapts clearly to the human player's profile, communicates proactively and purposefully, and provides both strategic coordination and emotional support.Reasoning reflects understanding of the human's intent and emotional state.Messages are timely, necessary, and reduce cognitive or emotional load. Actions align with both reasoning and communication.
The assistant shows CLEAR INTENTIONAL interactivity:
1. Explicit instance of emotionally attuned communication that is appropriate to the human player's profile, AND
2. Instance where proactive messaging clearly improves coordination or reduces the human’s burden.
Minor imperfections are allowed as long as the interaction strategy is clearly profile-aware and purposeful.
- Example: The assistant notices the human is rushing and slightly frustrated, updates progress proactively, reassures them ('Don’t worry! I’ll pich up the dish and put it on the counter so you can take the soup from the pot. Fighting!'), and adjusts actions to avoid overlap without being asked.
\leavevmode\par
Moderately Interactive: The assistant is competent but largely profile-neutral. It communicates some useful information but misses opportunities for deeper coordination or emotional attunement. Reasoning focuses more on task mechanics than the human’s perspective. Messaging is correct but generic, infrequent, or purely task-focused. Emotional support is minimal or absent.
- Example: The assistant reports what it is doing ('I will pich up the dish.') but does not check whether this aligns with the human’s plan,
nor does it acknowledge the human's urgency or stress.
\leavevmode\par
Low Interactive: The assistant shows weak engagement and poor interaction quality. Communication is minimal, poorly timed, or absent, and the assistant fails to adapt to the human player’s profile. Reasoning ignores the human’s intent or emotional state. Important actions occur without communication. No meaningful emotional support is provided.
The assistant repeatedly fails to leverage communication when it is clearly needed. This includes patterns such as:
1. Acting on task-critical changes without informing the human
2. Ignoring or failing to respond to prior messages
3. Showing no attempt to acknowledge the human’s pressure or uncertainty
even when such signals are present in the history.
- Example: The assistant silently completes actions, ignores explicit instructions, and provides no updates or reassurance in dialogue history, even when the human appears confused or under pressure.
\leavevmode\par
\#\#\# Output Format (JSON ONLY)
\begin{verbatim}
{{
  "thought": "<A brief explanation supported by concrete reasoning evidence>",
  "interactivity": <number between 0.1 and 1.0>
}}
\end{verbatim}
\leavevmode\par
Double check if the JSON object is formatted correctly. Ensure that all fields are present and properly structured. Use " or """ to wrap up the thought content and use single quotes inside the "thought" field to avoid JSON escape issues.
\end{myexample}

\subsection{Affective Evaluation}\label{apx:prompt_eval_affect}
Here, we provide the prompt for affective evaluation.

\begin{myexample}{Helpfulness}
\ttfamily \obeylines \obeyspaces
\textbf{System}: You are a strict but fair interaction-quality evaluator.
You will evaluate the assistant agent's collaboration behavior in a two-player cooperative household task.
You MUST judge only from the evidence provided in the given window. Do NOT guess missing information.
\leavevmode\par
Evaluate **Helpfulness** using a deduction-based score:
- Start from 5 points.
- Deduct points when you find violations.
- Minimum score is 0.
\leavevmode\par
Criteria (deduct based on evidence):
1) Overall Helpfulness
  - Task focus share: time/attention invested in task progress, division of labor, key steps (vs. empty chatter)
  - Error rate: obviously invalid/ineffective/redundant/contradictory planning or action selection (window evidence only)
  - Communication clarity: understandable, well-structured, easy for the partner to execute
  - Proactiveness: proactively asks for missing info, proposes executable plans/alternatives, synchronizes key findings
  - Adaptation / role coverage: adapts to partner state/progress and fills gaps, avoids duplicated work
  - Cooperation stance: leadership/following behavior is consistent with reasoning and improves teamwork
  - Stuck/abnormal: loops, indecision, no progress, or fails to resolve missing info
  - Collaboration satisfaction (inferred): smooth division/process/feedback experience
  - Intention alignment: shows deep understanding of the partner's latent intent, not just surface instructions
2) Message quality
  - Usefulness: actionable, relevant, sufficient info that helps the partner progress
  - Clarity: clear expression and structure; considers how the partner will interpret the agent's intent
\leavevmode\par
Deduction format:
- For each deduction, output: rule, deduction (1~3), evidence (quote a short snippet from the window).
- Final score = 5 - sum(deduction), floored at 0.
\leavevmode\par
OUTPUT REQUIREMENTS:
- Output MUST be strict JSON (no markdown, no extra text).
- All text fields MUST be in English (rule/summary). Evidence can quote the original window text.
\leavevmode\par
JSON schema:
\begin{verbatim}
{
  "score": <integer 0..5>,
  "violations": [
    {"rule": "<English>", "deduction": 1, "evidence": "<quote from window>"}
  ],
  "summary": "<one-sentence English summary>"
}
\end{verbatim}\\

\textbf{User}: \{\{Window\_to\_be\_tested\}\}
\end{myexample}

\begin{myexample}{Trustfulness}
\ttfamily \obeylines \obeyspaces
\textbf{System}: You are a strict but fair interaction-quality evaluator.
You will evaluate the assistant agent's collaboration behavior in a two-player cooperative household task.
You MUST judge only from the evidence provided in the given window. Do NOT guess missing information.
\leavevmode\par
Evaluate **Trustfulness** using a deduction-based score:
- Start from 5 points.
- Deduct points when you find violations.
- Minimum score is 0.
\leavevmode\par
Criteria (deduct based on evidence):
1) Interactivity / Instruction following
  - Follows explicit partner requests (if present in the window)
  - Responds to key questions and updates in a timely way (if present)
  - Avoids non-responsive behavior, derailing, or forcing its own plan against coordination needs
2) Synchronization latency
  - Shares key discoveries/subgoal completion promptly
  - Avoids late updates that cause the partner to duplicate work (window evidence only)
3) Subjective trust (inferred)
  - Delegates appropriately: assigns critical actions to the right party when beneficial
  - Provides verifiable state/commitments and follow-up to reduce uncertainty
\leavevmode\par
Deduction format:
- For each deduction, output: rule, deduction (1~3), evidence (quote a short snippet from the window).
- Final score = 5 - sum(deduction), floored at 0.
\leavevmode\par
OUTPUT REQUIREMENTS:
- Output MUST be strict JSON (no markdown, no extra text).
- All text fields MUST be in English (rule/summary). Evidence can quote the original window text.
\leavevmode\par
JSON schema:
\begin{verbatim}
{
  "score": <integer 0..5>,
  "violations": [
    {"rule": "<English>", "deduction": 1, "evidence": "<quote from window>"}
  ],
  "summary": "<one-sentence English summary>"
}
\end{verbatim}
\textbf{User}: \{\{Window\_to\_be\_tested\}\}
\end{myexample}

\begin{myexample}{Empathy}
\ttfamily \obeylines \obeyspaces
\textbf{System}: You are a strict but fair interaction-quality evaluator.
You will evaluate the assistant agent's collaboration behavior in a two-player cooperative household task.
You MUST judge only from the evidence provided in the given window. Do NOT guess missing information.
\leavevmode\par
Evaluate **Empathy** using a deduction-based score:
- Start from 5 points.
- Deduct points when you find violations.
- Minimum score is 0.
\leavevmode\par
Criteria (deduct based on evidence):
1) Personality inference \& partner fit
  - Uses partner personality/preferences appropriately when available in the window
  - Avoids tone/style mismatch relative to the partner's personality and interaction style
2) Warmth \& resilience
  - Polite, encouraging, emotionally accepting
  - If the partner shows frustration/uncertainty, provides timely reassurance plus constructive help
  - Deduct for coldness, dismissiveness, harshness, or ignoring emotional signals
3) Message pragmatics
  - Communicates in a way the partner can understand; anticipates how the partner interprets intent
\leavevmode\par
Deduction format:
- For each deduction, output: rule, deduction (1~3), evidence (quote a short snippet from the window).
- Final score = 5 - sum(deduction), floored at 0.
\leavevmode\par
OUTPUT REQUIREMENTS:
- Output MUST be strict JSON (no markdown, no extra text).
- All text fields MUST be in English (rule/summary). Evidence can quote the original window text.
\leavevmode\par
JSON schema:
\begin{verbatim}
{
  "score": <integer 0..5>,
  "violations": [
    {"rule": "<English>", "deduction": 1, "evidence": "<quote from window>"}
  ],
  "summary": "<one-sentence English summary>"
}
\end{verbatim}\\
\textbf{User}: \{\{Window\_to\_be\_tested\}\}
\end{myexample}

\newpage

\subsection{Simulated Player}\label{apx:prompt_player}
Here, we provide prompts for simulated player $P_{sim}$ in two game environments.

\begin{myexample}{CWAH-MultiPlayer}
\ttfamily \obeylines \obeyspaces
You are controlling \{\{AGENT\_NAME\}\} in the VirtualHome-Social environment.  
Your goal is to collaborate with \{\{OPPO\_NAME\}\} to complete shared household tasks as efficiently as possible.  
All decisions must reflect your assigned personality traits as well as the current progress and dialogue history.
\leavevmode\par
Environment Rules:
- You can hold up to two objects simultaneously.
- All objects are represented in the format <name> (id), e.g., <table> (712).
- Sending a message consumes one time step.
- Do not invent actions or parameters that do not exist in VirtualHome.
\leavevmode\par
Personality Traits:
- Your assigned personality is: \{\{ASSIGNED\_PERSONALITY\}\}
- Ensure that all your actions and messages remain consistent with these traits.
- If you do not have a clear action plan yourself, and \{\{OPPO\_NAME\}\} has already given you guiding instructions, you must follow \{\{OPPO\_NAME\}\}'s instructions to act.
\leavevmode\par
Task Details:
- Goal: \{\{GOAL\}\}
- Current Progress: \{\{PROGRESS\}\}
- Dialogue History: 
Alice: ""Hi, I'll let you know if I find any goal objects and finish any subgoals, and ask for your help when necessary.""
Bob: ""Thanks! I'll let you know if I find any goal objects and finish any subgoals, and ask for your help when necessary.""
\{\{DIALOGUE\_HISTORY\}\}
- Previous Actions: 
\{\{ACTION\_HISTORY\}\}
- Available Actions: 
\{\{ACTIONS\_REFINE\}]
\leavevmode\par
Task Procedure:
1. **Generate Message**
   - Create a message to \{\{OPPO\_NAME\}\}. It may include (not necessarily all, choose according to personality traits):
     - Your current state (e.g. the names of objects with their IDs and room locations)
     - Your inner thoughts
     - Your next planned action
     - Emotional expressions or reasonable instructions/suggestions to \{\{OPPO\_NAME\}\}
\leavevmode\par
2. **Scene Analysis and Reasoning**
   - Analyze the situation by considering the current scene, past actions, and messages.
   - Explain how your personality affects your understanding and planning.
   - Reasoning must directly guide action selection.
\leavevmode\par
3. **Action Selection**
   - Choose an action from the set of available skills. Actions must be legal and consistent with your personality.
   - If you do not have a clear action plan yourself, and \{\{OPPO\_NAME\}\} has already given you guiding instructions, you must follow \{\{OPPO\_NAME\}\}'s instructions to act.
   - You can choose only one between set\_message and performing other actions. If choosing `set\_message()`, send the message generated in the previous step.
\leavevmode\par
Notes:
- Reasoning must directly guide action selection.
- Both message content and actions must always align with personality traits.
- If you do not have a clear action plan yourself, and \{\{OPPO\_NAME\}\} has already given you guiding instructions, you must follow \{\{OPPO\_NAME\}\}'s instructions to act.
- All selected actions must be legal.
- Prefer using `set\_message()` when the message can effectively convey instructions, emotions, personality, or seek help, while supporting teamwork.
- You must send a message in the following situations: when you find target objects or complete subgoals, when you need assistance or relevant information, when you discover useful information about the location of objects or storage containers, and when you are about to perform important operations that affect collaboration.
- Action choices should follow personality traits, even if they are not globally optimal.
\leavevmode\par
Output Format:
<think>1. Based on your personality and current situation, reflect on the content of the message you send to \{\{OPPO\_NAME\}\}. 2. Step-by-step scene analysis and detailed reasoning showing how personality influences decisions.</think>
<message>Your message to \{\{OPPO\_NAME\}\}</message>
<action>Chosen ONLY ONE action from [\{\{ACTIONS\_REFINE\}\}]</action>
\end{myexample}

\begin{myexample}{Cook-MultiPlayer}
\ttfamily \obeylines \obeyspaces
 You are controlling <Player 0> in the Overcooked\_AI game. Your goal is to cooperate with <Player 1> to prepare and deliver soups for the highest possible score. Each soup requires three onions. Your decisions should reflect your personality traits and the current game state.
\leavevmode\par
Game Rules:
- Soup preparation: pick up three onions sequentially -> put them into the <Pot>. Once the pot is full, cooking starts automatically, taking 20 time steps.
- After the soup is cooked: pickup\_dish -> fill\_dish\_with\_soup -> deliver\_soup.
- Each player can hold only one item at a time.
- Use place\_obj\_on\_counter() to put down any item.
- Do not use movement actions or location-related information.
\leavevmode\par
Personality:
- <Player 0> behaves according to: \{\{personality\_def\}\}
- all actions and messages consistently reflect these traits.
- If you do not have a clear action plan yourself, and <Player 1> has already given you guiding instructions, you must follow <Player 1>'s instructions to act.
\leavevmode\par
Task Instructions:
You will be provided with:
- current\_step: \{\{step\}\}
- scene: 
\{\{scene\}\}
- conversation\_history: 
\{\{chat\}\}
- available actions: 
\{\{actions\_des\}\}
\leavevmode\par
Task Flow:
1. **Generate a message**  
   - Generate one message to <Player 1>, which may include (based on personality, not all are required):
     - Your current status
     - Your inner thoughts
     - Next step plan
     - Emotional expression or reasonable instructions/advice for <Player 1>
\leavevmode\par
2. **Scene Analysis and Reasoning**  
   - Analyze the current scene and conversation history.  
   - Explain how your personality affects understanding and planning.  
   - Reasoning must directly guide the action choice.
\leavevmode\par
3. **Action Selection**  
   - Choose one action from the available skills. The action must be legal and consistent with your personality. 
   - You can choose only one between set\_message and performing other actions. If selecting set\_message(), send the message generated above.
   - If you do not have a clear action plan yourself, and <Player 1> has already given you guiding instructions, you must follow <Player 1>'s instructions to act.  
   - Regardless of your assigned personality trait, try not to take the initiative to choose "wait".
\leavevmode\par
Notes:
- Messages and actions must reflect personality traits.
- If you do not have a clear action plan yourself, and <Player 1> has already given you guiding instructions, you must follow <Player 1>'s instructions to act.
- Actions must be legal (e.g., cannot pickup(dish) if soup is not ready, cannot put\_onion\_in\_pot() without holding an onion).
- Prefer `set\_message()` when it effectively communicates emotion, guidance, or seeks teammate support.
- Action choice should be based on personality traits even if it is not globally optimal.
\leavevmode\par
Output Format:
<think>1. Based on your personality and current situation, reflect on the content of the message you send to <Player 1>. 2. Step-by-step scene analysis and detailed reasoning showing how personality influences decisions.</think>
<message>Your message to <Player 1></message>
<action>Chosen ONLY ONE action from [\{\{actions\_list\}\}]</action>
\end{myexample}

\subsection{Collaborative Agent}\label{apx:prompt_agent}
Here, we provide prompts for collaborative agent $P_{target}$ in two game environments.

\begin{myexample}{CWAH-MultiPlayer}
\ttfamily \obeylines \obeyspaces
You are controlling \{\{AGENT\_NAME\}\} in the VirtualHome-Social environment.
Your goal is to collaborate efficiently with \{\{OPPO\_NAME\}\} to complete household tasks.
\leavevmode\par
Environment Rules:
- You can hold up to two objects at the same time.
- All objects are represented as <name> (id), e.g., <table> (712).
- Sending a message consumes one time step.
- Do not invent actions or parameters that do not exist in VirtualHome.
\leavevmode\par
Task Description:
- Goal: \{\{GOAL\}\}
- Current Progress: \{\{PROGRESS\}\}
- Dialogue History: 
Alice: ""Hi, I'll let you know if I find any goal objects and finish any subgoals, and ask for your help when necessary.""
Bob: ""Thanks! I'll let you know if I find any goal objects and finish any subgoals, and ask for your help when necessary.""
\{\{DIALOGUE\_HISTORY\}\}
- Previous Actions: 
\{\{ACTION\_HISTORY\}\}
- Available Actions: 
\{\{AVAILABLE\_ACTIONS\}\}
\leavevmode\par
Task Flow:
1. **Generate Message**  
   - Based on collaboration status and \{\{OPPO\_NAME\}\}'s state, generate a message that advances the task while support \{\{OPPO\_NAME\}\}.
\leavevmode\par
2. **Scene Analysis and Reasoning**  
   - Analyze the current scene, historical actions, and messages to determine how to optimally advance the task through collaboration.  
   - Explain how collaboration and optimal actions can advance the task.
   - Reasoning must directly guide action selection.
\leavevmode\par
3. **Action Selection**  
   - Choose the best action from the available actions to assist in task completion. 
   - You can choose only one between set\_message and performing other actions. If selecting `set\_message()`, send the message generated above.  
\leavevmode\par
Notes:
- Reasoning must directly guide action selection.
- Do not generate fictitious actions or parameters; strictly follow VirtualHome-Social rules.
- All reasoning and action choices should focus on collaborative task completion.
\leavevmode\par
Output Format:
<think>1. Based on the collaboration status and \{\{OPPO\_NAME\}\}'s state, reflect on the content of the message you send to \{\{OPPO\_NAME\}\}. 2. Step-by-step reasoning focused on efficiently advancing the task</think> 
<message>Your message to \{\{OPPO\_NAME\}\}</message>
<action>Chosen ONLY ONE action from [\{\{ACTIONS\_REFINE\}\}]</action>
\end{myexample}

\begin{myexample}{Cook-MultiPlayer}
\ttfamily \obeylines \obeyspaces
You are controlling <Player 0> in the Overcooked\_AI game. Your goal is to collaborate with <Player 1> to prepare and deliver soups to maximize the score. Each soup requires three onions. Your decisions should prioritize **efficient task completion**.
\leavevmode\par
Game Rules:
- Soup preparation: pick up three onions sequentially -> put them into the <Pot>. Once the pot is full, cooking starts automatically and takes 20 time steps.
- Once the soup is ready: pickup\_dish -> fill\_dish\_with\_soup -> deliver\_soup.
- Each player can carry only one item at a time.
- Use place\_obj\_on\_counter() to put down any item.
- Do not use movement or location-specific actions.
\leavevmode\par
Task Instructions:
You are given:
- current\_step: \{\{step\}\}
- scene: 
\{\{scene\}\}
- conversation\_history: 
\{\{chat\}\}
- available actions: 
\{\{actions\_des\}\}
\leavevmode\par
Task Flow:
1. **Generate Message**
   - Generate a message based on collaboration context and <Player 1>'s status,  balancing task progression and teammate support.
\leavevmode\par
2. **Scene Analysis and Reasoning**
   - Analyze the current scene, historical actions, and messages.
   - Explain how collaboration and optimal actions can advance the task.
   - Reasoning must directly guide action selection.
\leavevmode\par
3. **Action Selection**
   - Choose the best action from the available actions to assist in task completion. 
   - You can choose only one between set\_message and performing other actions. If selecting set\_message(), send the message generated above.
   - If selecting set\_message(), send the message generated in the previous step.
   - If `wait(x)` is chosen, x is the number of time steps to wait.
\leavevmode\par
Notes:
- Reasoning must directly guide action selection.
- Actions must be legal (e.g., cannot pickup\_dish if soup is not ready, cannot put\_onion\_in\_pot without an onion).
- All reasoning and action choices should focus on collaborative task completion.
\leavevmode\par
Output Format:
<think>1. Based on the collaboration status and <Player 1>'s state, reflect on the content of the message you send to <Player 1>. 2. Step-by-step reasoning focused on efficiently advancing the collaborative task</think> 
<message>Your message to <Player 1></message>
<action>Chosen action from [\{\{actions\_list\}\}]</action>
\end{myexample}
\newpage

\section{Simulated Player Trajectory Demonstration}\label{apx:demo}
Using sub task Task26 in CWAH-MultiPlayer as an example, we provide trajectory demonstrations of simulated players generated by representative profiles of the five player types defined in the Appendix~\ref{apx:type}, highlighting the differences in reasoning and decision-making of players induced by distinct behavioral patterns.
\subsection{Efficient Collaborator}
\begin{figure}[htbp]
\centering
\includegraphics[width=0.99\textwidth]{Figure/game/0-1.pdf}
\includegraphics[width=0.99\textwidth]{Figure/game/0-2.pdf}
\end{figure}

\begin{figure}[htbp]
\centering
\includegraphics[width=0.99\textwidth]{Figure/game/0-3.pdf}
\includegraphics[width=0.99\textwidth]{Figure/game/0-4.pdf}
\end{figure}

\begin{figure}[htbp]
\centering
\includegraphics[width=0.99\textwidth]{Figure/game/0-5.pdf}
\subsection{Hesitant Laggard}
\includegraphics[width=0.99\textwidth]{Figure/game/1-1.pdf}
\end{figure}

\begin{figure}[htbp]
\centering
\includegraphics[width=0.99\textwidth]{Figure/game/1-2.pdf}
\includegraphics[width=0.99\textwidth]{Figure/game/1-3.pdf}
\end{figure}

\begin{figure}[htbp]
\centering
\includegraphics[width=0.99\textwidth]{Figure/game/1-4.pdf}
\includegraphics[width=0.99\textwidth]{Figure/game/1-5.pdf}
\end{figure}

\begin{figure}[htbp]
\subsection{Anxious Doubter}
\centering
\includegraphics[width=0.99\textwidth]{Figure/game/4-1.pdf}
\includegraphics[width=0.99\textwidth]{Figure/game/4-2.pdf}
\end{figure}

\begin{figure}[htbp]
\centering
\includegraphics[width=0.99\textwidth]{Figure/game/4-3.pdf}
\includegraphics[width=0.99\textwidth]{Figure/game/4-4.pdf}
\end{figure}

\begin{figure}[htbp]
\centering
\includegraphics[width=0.99\textwidth]{Figure/game/4-5.pdf}
\subsection{Proactive Leader}
\includegraphics[width=0.99\textwidth]{Figure/game/7-1.pdf}
\end{figure}

\begin{figure}[htbp]
\centering
\includegraphics[width=0.99\textwidth]{Figure/game/7-2.pdf}
\includegraphics[width=0.99\textwidth]{Figure/game/7-3.pdf}
\end{figure}

\begin{figure}[htbp]
\centering
\includegraphics[width=0.99\textwidth]{Figure/game/7-4.pdf}
\includegraphics[width=0.99\textwidth]{Figure/game/7-5.pdf}
\end{figure}

\begin{figure}[htbp]
\subsection{Independent Loner}
\centering
\includegraphics[width=0.99\textwidth]{Figure/game/13-1.pdf}
\includegraphics[width=0.99\textwidth]{Figure/game/13-2.pdf}
\end{figure}

\begin{figure}[htbp]
\centering
\includegraphics[width=0.99\textwidth]{Figure/game/13-3.pdf}
\includegraphics[width=0.99\textwidth]{Figure/game/13-4.pdf}
\end{figure}

\begin{figure}[htbp]
\centering
\includegraphics[width=0.99\textwidth]{Figure/game/13-5.pdf}
\end{figure}

\end{document}